\documentclass{article}
\usepackage{fancybox}
\usepackage{soul}
\usepackage{color}
\usepackage{url}
\usepackage[running, mathlines]{lineno}
\usepackage[pdftex]{graphicx}
\usepackage[cmex10]{amsmath}
\usepackage{amsfonts}

\newcommand{\mb}{\mathbf}

\newcommand{\blue}[1]{\textcolor{black}{#1}}

\newif\ifREVIEW
\REVIEWfalse

\newif\ifDCOL
\DCOLfalse

\ifREVIEW
  \linenumbers
\fi

\title{Dependent landmark drift: robust point set registration with 
       a Gaussian mixture model and a statistical shape model}
\author{ Osamu Hirose }

\date{}

\begin{document}
\maketitle
\thispagestyle{empty}

\vspace{-6mm}
\begin{center}
Faculty of Electrical and Electronic Engineering,
Institute of Science and Engineering, Kanazawa University, \\
Kakuma, Kanazawa, Ishikawa 920-1192, JAPAN. \\
E-mail: hirose@se.kanazawa-u.ac.jp\\
\end{center}

\noindent

\noindent {\bf Abstract.}
The goal of point set registration is to find point-by-point correspondences between point sets,
each of which characterizes the shape of an object.
Because  local preservation of object geometry is assumed,
prevalent algorithms in the area can often elegantly solve the problems without 
using geometric information specific to the objects.
This means that registration performance can be further improved by using prior knowledge 
of object geometry.
In this paper, we propose a novel point set registration method using the Gaussian mixture model 
with prior shape information encoded as a statistical shape model.
Our transformation model is defined as a combination of the similarity transformation, motion coherence, 
and the statistical shape model. Therefore, the proposed method works effectively if the target point set 
includes outliers and missing regions, or if it is rotated. 
The computational cost can be reduced to linear, and therefore the method is scalable to large point sets.
The effectiveness of the method will be verified through comparisons with existing algorithms
using datasets concerning human body shapes, hands, and faces.

\vspace{1mm}
\noindent {\bf Keywords:}
Point set registration, Gaussian mixture model, statistical shape model, EM algorithm

\vspace{-3mm}
\section{Introduction}

Point matching is the problem of finding point-by-point correspondence between point sets
where each set characterizes the geometry of an object. Finding such geometrical correspondences
between point sets is studied in the fields of image recognition, computer vision, and, computer graphics.
A major class of point matching problems is point set registration---the problem of finding
a transformation between two point sets in addition to point-by-point correspondences.
Point set registration problems can be roughly classified into two classes according to the 
transformation models: rigid and non-rigid transformations. % \cite{Ma2016}.
A rigid transformation model is defined as a linear map that preserves the relative positions of points
in a point set, i.e., rotation and translation. The rigid point set registration problem
is a relatively simple problem that has been intensively studied
\cite{Brown1992,Besl1992,Rusinkiewicz2001,Fitzgibbon2003,Makadia2006}.
Non-rigid registration is a more complex problem that involves transforming an object's geometry.
Typical transformation models used for point set registration problems are the thin-plate spline functions
\cite{Chui2000,Chui2003,Jian2011,Chen2015,Yang2015} and Gaussian functions
\cite{Myronenko2006,Myronenko2010,Ma2015,Ma2016}.
These methods are differently classified according to the definitions of the point set registration problem:
energy minimization \cite{Chui2000,Chui2003,Yang2015} and probabilistic density estimation
using a Gaussian mixture model \cite{Jian2011,Chen2015,Myronenko2006,Myronenko2010,Ma2015,Ma2016}.

Crucial to the success of these registration methods is a robustness to outliers, points that are
irrelevant to the true geometry of the object. Several approaches have been proposed to deal with outliers,
such as statistical analysis for distances between correspondent points \cite{Zhang1994,Granger2002,Stewart2003},
soft assignments \cite{Chui2003,Rangarajan1997a}, trimming point sets through iterative random sampling
\cite{Chetverikov2005}, kernel correlation \cite{Tsin2004}, explicit probabilistic modeling of outliers
\cite{Myronenko2006,Myronenko2010,Ma2015,Ma2016}, and the use of robust estimators 
such as the $L_2E$ estimator \cite{Jian2011} and a scaled Geman--McClure estimator \cite{Zhou2016}.
Another factor crucial to the success of registration methods is the assumption of a smooth displacement field, 
which forces neighboring points to move coherently. %in a floating point set to move coherently.
The smoothness of a displacement field, also known as the motion coherence, is imposed by a regularization
techniques \cite{Chui2000,Chui2003,Yang2015,Jian2011,Chen2015,Myronenko2006,Myronenko2010,Ma2015,Ma2016}.
%% technique defined as a penalty term for energy minimization \cite{Chui2000,Chui2003,Yang2015} 
%% and a penalty term for log-likelihood functions \cite{Jian2011,Chen2015,Myronenko2006,Myronenko2010,Ma2015,Ma2016}.
Owing to the assumption of a smooth displacement field, such non-rigid registration algorithms seek to find 
transformations with sufficient global flexibility while preserving the local topology of a point set.

These methods are universally applicable to general point set registration problems as no prior knowledge,
except for that concerning the smoothness of a displacement field, is assumed for the geometry of objects to be registered. 
This means that registration performance can be improved by using prior knowledge of object geometry.
%instead of sacrificing applicability.
%
One approach to incorporating such prior information is the use of a kinematic motion for articulated
objects, such as the human body \cite{Mateus2008,Horaud2011,Ge2015a,Ge2015b}.
These methods show promising results but cannot be applied to objects with no kinematics.
Another approach is the use of partial correspondence across two point sets \cite{Kolesov2016,Golyanik2016}. 
These methods also show promising results, but better performance is not expected if partial correspondence 
is not available.
Yet another approach is the use of supervised learning techniques. One candidate of such supervised learning 
techniques is a statistical shape model \cite{Cootes1995,Cootes2004}
that describes the mean shape and statistical variation of geometrical objects. Shape variations
represented by statistical shape models are constructed from shape statistics of landmark displacements.
%which are usually obtained manually. 
Therefore, movements of neighboring landmarks are not assumed to be
correlated, and those of distant landmarks are allowed to be dependent on one another.
The statistical shape models also do not require any physical models such as object kinematics.

Statistical shape models have been widely used as prior shape information for various tasks,
including image registration, image segmentation, and surface reconstruction, in the fields of 
computer graphics, computer vision, and medical imaging. A broad range of these works were 
reviewed in several survey papers \cite{Heimann2009,Oliveira2012,Sotiras2013,Berger2014},
and lecture notes \cite{Bouaziz2013,Bouaziz2016}.
In the field of computer graphics and vision, statistical shape models have mainly been used for reconstruction
or completion of smooth 3D surfaces \cite{Blanz1999,Anguelov2005,Park2008,Jain2010,Yang2014}, which can be
interpreted in a wider sense as point set registration problems with extra information such as 
colors, surface normals, and the articulation of the human body shape.
These works include applications related to the reconstruction of 3D human surfaces such as
faces \cite{Blanz1999}, body shapes and poses \cite{Anguelov2005,Jain2010}, skins and muscles \cite{Park2008}, 
and body shapes \cite{Yang2014}. 

In the field of medical imaging, the point set registration problems with statistical
shape models have often been developed in noisier settings, i.e., in the presence of outliers
\cite{Hufnagel2008,Rasoulian2012,Ravikumar2016}, and these methods were later extended to 
image registration \cite{Wang2016} and image segmentation \cite{Kruger2016}. 
All these methods are designed on the basis of non-rigid registration techniques such as
probabilistic density estimation with a Gaussian mixture model \cite{Hufnagel2008,Rasoulian2012},
the one with a $t$-mixture model \cite{Ravikumar2016}, and the minimization of an
energy function \cite{Wang2016}.
Typically, these methods solve either or both of the two different problems.
The first problem, which is sometimes called group-wise point set registration, involves 
finding point-by-point correspondence among multiple unstructured point sets, 
aiming at learning statistical shape models \cite{Hufnagel2008,Rasoulian2012,Ravikumar2016,Wang2016}.
The second problem involves finding point-by-point correspondence between a point set that forms the average shape
in a pre-trained statistical shape model and a new unstructured point set that forms an object shape
\cite{Rasoulian2012,Wang2016}. More recently, a method of 3D surface reconstruction,
originating in these works, has been proposed with an efficient use of surface normals \cite{Bernard2017}.
 
In this article, we focus on the second problem, motivated by the fact that 
(a) the scalability of the algorithms for the second problem is relatively limited, because their
computational costs are proportional to the product of the numbers of the two point sets to be registered, 
and (b) their formulation did not explicitly include the rotational term and 
they might not be robust against a rotated target shape. 
As in \cite{Bernard2017}, we assume that 
training shapes or pre-trained statistical shape models are available as prior shape information.
This assumption is somewhat strong but is not unrealistic, because training shapes with point-by-point 
correspondence or pre-trained shape models are often publicly available; for example, 
2D human hands and faces \cite{Stegmann2002}, 3D human poses and shapes \cite{Anguelov2005},
and 3D human shapes \cite{Yang2014}. 
In test cases, we do not assume the use of any extra information such as color, surface normals,
the articulation in the body parts, or 
the partial correspondence between point sets, in order not to reduce the applicability of the method.

\subsubsection*{Contributions of this work}

We propose a novel non-rigid point set registration algorithm based on a supervised
learning approach that we call dependent landmark drift (DLD). 
In the context of statistical shape modeling,
the proposed algorithm is a novel optimization algorithm for fitting a statistical shape model, 
where one input point set forms the mean shape and another forms a target shape. 
The proposed algorithm is designed as a Gaussian mixture model in which the transformation
model is defined as a combination of the similarity transformation and the statistical shape model.
Among the related algorithms, %with statistical shape models 
%\cite{Rasoulian2012,Bernard2017}, 
our approach is novel in that it simultaneously provides
(1) fast computation, with computational costs proportional to the sum of the numbers of 
two point sets to be registered,
(2) simultaneous optimization of scale, translation, rotation, and shape deformation, and
(3) adaptive control of the smoothness of a displacement field.
In addition, our algorithm retains the merits of the registration algorithms based on the Gaussian mixture model,
i.e., the automatic radius control and robustness against outliers,
originating in the explicit probabilistic modeling of outliers.
The effectiveness of the proposed algorithm was tested through comparisons
with supervised and unsupervised point set registration algorithms.

\subsubsection*{Related works}
%\section{Related works}

In computer graphics, shape matching algorithms, originating in point set registration algorithms,
have been actively developed. % for the reconstruction and automatic generation of 3D surfaces. 
The most common method for matching two shapes in computer graphics is the iterative closest point algorithm
(ICP) \cite{Bouaziz2016}. The ICP solves the registration problems by iterating the two-step procedure:
(1) to find the point in the target surface that is closest to each point in the current deformed surface and
(2) to update the deformed surface using the current pairs of closest points.
In this field, the ICP has been customized to solve shape matching problems specific to the types of the object
such as a human face \cite{Cosker2011,Li2017}, hand shape and pose \cite{Hirshberg2012,Romero2017}, 
body shape and pose \cite{Allen2003,Kraevoy2005,Anguelov2005,Hasler2009,Bogo2014}, and dynamic human shape 
in motion \cite{Pons-Moll2015}. The shapes registered by these algorithms can be used for building 
blend shapes and shape models, and they have been successfully applied to video-retouching \cite{Jain2010} 
and the automatic synthesis of 3D human animation \cite{Hasler2009,Loper2015,Pons-Moll2015}.

The main difference between shape matching algorithms in computer graphics and the general point set 
registration algorithms is in the design of the objective function that measures a dissimilarity between
two shapes. The objective functions of general point set registration problems are typically based on the 
distance between points and the smoothness of a displacement field. On the contrary, the objective functions
of the shape matching algorithms is designed as a combination of the distance between points and
various information such as the user-defined marker errors and the smoothness of the deformed surface 
\cite{Allen2003,Hasler2009}, constraints on pose and shape deformation \cite{Anguelov2005,Hirshberg2012,Romero2017},
and consistency with additional information such as color and brightness \cite{Bogo2014,Li2017}.

The registration algorithms that are not domain-specific have also been developed in computer graphics.
%% The main issue to be addressed is that the quality of the registration strongly depends on the
%% the quality of the pre-alignment. To address this issue, 
Mitra et al. proposed a rigid alignment algorithm with the objective function that efficiently approximates 
the point-to-surface distance using the d2Tree \cite{Mitra2004}. Li et al. proposed a non-rigid registration 
method that incorporates the rigid transformation, local affine transformation, and the spatial coherence 
of the deformation \cite{Li2008}. Gao et al. proposed a representation of a shape, called rotation-invariant 
local mesh differences (RIMD) to remove the rotational effect on the dissimilarity measure between two shapes 
\cite{Gao2016}. The global optimization method that solves non-rigid registration problems using a convex 
optimization technique was reported by \cite{Maron2016}.

%% point-by-point correspondence can be spurious
%% unless the initialization with sufficient quality is conducted before the registration.

%% the estimation of point-by-point correspondence by the ICPs;
%% the correspondence is typically estimated as a pair of points with the minimum distance,
%% and thereby, can be spurious unless the initialization with sufficient quality is conducted before the registration.

%% \ifDCOL
%% \else
%% The implementation of the DLD algorithm is available at \url{https:/github.com/ohirose/dld}. 
%% \fi

\section{Methods}
The goal of the point set registration is to find the map $\mathcal{T}$ that
transforms the geometric shape represented as a point set $Y=\{y_1,\cdots,y_M\}$
and matches the target shape represented as the other point set $X=\{x_1,\cdots,x_N\}$.
The set of $\mathcal{T}(y)-y$ for any $y\in\mathbb{R}^D$ is called a {\em displacement field}
as it defines the displacement of each point in $Y$. Generally, the point set 
registration problem is defined as a minimization problem as follows:
\begin{linenomath*}
\begin{align}
  \hat{\mathcal{T}} = \text{argmin}_\mathcal{T} \mathcal{L} (X, \mathcal{T}(Y)) 
   + \gamma \mathcal{R}(\mathcal{T}),
  \label{eq:registration}
\end{align}
\end{linenomath*}
where $\mathcal{L}$ is a loss function that measures the dissimilarity between two geometric shapes,
$\mathcal{R}$ is a functional that measures the complexity of a displacement field, and
$\gamma >0$ is a parameter that balances the similarity of the shapes and the smoothness of the
displacement field. 
In this study, we use a statistical shape model (SSM) with the similarity transformation
as the transformation model $\mathcal{T}$ and use the negative log-likelihood of a 
Gaussian mixture model (GMM) as the loss function $\mathcal{L}$. 
To clarify the merits in the choice, we review the SSM based on the principal component analysis
\cite{Cootes2004} and then introduce a GMM as a technique of the point set registration
\cite{Myronenko2006,Myronenko2010}. We then propose a novel registration method, and finally,
discuss the reduction in its computational cost. The list of symbols used throughout this paper
is available in Appendix.

\subsection{Statistical shape model}
We begin with definitions of a landmark and a shape, required to define a statistical
shape model. To obtain shape statistics from multiple geometric objects, it is common to
define correspondent points across them. These points of correspondence
are called {\em landmarks}. A {\em shape} is typically defined as a set of landmarks for one
of the geometric objects with scale, rotation, and translation effects removed \cite{Stegmann2002,Dryden2016}.
Note that a point set and a shape are distinguished from each other in that
(1) shapes are composed of the same number of points whereas the number of points in a point set
is generally different from those in other point sets;
and (2) points in shapes are correspondent across all shapes, whereas points in multiple point sets
are not correspondent. Statistical shape models are constructed from training shapes, i.e.,
multiple point sets with point-by-point correspondence.

\subsubsection*{Definition of a statistical shape model}
The SSM is a representation of a geometrical shape and its statistical variations
in an object.
Because definitions of SSMs diverge according to the aim of the application or the method of construction
\cite{Heimann2009}, we introduce a definition based on the principal component analysis (PCA)
\cite{Cootes1995,Cootes2004,Stegmann2002} that can adequately describe the proposed algorithm.
Suppose a shape is composed of $M$ landmarks $(v_1,\cdots,v_M)$, each of which lies in a
$D$-dimensional space. Then, the shape is represented as a vector 
$\mathbf{v}=(v_{1}^{T},\cdots,v_{M}^{T})^T \in \mathbb{R}^{MD}$.
The PCA-based statistical shape model is defined as follows:
\begin{linenomath*}
\begin{align}
  \mb{v}= \mb{u} +\sum_{k=1}^K z_k \mb{h}_k +\mb{w},
  \label{eq:ssm}
\end{align}
\end{linenomath*}
where $\mb{u}= (u_{1}^{T},\cdots,u_{M}^{T})^T\in \mathbb{R}^{MD}$ is the mean shape,
$\mb{h}_k \in \mathbb{R}^{MD}$ is the $k$th leading shape variation, $z_k \in \mathbb{R}$ is the $k$th
weight corresponding to the $k$th shape variation, $K$ is the number of shape variations, 
and $\mb{w} \in\mathbb{R}^{MD}$ is a residual vector. To reduce information sharing in shape variations
to the maximum extent, $\mb{h}_1,\cdots,\mb{h}_K$ are usually assumed to satisfy the orthonormality
condition $\mb{h}_{i}^T \mb{h}_{j}=\delta_{ij}$ % for all pairs of $i$ and $j$ such that $i\neq j$.
where $\delta_{ij}$ is Kronecker's delta.

\subsubsection*{Estimation of shape variations from training data}
Mean shape $\mb{u}$ and shape variations $\mb{h}_1, \cdots, \mb{h}_K$
are unknown, and should be estimated from multiple shapes ${\mb{v}_1,\cdots,\mb{v}_B}$, 
i.e., training data. The mean shape $\mb{u}$ is simply estimated as the average of sample shapes.
Suppose $\mb{C}\in \mathbb{R}^{MD\times MD}$ is a shape covariance matrix defined as
\begin{linenomath*}
\begin{align}
  \mb{C}=\frac{1}{B-1} \sum_{j=1}^B (\mb{v}_j-\bar{\mb{v}})(\mb{v}_j-\bar{\mb{v}})^{T},
\end{align}
\end{linenomath*}
where $\bar{\mb{v}}$ is the sample average of shapes $\mb{v}_1,\cdots,\mb{v}_B$.
The shape covariance matrix $\mb{C}$ represents statistical dependencies for landmark displacements.
The $k$th shape variation can be estimated as the $k$th eigenvector of $\mb{C}$ corresponding
to the $k$th largest eigenvalue. 
We note that the shape covariance matrix is dependent on rotations of shapes
$\mb{v}_1,\cdots,\mb{v}_B$. On the contrary, coherent moves of landmarks caused by the rotation 
of a whole shape cannot be represented as their covariance if the rotation angle is relatively large. 
Therefore, effects of rotations should be eliminated from the training shapes $\mb{v}_1 \cdots, \mb{v}_B$
in computing the shape covariance matrix $\mb{C}$.

\subsection{Gaussian mixture model for point set registration}
We summarize the Gaussian mixture modeling approach for solving point set registration problems
proposed by Myronenko et al. \cite{Myronenko2010} as this approach is the basis for the proposed algorithm.
They defined a registration problem of two point sets as a problem of probabilistic density estimation,
where one point set is composed of centroids for a GMM, and the other consists of samples generated from the GMM. 
Here, we refer to a point set to be deformed as a source point set and the other point set 
that remains fixed as a {\em target} point set.
We also refer to a point that is irrelevant to the true object geometry as an {\em outlier} and one
that represents the true object geometry as an {\em inlier}.

\subsubsection*{Definition of the model}

Let $x_n\in \mathbb{R}^{D}$ and $y_m\in \mathbb{R}^{D}$
be the $n$th point in a target point set $X=\{x_1,\cdots,x_N\}$ and the $m$th point 
in a source point set ${Y}=\{y_1,\cdots,y_M\}$, respectively. 
The probabilistic model for registering the two point sets ${X}$ and ${Y}$ %proposed by Myronenko et al. 
is designed as a mixture model for generating target point $x_n$ in the four-step procedure: 
(1) a label is selected as an outlier or an inlier based on the Bernoulli distribution 
with outlier probability $\omega$, 
(2) if the label is an inlier, a point is selected from the source point set 
    $Y=\{y_1,\cdots,y_M\}$ with equal probability $p(m)={1}/{M}$,
(3) the selected point $y_m$ is moved by the transformation model $\mathcal{T}$, and 
(4) target point $x_n$ is generated by the Gaussian distribution whose center is the moved point.
More formally, the mixture model is defined as follows:
\ifDCOL
  \begin{align}
    p(x_n;\Theta) = & \omega\cdot p_{\text{outlier}}(x_n) \nonumber \\
                    & + (1-\omega) \cdot \sum_{m=1}^{M} p(m) p(x_n|m;\Theta),
    \label{eq:mixture}
  \end{align}
\else
  \begin{linenomath*}
  \begin{align}
    p(x_n;\Theta) = \omega\cdot p_{\text{outlier}}(x_n) + (1-\omega) \cdot \sum_{m=1}^{M} p(m) p(x_n|m;\Theta),
    \label{eq:mixture}
  \end{align}
  \end{linenomath*}
\fi
where $\Theta$ is a set of parameters of the mixture model, and $p_{\text{outlier}}(x_n)$ is a distribution of outliers. 
The prior distribution of inliers is defined as $p(m)=1/M$. 
The inlier distribution $p(x_n|m;\Theta)$ for $m=1,\cdots,M$ is defined as a Gaussian distribution
\begin{linenomath*}
\begin{align}
p(x_n|m;\Theta)=
  \frac{1}{(2\pi\sigma^2)^{D/2}}
    \exp
    \bigg(
      -\frac{||x_n-\mathcal{T}(y_m;\theta)||^2}{2\sigma^2}
    \bigg),
  \label{eq:gauss}
\end{align}
\end{linenomath*}
where $\sigma^2$ is the variance of the Gaussian distribution, $\mathcal{T}(y_m;\theta)$ is 
a transformation model for source point $y_m$ with
a set of parameters $\theta$, and $\Theta=(\theta,\sigma^2)$ is a set of parameters of the GMM.

\subsubsection*{Problem definition as a probabilistic density estimation}
Under the model construction, the problem is to find the best transformation $\mathcal{T}(y_m;\theta)$
that matches the target point set $X$. Because the parameter $\theta$ characterizes the transformation 
$\mathcal{T}(y_m;\theta)$, the solution of the point set registration problem is 
obtained by finding the best parameter $\hat\theta$. More formally, the point set registration problem is defined
as a probabilistic density estimation as follows:
\begin{linenomath*}
\begin{align}
  \hat{\Theta}=\text{argmin}_\Theta 
                   \bigg\{
                     -\sum_{n=1}^N \log p(x_n;\Theta) +\gamma\mathcal{R}(\mathcal{T})
                   \bigg\},
  \label{eq:likelihood}
\end{align}
\end{linenomath*}
where $\Theta=(\theta,\sigma^2)$, and $\mathcal{R}(\mathcal{T})$ is the regularizer in 
the motion coherence theory \cite{Yuille1989}.
We note that this formulation is a special case of 
the general definition of the point set registration problems, defined as Eq. (\ref{eq:registration}),
as the negative log-likelihood function is a type of loss function $\mathcal{L}$ that measures 
the dissimilarity between geometric shapes represented as point sets.

\subsubsection*{Optimization by the EM algorithm}
Because the analytic solution for Eq. (\ref{eq:likelihood}) is not available, the EM algorithm is used 
to search for a local minimum of the function. 
The EM algorithm iteratively improves a solution by updating the upper bound of the negative
log-likelihood function, called the $Q$-function. Given a current estimate
$\bar\Theta=(\bar\sigma^2,\bar\theta)$, the $Q$-function for the GMM is derived as
\ifDCOL
  \begin{align}
    Q(\Theta,\bar{\Theta}) &=
          \frac{N_P D}{2}\log \sigma^2 + \gamma \mathcal{R}(\mathcal{T}) \nonumber \\   
        &+ \frac{1}{2\sigma^2} \sum_{n=1}^{N} \sum_{m=1}^M p(m|x_n;\bar\Theta) ||x_n-\mathcal{T}(y_m;\theta)||^2,
  \label{eq:Q}
  \end{align}
\else
  \begin{linenomath*}
  \begin{align}
    Q(\Theta,\bar{\Theta}) =
          \frac{N_P D}{2}\log \sigma^2
        + \frac{1}{2\sigma^2} \sum_{n=1}^{N} \sum_{m=1}^M p(m|x_n;\bar\Theta) ||x_n-\mathcal{T}(y_m;\theta)||^2
        + \gamma \mathcal{R}(\mathcal{T})
  \label{eq:Q}
  \end{align}
  \end{linenomath*}
\fi
where $N_P=\sum_{n=1}^{N}\sum_{m=1}^{M} p(m|x_n,\bar\Theta)\leq N$ is the effective number
of matching points, and $p(m|x_n,\bar\Theta)$ denotes the posterior probability that source point $y_m$ is
correspondent to target point $x_n$ under the current estimate $\bar\Theta$. 
The posterior probability $p(m|x_n;\bar{\Theta})$ can be calculated as
\ifDCOL
  \begin{align}
    p(&m|x_n;\bar\Theta)  \nonumber\\
     &= \frac
     {          (1-\omega)      p(x_n|m; \bar\Theta)                           }
     { \omega p_{\text{outlier}}(x_n) +(1-\omega)\frac{1}{M}\sum_{m'=1}^{M} p(x_n|m'; \bar\Theta) }.
    \label{eq:posterior}
  \end{align}
\else
  \begin{linenomath*}
  \begin{align}
    p(m|x_n;\bar\Theta) = \frac
     {          (1-\omega)      p(x_n|m; \bar\Theta)                           }
     { \omega p_{\text{outlier}}(x_n) +(1-\omega)\frac{1}{M}\sum_{m'=1}^{M} p(x_n|m'; \bar\Theta) }.
    \label{eq:posterior}
  \end{align}
  \end{linenomath*}
\fi
Based on the theory of the EM algorithm,
a solution of the point set registration problem is obtained by iterating the following procedure:
(i) updating the posterior probability $p(m|x_n; \bar\Theta)$,
(ii) finding the $\Theta^*$ that minimizes the $Q$-function for $\Theta$, given the parameter set $\bar\Theta$,
and (iii) replacing the given parameter set $\bar\Theta$ with the minimizer $\Theta^*$ of the $Q$-function.
This procedure is iterated until a suitable convergence criterion is satisfied.

\subsubsection*{Relation to ICPs and the automatic radius control}

There is a close relationship between the methods based on the GMM and iterative closest point
algorithms (ICPs), which are more common in computer graphics. 
Let $\mathcal{N}_m\subset \{1,\cdots,N\}$ be an index set composed of the neighbors of $\mathcal{T}(y_m)$
in the target point set $X$. The ICPs for point set registration problems can be defined as follows:
\begin{linenomath*}
\begin{align}
    \hat{\mathcal{T}}=\text{argmin}_{\mathcal{T}} 
       \sum_{m=1}^M \sum_{n\in\mathcal{N}_m} g_{mn} ||x_n-\mathcal{T}(y_m)||^2 
       + \gamma \mathcal{R}(\mathcal{T}),
    \label{eq:icp}
\end{align}
\end{linenomath*}
where $g_{mn}$ is the matching weight between point $x_n$ and $y_m$.
%and $\mathcal{N}_m$ is typically composed of the point closest to $\mathcal{T}(y_n)$ only.  
The loss functions of ICPs and the GMM are the sum of the squared distance weighted 
by $g_{mn}$ and $p_{mn}=p(m|x_n;\Theta)$, respectively.
Therefore, the GMM and ICPs are the same in that they are included in the general class of
the point set registration problem, defined as Eq. (\ref{eq:registration}), with the same type of the
loss functions.
%% On the other hand, these methods are diffenrent from the existence of the residual vaarince $\sigma^2$
%% in the definitions of the registration problem
%%  and can be estimated on the basis of
%% the maximum likelihood esitmation whereas it is not included in the 

On the contrary, there exists a difference in the definitions of the registration problems that
yields the difference in the optimization trajectories and resulting deformed shapes:
the residual variance $\sigma^2$.
%The residual variance $\sigma^2$ is included in the $Q$-function for the GMM.
It can be intuitively interpreted as the parameter that defines the set of neighbors $\mathcal{N}_m$ 
since $p_{mn}\propto \exp \{-||x_n-T(y_m;\theta)||^2/2\sigma^2\}$ and the corresponding loss term 
$p_{mn} ||x_n-T(y_m;\theta)||^2$ in the $Q$-function 
%defined as Eq. (\ref{eq:Q}) 
become nearly zero if the distance between $x_n$ and $T(y_m;\theta)$ is sufficiently large e.g., $5\sigma$. 
Therefore,
the sequential update of $\sigma^2$ based on the EM algorithm dynamically changes the set of neighbors
$\mathcal{N}_m$ with a mutable radius during the optimization.
% based on the maximum likelihood principle.
For the ICPs, the set of neighbors $\mathcal{N}_m$ is typically set to the closest point
to the target surface \cite{Bouaziz2016}.
We, therefore, call the characteristic that yields the difference between the GMM and ICPs the 
{\em automatic radius control}.

%% We note that the residual variance $\sigma^2$ corresponds to temperature ${T}$ used in the annealing step 
%% for thin plate spline robust point matching (TPS-RPM) \cite{Chui2000}. 
%% From Eq.~(\ref{eq:gauss}) and Eq.~(\ref{eq:posterior}), 
%% we see that matching between point sets tends to be more random if $\sigma^2$ is large and less 
%% random if it is small. 
%% To avoid local minima, TPS-RPM requires some heuristics to reduce
%% the temperature ${T}$ during optimization. On the contrary, $\sigma^2$ can be estimated during
%% optimization, and thus no heuristics are required for decreasing $\sigma^2$.

\subsection{Our approach}

We propose a novel algorithm called dependent landmark drift (DLD) for registering 
the mean shape and a novel point set. The algorithm uses the same GMM framework as the CPD algorithm.
The merits of using a GMM, such as the explicit modeling of outliers and the automatic radius control,
are thereby inherited by the proposed algorithm.
The main difference between the CPD and the DLD is in the definition of transformation models for a source
point set:
the transformation model of the CPD is based on motion coherence, i.e., moving points under a smooth displacement
field, whereas that of the DLD is a combination of a statistical shape model, a similarity transformation,
and motion coherence. 

\subsubsection*{Statistical shape model as a transformation model}

We first describe that statistical shape models can be utilized as a transformation model for 
GMM-based point set registration. 
Suppose $\mb{H}=(\mb{h}_1,\cdots,\mb{h}_K)\in \mathbb{R}^{MD\times K}$
is the matrix notation of $K$ shape variations.  We also suppose $H_{m}\in \mathbb{R}^{D \times K}$ is
the submatrix of $\mb{H}$ that corresponds to the $m$th landmark  $u_m \in \mathbb{R}^{D}$ in the mean 
shape $\mb{u}\in\mathbb{R}^{MD}$.
Then, the statistical shape model (\ref{eq:ssm}) is denoted by a point-by-point transformation model
\begin{linenomath*}
\begin{align}
  v_m = {u_m} + H_{m}z+{w_m},
  \label{eq:transform}
\end{align}
\end{linenomath*}
where $v_m$ is the $m$th point in shape $\mb{v}\in \mathbb{R}^{D}$, $z=(z_1,\cdots,z_K)^T$ $\in\mathbb{R}^K$ 
is a weight vector for $K$ shape variations, and $w_m \in \mathbb{R}^D$ is a subvector of the residual vector 
$\mb{w}$ corresponding to landmark $u_m$. Therefore, the statistical shape model can be used as a transformation 
model of a point set registration.
A merit of using statistical shape models for point set registration problems is that moves of landmarks
are estimated based on the {\em statistical dependency} of landmark displacements. We therefore
call our algorithm {\em dependent landmark drift}.

\subsubsection*{Transformation model}

We incorporate prior shape information into the transformation model.
We suppose that the training phase has already completed, that is, the mean shape $\mb{u}=(u_1^T,\cdots,u_M^T)^T$
and shape variations $\mb{H}=(\mb{h}_1,\cdots,\mb{h}_K)$, also denoted by $\mb{H}=(H_1^T,\cdots,H_M^T)^T$,
have been calculated from training shapes.
We then define the transformation model $\mathcal{T}_\text{DLD}$ as a combination of the similarity transformation 
and the statistical shape model, encoding prior shape information as follows:
\begin{linenomath*}
\begin{align}
  \mathcal{T}_\text{DLD}(u_m;\theta)= sR(u_m+H_mz)+d,
  \label{eq:transform:dld}
\end{align}
\end{linenomath*}
where $s\in\mathbb{R}$ is a scale parameter, $R\in \mathbb{R}^{D\times D}$ is a rotation matrix, 
and $d\in\mathbb{R}^{D}$ is a translation vector. 
We incorporated the similarity transformation into the transformation model %to register $X$ and $U$
because the effects of scale, translation, and rotation are typically eliminated from the statistical shape model 
during training. We refer to the triplet $\rho=(s,R,d)$ as {\em location parameters} and $z$ as {\em shape parameters}.
We note that the transformation model is different from \cite{Rasoulian2012,Wang2016} in that
our model combines the similarity transformation and the non-rigid transformation with a PCA model,
whereas their transformation model is defined as the PCA model only.

\subsubsection*{Definition of $Q$-function}
We use the Gaussian mixture model, defined as Eqs. (\ref{eq:mixture}) and (\ref{eq:gauss}), 
as the basis of solving the point set registration problems. 
We suppose $\lambda_k \in \mathbb{R}$ is the $k$th largest eigenvalue of the covariance matrix $\mb{C}$ 
computed from training data and $\Lambda_K \in \mathbb{R}^{K\times K}$ is the diagonal matrix defined as 
$\Lambda_K=\text{d}(\lambda_1,\cdots,\lambda_K)$, 
where $\text{d}(\cdot)$ denotes the operation of converting a vector into a diagonal matrix.
Replacing the transformation model $\mathcal{T}$ in Eq. (\ref{eq:Q}) by 
$\mathcal{T}_\text{DLD}$, we define the $Q$-function of the point set registration problem as
%
%\vspace{-0.1cm}
\ifDCOL
  \begin{align}
      Q_\text{DLD}(&\Theta,\bar{\Theta}) =
             \frac{N_P D}{2}\log \sigma^2 +\gamma z^T\Lambda_K^{-1} z \nonumber \\
          &+ \frac{1}{2\sigma^2}
               \sum_{n=1}^{N} \sum_{m=1}^M p_{mn} %`p(m|x_n,\bar\Theta)
                 ||x_n-sR(u_m+H_m z)-d||^2,  
  \end{align}
\else
  \begin{linenomath*}
  \begin{align}
      Q_\text{DLD}(\Theta,\bar{\Theta})=
            \frac{N_P D}{2}\log \sigma^2
          + \frac{1}{2\sigma^2}
               \sum_{n=1}^{N} \sum_{m=1}^M p_{mn} %`p(m|x_n,\bar\Theta)
                 ||x_n-sR(u_m+H_m z)-d||^2
          + \gamma z^T \Lambda_K^{-1} z,
  \end{align}
  \end{linenomath*}
\fi
where $\Theta=(z,\rho,\sigma^2)$, $p_{mn}=p(m|x_n;\bar\Theta)$, and 
\ifDCOL
  $N_P=\sum_{n=1}^N$ $\sum_{m=1}^M p_{mn}$.
\else
  $N_P=\sum_{n=1}^N\sum_{m=1}^M p_{mn}$.
\fi
We used the Tikonov regularizer $z^T \Lambda_K^{-1} z$ as a regularization term 
to avoid searching for extreme shapes, where $\gamma >0$ is a parameter that controls the search space of $z$. 
We note that this regularization is interpreted as a smooth displacement field because the resulting 
transformed shape becomes increasingly similar to the mean shape as $\gamma$ increases, i.e., the neighboring 
points move coherently. The prior shape information in our approach is therefore a combination of the
statistical shape model, the similarity transformation, and motion coherence. 

\subsubsection*{Outlier distribution and matching probabilities}
We note that the outlier distribution $p_{\text{outlier}}(x_n)$ included in the mixture model
defined as Eq. (\ref{eq:mixture}) is an arbitrary choice according to applications. 
For example, $p_\text{outlier}(x_n)$ can be removed from the mixture model if no outliers exist
in the target point set. 
To register point sets with outliers, we use a $D$-dimensional uniform distribution defined as 
\begin{linenomath*}
\begin{align}
  p_{\text{outlier}}(x_n)=1/S,
  \label{eq:outlier}
\end{align}
\end{linenomath*}
where $S$ is the volume of the region in which outliers can be generated. The volume $S$ is 
of course unknown and we therefore use the minimum-variance unbiased estimator of the $D$-dimensional
uniform distribution as an estimate of the volume $S$, which is estimated from the target point set $X$.

In this setting of the outlier distribution, the posterior matching probability
$p_{mn}=p(m|x_n;\bar\Theta)$, which is defined as Eq. (\ref{eq:posterior}), is reformulated as
  \begin{linenomath*}
  \begin{align}
    p(m|x_n;\bar\Theta) = \frac
     { \exp \big(-\frac{1}{2\sigma^2} ||x_n-y_m||^2 \big) }
     { c + \sum_{m'=1}^{M} \exp \big(-\frac{1}{2\sigma^2} ||x_n-y_{m'}||^2 \big)}, 
    \label{eq:posterior2}
  \end{align}
  \end{linenomath*}
where $y_m=\mathcal{T}_\text{DLD}(u_m;\bar\theta)$ and  $c=(2\pi\sigma^2)^{D/2} \frac{\omega}{1-\omega}\frac{M}{S}$.
We also note that the computation of the matching probabilities can be prohibitive if $M$ and $N$ are 
relatively large. The acceleration of this bottleneck computation will be discussed in Section 2.4.

\subsubsection*{Optimization}

Based on the theory of the EM algorithm, a solution to the point set registration problem is obtained 
by iterating the following procedure:
(i) updating the posterior probability $p_{mn}=p(m|x_n; \bar\Theta)$,
(ii) finding the ${\Theta}^*$ that minimizes the $Q$-function for $\Theta$ given the current estimate  $\bar\Theta$,
and (iii) replacing the given parameter set $\bar\Theta$ with the minimizer $\Theta^*$ of the $Q$-function.
Because the analytic solution $\Theta^*$ is unavailable, we divide the optimization procedure (ii) 
%of $\Theta=(z,\rho,\sigma^2)$ 
into three steps: 
\ifDCOL
\begin{center}
  \begin{enumerate}[label=(\alph*)]
    \item Optimization of shape parameter $z$ and translation vector $d$ given $(s,R,\sigma^2)$,
    \item Optimization of location parameter $\rho=(s,R,d)$ given $(z,\sigma^2)$,                   
    \item Optimization of residual variance $\sigma^2$ given $(z,\rho)$.                        
  \end{enumerate}
\end{center}
\else
\begin{center}
  \begin{tabular}{ll}
    (a) & Optimization of shape parameter $z$ and translation vector $d$ given $(s,R,\sigma^2)$, \vspace{2mm}\\
    (b) & Optimization of location parameter $\rho=(s,R,d)$ given $(z,\sigma^2)$,                    \vspace{2mm}\\
    (c) & Optimization of residual variance $\sigma^2$ given $(z,\rho)$.                                     \\
  \end{tabular}
\end{center}
\fi
For each step, it is possible to find the exact minimizer of the $Q$-function given corresponding fixed parameters.
For steps (a) and (c), the exact minimizers are obtained by taking partial derivatives of $Q_\text{DLD}$, 
and equating them to zero. For step (b), $Q_\text{DLD}$ must be optimized under the orthonormality 
condition $R^T R =I_D$ as $R$ is a rotation matrix. This constrained optimization problem can be analytically 
solved by using the result reported in \cite{Umeyama1991,Myronenko2009}.

The mean shape $\mb{u}$ and shape variations $\mb{H}$ %=(\mb{h}_1,\cdots,\mb{h}_K)=(H_1^T,\cdots,H_M^T)^T$ 
are updated in step (b) after updating the scale parameter $s$ and the rotation matrix $R$ because 
$\mb{u}$ and $\mb{H}$ are dependent on the coordinate system in which the statistical shape model is defined.
Translation vector $d$ is updated in both (a) and (b) of the M-step to find a better approximation of 
the exact minimizer $\Theta^*$ in a heuristic manner. The residual variance $\sigma^2$ can be estimated in 
the same manner as in the general case of the GMM-based registration, and thereby, the automatic radius control 
in the GMM is retained. %The DLD algorithm is summarized in Figure \ref{fig:algo}.

\subsubsection*{Adaptive control of the smoothness of a displacement field}

The choice of parameter $\gamma$, defined in Eq. (\ref{eq:Q}), is crucial to performing accurate registration
because $\gamma$ controls the smoothness of a displacement field.
If $\gamma$ is excessively large, the shape deformation is less affected by the target point set 
and tends to be more conservative.  
On the contrary, if $\gamma$ is excessively small, the registration often fails as
the displacement field becomes less smooth and the optimization tends to be stuck in local minima.
To relax the difficulty in choosing an appropriate $\gamma$, we adopt an optional search strategy
based on the adaptive control of $\gamma$; we initially set $\gamma$ to a moderately large value,
and if the optimization process approaches convergence, we set it to zero or a value close to zero.
The rationale of this approach is to register two point sets in a hierarchical manner; 
the average shape is deformed globally and conservatively in the early stage of the optimization, 
and is deformed more locally and flexibly near the convergence.

\subsection{\blue{Linear time algorithm}}
%\subsection{Dependent landmark drift}

%% We here show that the computational cost of the EM optimization %which requires $O(MN)$ computation based on the na\"ive algorithm
%% can be reduced to linear. %Keys to reduce computational costs

\begin{figure*}
\begin{center}
  \fbox{
    \begin{tabular}{l}  \\
      {\bf Algorithm 1}: Dependent Landmark Drift (Linear time algorithm)                                                \\ \\
      $\cdot$  Input:  $\displaystyle X=(x_1,\cdots,x_N)^{T},~ \mb{u}=(u_1^T,\cdots, u_M^T)^T$,
                       %$\displaystyle \mb{H}=(\mb{h}_1,\ldots,\mb{h}_K)^T$, $\Lambda_K$, $\omega\in(0,1)$, $\gamma>0$.   \\ \\
                       $\displaystyle \mb{H}=(H_1^T,\cdots,H_M^T)^T$, $\Lambda_K$, $\omega\in(0,1)$, $\gamma>0$.         \\ \\
      $\cdot$  Output: $Y=(y_1,\cdots,y_M)^T$.                                                                           \\ \\
      $\cdot$  Initialization:                                                                                           \\ \\
      \hspace{0.5cm}   $\displaystyle v(Y)=\mb{u}$,~ $S=\text{volume}(X)$,~
                       $\displaystyle \sigma^2 
                                        =\frac{1}{N M D}
                                         \big\{
                                              M \text{tr}(X^{T} X)
                                            - 2 (1_M^T Y X^T 1_N)
                                            + N \text{tr}(Y^T Y)
                                         \big\}$.                                                                        \\ \\
      $\cdot$  EM optimization: repeat until convergence.                                                                \\ \\
      \hspace{0.5cm} - E-step: calculate $P1_N$, $P^{T}1_M$, and $PX$ using the Nystr{\"o}m method.                      \\ \\
        \hspace{1.0cm} (a) \blue{Calculate $K_{YY}, K_{VV}^{-1}, K_{VX}$} 
                                 using the random samples $V\subset X\cup Y$ without replacement s.t.           \vspace{2mm}\\
          \hspace{1.5cm}
          $\displaystyle K_{YX}\approx K_{YV} K_{VV}^{-1} K_{VX}$,~   \\ \\
        \hspace{1.0cm} (b) Calculate $P1_N$, $P^{T}1_M$, and $PX$.    \vspace{2mm}\\
          \hspace{1.5cm}
          $\displaystyle q =1_N./(K_{YX}^T  1_M + c 1_N), ~ P^T1_M=1_N-cq,~ P1_N  =K_{YX} q,~ PX=K_{YX}\text{d}(q)X$,    \\ \\             
        \hspace{0.5cm} - M-step: update $z,s,R,d,\mb{u},\mb{H},Y$ and $\sigma^2$.                                        \\ \\
        \hspace{1.0cm} (a) Update shape parameters $z$, translation vector $d$, and shape $Y$.                  \vspace{2mm}\\
          \hspace{1.5cm}
          $\displaystyle N_P=1_M^TP1_N,~ 
                         \mb{P}=\text{d}((P1_N)\otimes 1_D),~
                         \mb{x}_P=v(PX),~ \mb{u}_P=\mb{P}\mb{u}$,                                               \vspace{2mm}\\
          \hspace{1.5cm}
          $\displaystyle H_P= N_P^{-1}(1_M^T\otimes I_D)\mb{P}\mb{H},~ 
                         x_P= N_P^{-1}  X^T P^T 1_M,~ u_P=N_P^{-1}  U^T P 1_N$,                                 \vspace{2mm}\\
          \hspace{1.5cm}
          $\displaystyle z = 
                            \big\{ \mb{H}^{T}\mb{P}\mb{H} - N_P H_P^TH_P + \gamma \Lambda_K^{-1} \big\}^{-1} 
                            \big\{ \mb{H}^{T} (\mb{x}_P-\mb{u}_P) - N_PH_P^T(x_P-u_P) \big\},             
                          $                                                                                     \vspace{2mm}\\
          \hspace{1.5cm} 
          $\displaystyle d = x_P-u_P-H_P z$,~  
          $\displaystyle v(Y) = \mb{u} + \mb{H} z  + 1_M\otimes d$, ~$y_P=N_P^{-1}Y^T P1_N$.                             \\ \\
        \hspace{1.0cm} (b) Update \blue{location} parameters $(s,R,d)$, shape model $(\mb{u},\mb{H})$, 
                           and shape $Y$.                                                                       \vspace{2mm}\\
          \hspace{1.5cm}                                                                                         
          $\displaystyle \tilde{X}=X-1_N x_P^T,~\tilde{Y}=Y-1_M y_P^T$,~                                         
          $\displaystyle A=(P\tilde{X})^{T}\tilde{Y}$, compute SVD of $A=\Phi L \Psi^T$,                        \vspace{2mm}\\
          \hspace{1.5cm}                                                                                         
          $\displaystyle R=\Phi \text{d}(1,\cdots,1,\text{det}(\Phi\Psi^T)) \Psi^T$,~                            
          $\displaystyle s=\text{tr}(A^TR) \big/ \text{tr}(\tilde{Y}^{T}\text{d}(P1_N)\tilde{Y})$,~                        
          $\displaystyle d=x_P-s R y_P$,                                                                        \vspace{2mm}\\
          \hspace{1.5cm}
          $\displaystyle \mb{u} \leftarrow s(I_M\otimes R) \mb{u}$,~
          $\displaystyle \mb{H} \leftarrow s(I_M\otimes R) \mb{H}$,~
          $\displaystyle Y \leftarrow sYR^T+1_M d^T$.                                                                    \\ \\
        \hspace{1.0cm} (c) Update residual variance ${\sigma}^2$.                                               \vspace{2mm}\\
          \hspace{1.5cm}
          $\displaystyle {\sigma}^2
                   = \frac{1}{N_PD} 
                        \Big\{   \text{tr}(X^T \text{d}(P^T1_M)X)
                              -2 \text{tr}(Y^TPX)
                              +  \text{tr}(Y^T \text{d}(P1_N)Y) \Big
                            \}$.                                                                                         \\ \\
    \end{tabular}
  }
  \caption{
    Linear time algorithm for registering mean shape $\mb{u}$ and point set $X$.
    The notation $\text{d}(\cdot)$ denotes the operation for converting a vector into a diagonal matrix, 
    $\text{det}(\cdot)$ denotes the determinant of a square matrix, and
    $v(\cdot)$ denotes the operation for converting a matrix into a vector in the row-major order.
    The mean shape $\mb{u}$ is also denoted by the matrix notation $U=(u_1,\cdots,u_M)^T$, i.e., $\mb{u}=v(U)$.
    Furthermore, $1_M$ and $1_N$ denote column vectors of all 1s of size $M$ and $N$, respectively. 
    The identity matrices of size $M$ and $D$ are denoted by $I_M$ and $I_D$. 
    The symbols ``$\leftarrow$'' and $\otimes$ denote the substitution and the Kronecker product, respectively. 
  }
  \label{fig:linalgo}
\end{center}
\end{figure*}

\blue{
We show that the computational cost of one step of the EM optimization described in the previous subsection
can be reduced to linear, i.e., $O(M+N)$. The bottleneck of the EM optimization is the computation of matching 
probabilities $p_{mn}$ for all $m=1,\cdots,M$ and $n=1,\cdots,N$; the na\"{i}ve computation requires $O(MN)$ computation.
The key to reducing the computational cost is to use the Nystr\"om method \cite{Williams2001}, 
%which is a computationally efficient technique for computing a low-rank approximation of a Gram matrix.
which avoids the direct evaluation of all the matching probabilities.
%% When computing matching probabilities, the Nystr\"om method is typically faster than the improved fast Gauss 
%% transform \cite{Yang2003}, a more conventional method for reducing the computational cost of matching probabilities.
The resulting linear time algorithm that we propose is shown in Figure \ref{fig:linalgo}.
}

\subsubsection*{\blue{Key observations for reducing the computational cost}}

\blue{
Here, we summarize the idea for reducing the computational cost of the EM optimization.
Suppose $K_{YX}\in \mathbb{R}^{M\times N}$ is a Gaussian affinity matrix, 
the $mn$th element of which is defined as $k_{mn}=\exp\{-||x_n-y_m||^2/2\sigma^2\}$. 
Key observations for reducing the computational cost are listed as follows:
\begin{itemize}
  \item
    All computations related to matching probabilities $P=(p_{mn})$ in the M-step of the EM algorithm can
    be expressed as one of the products $P1_N$, $P^{T}1_M$, and $PX$. 
  \item
    The computations of $P1_N$, $P^{T}1_M$, and $PX$ can be replaced by
    the products related to $K_{XY}$.
  \item
    The matrix-vector products related to $K_{YX}$ can be calculated in $O(M+N)$ time
    by using the Nystr\"om method.
  \item
    All matrices of size $M\times M$ and $MD\times MD$, e.g., $\text{d}(q)$, $\text{d}(P1_N)$, 
    $\text{d}(P^T 1_M)$, $\mb{P}$ and $I_M$, are diagonal, and their products can be computed in $O(M)$ time. 
\end{itemize}
The first observation is obtained by solving the optimization subproblems for the M-step
and factoring out $P1_N$, $P^{T}1_M$, and $PX$ from each term of their solutions, as shown in the Algorithm 1.
The fourth observation is easily obtained by avoiding the evaluation of off-diagonal elements
in the diagonal matrices. The number of nonzero elements of each diagonal matrix is at most $M$,
and thereby, the computational cost of evaluating the products related to the diagonal matrices is $O(M)$.
We next see how the second and third observations are obtained.
}

\subsubsection*{\blue{Computation via the Gaussian affinity matrix}}

\blue{
The second observation is guaranteed by \cite{Myronenko2010}. In the same manner as the CPD,
the products $P1_N$, $P^T1_M$, and $PX$ can be replaced by the matrix-vector products related to 
the Gaussian affinity matrix $K_{YX}$
as the definitions of the products $P1_N$, $P^T1_M$, and $PX$ are exactly the same as those of the CPD
except for the definition of the outlier distribution. Suppose $q\in\mathbb{R}^N$ is the vector defined as
%% Suppose that $q\in\mathbb{R}^N$ is the vector, $n$th element of which is the denominator defined in Eq. (??).
%% By using $K_{XY}$, the vector $q$ is represeted as
\begin{align}
q=1_N./(K_{YX}^T1_M + c 1_N) \nonumber,
\end{align}
where ``$./$'' denotes element-wise division and 
$c=(2\pi\sigma^2)^{D/2}$ $\frac{\omega}{1-\omega}\frac{M}{S}$ is a constant. 
%% Considering (1) $cq$ is the vector, $n$th element of which is the posterior probability that $x_n$ is an outlier,
%% (2) the $n$th element of $P^T1_M$
%% Then, the poste
%Suppose $\text{d}(\cdot)$ denotes the operation for converting a vector into a diagonal matrix.
Then, the matrix-vector products $P1_N$, $P^T1_M$, and $PX$ can be reformulated by using $K_{YX}$ and $q$ as follows:
\ifDCOL
  \begin{align}
    \hspace{1cm}
    \begin{tabular}{cl}
      $\displaystyle P^T1_M$ \hspace{-5mm} &$\displaystyle =1_N-cq$,                     \vspace{1mm}\\
      $\displaystyle P1_N  $ \hspace{-5mm} &$\displaystyle =K_{YX} q$,                   \vspace{1mm}\\
      $\displaystyle PX    $ \hspace{-5mm} &$\displaystyle =K_{YX}\text{d}(q)X$,
    \end{tabular}
    \label{eq:bottleneck}
  \end{align}
\else
  \begin{linenomath*}
  \begin{align}
    P^T1_M=1_N-cq,             ~~  
    P1_N  =K_{YX} q,           ~~
    PX    =K_{YX}\text{d}(q)X,
    \label{eq:bottleneck}
  \end{align}
  \end{linenomath*}
\fi
where $\text{d}(\cdot)$ denotes the operation for converting a vector into a diagonal matrix.
That is, all the products in the bottleneck computations $P1_N$, $P^T 1_M$, and $PX$ can be
computed through the products related to $K_{XY}$.
We note that the matrix-matrix products $K_{YX}\text{d}(q)X$ can be expressed as the sum of the 
matrix-vector products related to $K_{YX}$ as follows:
\begin{linenomath*}
\begin{align}
  K_{YX}\text{d}(q)X= \sum_{d=1}^D K_{YX} (q \circ x_{(d)}), \nonumber
\end{align}
\end{linenomath*}
where ``$\circ$'' denotes element-wise product and $x_{(d)}$ is the $d$th column of $X$.
Therefore, if the matrix-vector products $K_{YX}^T a$ and $K_{YX} b$ for arbitrary vectors
$a\in \mathbb{R}^M$ and  $b\in \mathbb{R}^N$ are computed in $O(M+N)$ time, 
all the bottleneck computations $P1_N$, $P^T1_M$, and $PX$ can also be computed in $O(M+N)$ time.
}

%% $D$-time repetition of the matrix-vector
%% products $K_{YX} b$ since $\text{d}(q) X$ is a matrix of size $M \times D$.

%% We note that the evaluation of $K_{YX}\text{d}(q)X$,                  
%% The direct evaluation of the diagonal matrix $\text{d}(q)$ of size $M \times M$ can be a bottleneck.

\subsubsection*{\blue{Nyst\"om approximation for reducing the computational cost}}

\blue{
We then proceed to the third observation.
The matrix-vector products $K_{YX}^T a$ and $K_{YX} b$ can be approximated in $O(M+N)$ time by using
the Nysr\"om method \cite{Williams2001}, which is a technique of generating a low-rank
approximation of a Gram matrix without its eigendecomposition in a computationally efficient manner.
%which is composed of all-by-all dot products among samples in a reproducing kernel Hilbert space,
%in a computationally efficient manner. 
The Gaussian affinity matrix for all points in the combined point set $X\cup Y$ is a Gram matrix, 
and thereby, the Nystr\"om method is applicable to the approximation of its submatrix $K_{YX}$. 
}
Suppose $V$ is the set of points that is randomly chosen $L$ times from the combined point set $X\cup Y$ 
without replacement. Based on the theory of the Nystr\"om method, the Gaussian affinity matrix $K_{YX}$ 
can be approximated as follows:
\begin{linenomath*}
\begin{align}
   K_{YX} \approx K_{YV} K_{VV}^{-1} K_{VX},
   \label{eq:nystrom}
\end{align}
\end{linenomath*}
where 
$K_{YV}\in \mathbb{R}^{M\times L}$, $K_{VV}\in \mathbb{R}^{L\times L}$, and $K_{VX}\in \mathbb{R}^{L\times N}$
are the Gaussian affinity matrices between $Y$ and $V$, for the subset $V$ itself, and between $V$ and $X$, respectively.
Therefore, if the number of sampled points $L$ is much less than $N$ and $M$,
the approximation of the products $K_{YX}^T a$ or $K_{YX} b$ can be efficiently calculated
by evaluating matrix-vector products on the right-hand side of the following equations
\begin{linenomath*}
\begin{align}
   K_{YX}^T a & \approx  K_{VX}^T K_{VV}^{-1} K_{YV}^T a, \nonumber \\\
   K_{YX}   b & \approx  K_{YV}   K_{VV}^{-1} K_{VX}   b,
\end{align}
\end{linenomath*}
from right to left. The computational costs of evaluating the right-hand side equations are $O(DL(M+N)+L^3+DL^2)$;
i.e., $O(M+N)$ if we assume that $L$ and $D$ are constants.
One issue of using the Nystr{\"o}m method is the trade-off between speed and accuracy;
setting $L$ to a small value leads to fast but less accurate computations.
%
%% During the early stage of the optimization, the accuracy of approximating the Gaussian functions is less important 
%% because the residual variance $\sigma^2$ is relatively large and the matching probabilities,
%% defined as Eq (?), are roughly the same, regardless of the accuracy of the approximation.
%% On the contrary, during the late stage of the optimization,
%% the accuracy of the approximation is important because the matching probabilities are strongly affected by it.
%
%
%% Therefore, we employ a different strategy for reducing the computational cost of evaluating $K_{YX}^T a$ and $K_{YX} b$
%% that is more accurate and computationally efficient if the residual variance $\sigma^2$ is sufficiently small.
Therefore, if the residual variance $\sigma$ is sufficiently small, we directly evaluate Eqs. (\ref{eq:bottleneck}),
reducing the computational cost by ignoring the points that are sufficiently distant from the mean point of
a Gaussian function. The points that lie within a small distance from the mean point can be efficiently 
found by using the KD tree search.

%\subsubsection*{Alternative approach to reducing the computational cost}

%% The matrix-vector products can also be accelerated by the use of a more conventional technique 
%% such as the improved fast Gauss transform \cite{Yang2003}
%% which reduces the computational cost of the summation of the Gaussian functions.

%% An advantage of the IFGT over the Nysr\"om method is the accuracy of the approximation, which is guaranteed by 
%% specifying the error bound.  On the contrary, the computational cost of the IFGT

\vspace{3mm}
\subsubsection*{\blue{Total computational cost of the EM algorithm}}

The computations of the remaining terms in Algorithm 1 can be evaluated in $O(M+N)$ time.
The product $P\tilde{X}$ in Algorithm 1 is computed in $O(M+N)$ time because $P\tilde{X}=PX-(P1_N)x_P^T$.
Furthermore, the initialization step requires $O(M+N)$ computations because
%$\sum_{n=1}^N \sum_{m=1}^M x_n^Ty_m =1_N^TXY^T 1_M$ 
$1_N^TXY^T 1_M$ 
can be computed by evaluating matrix-vector products of the right-hand side equation from left to right. 
Therefore, the total computational cost of the initialization and one iteration of the EM algorithm is
proportional to $M+N$.
 
We also note that the Nysr\"om method is typically faster than the improved fast Gauss transform \cite{Yang2003},
a more conventional technique for reducing the computational cost of evaluating the matching probabilities.
The IFGT is the method for accelerating the sum of Gaussian functions by using the truncated Taylor series expansion.
An issue with the IFGT is that its computational cost depends on the radius of the Gaussian functions; 
the modified IFGT algorithms that are practically faster than the original version require direct computation,
i.e., $O(MN)$ computation, in a worst-case scenario \cite{Morariu2009,Raykar2005}.
If we use the IFGT as an alternative method for speeding up the E-step, direct computation cannot be avoided 
because the radius, which corresponds to residual variance $\sigma^2$ in our method,
gradually decreases as the optimization proceeds. On the contrary,
the computational cost of the Nystr{\"o}m method is unaffected by the radius of the Gaussian function,
and is always $O(M+N)$ during the optimization. 

\section{Experiments}

\ifDCOL
\else
  %%version 3 (document)
\ifDCOL
  \begin{teaserfigure}
     \centering \vspace{-1mm} \hspace{0mm}
       \\ \vspace{-8mm}
       \includegraphics[width=0.160\textwidth]{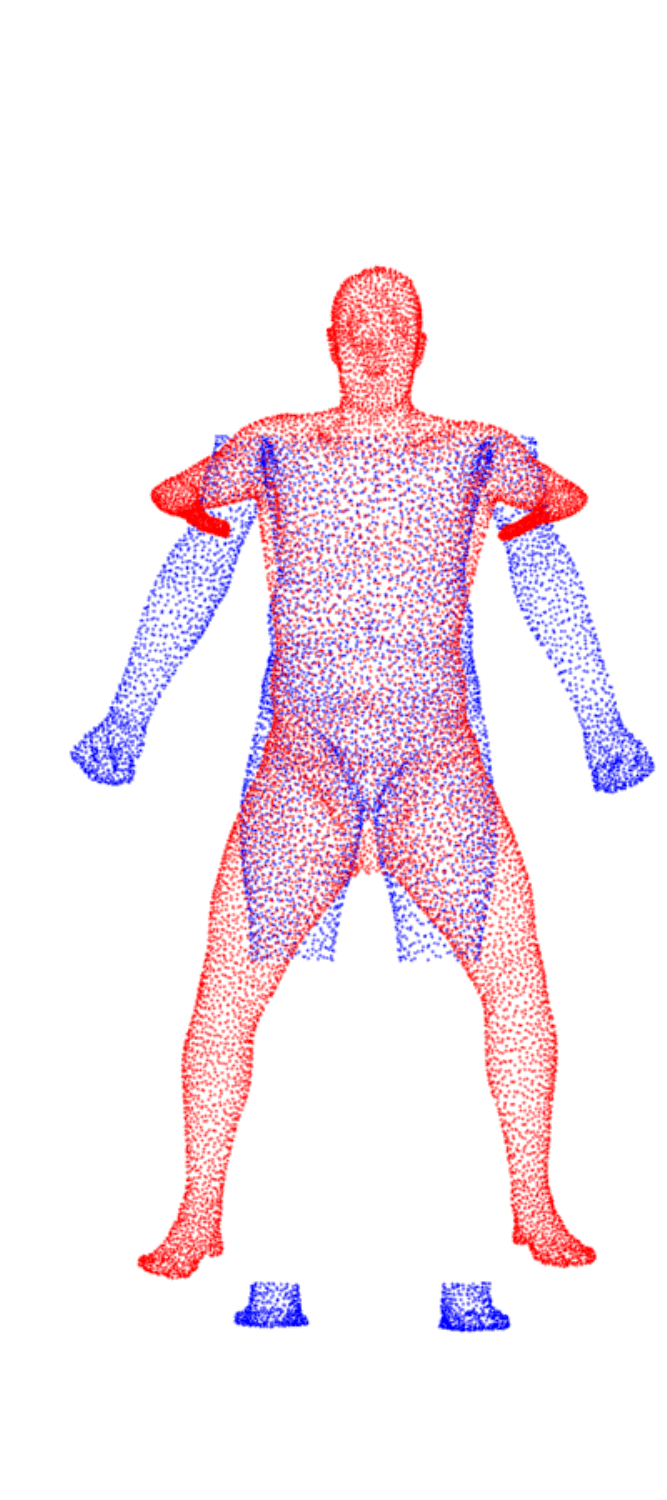}
       \includegraphics[width=0.160\textwidth]{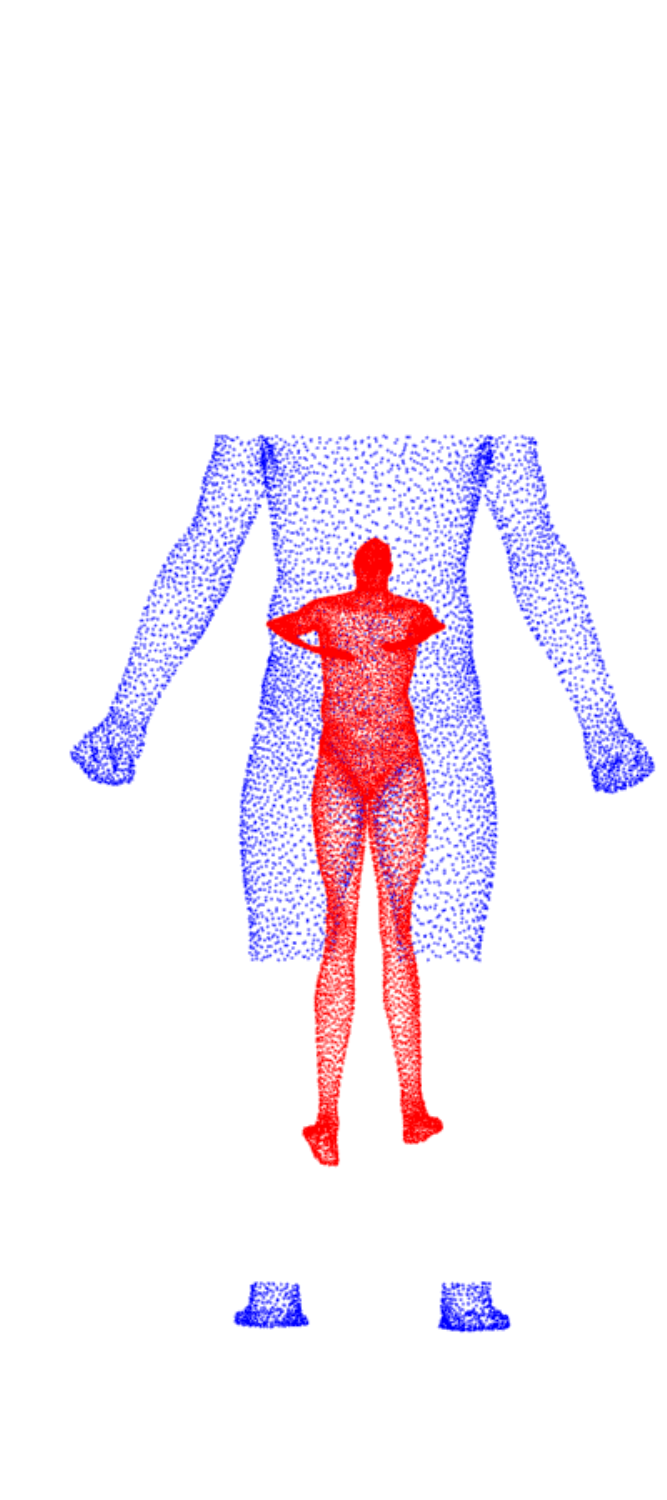}
       \includegraphics[width=0.160\textwidth]{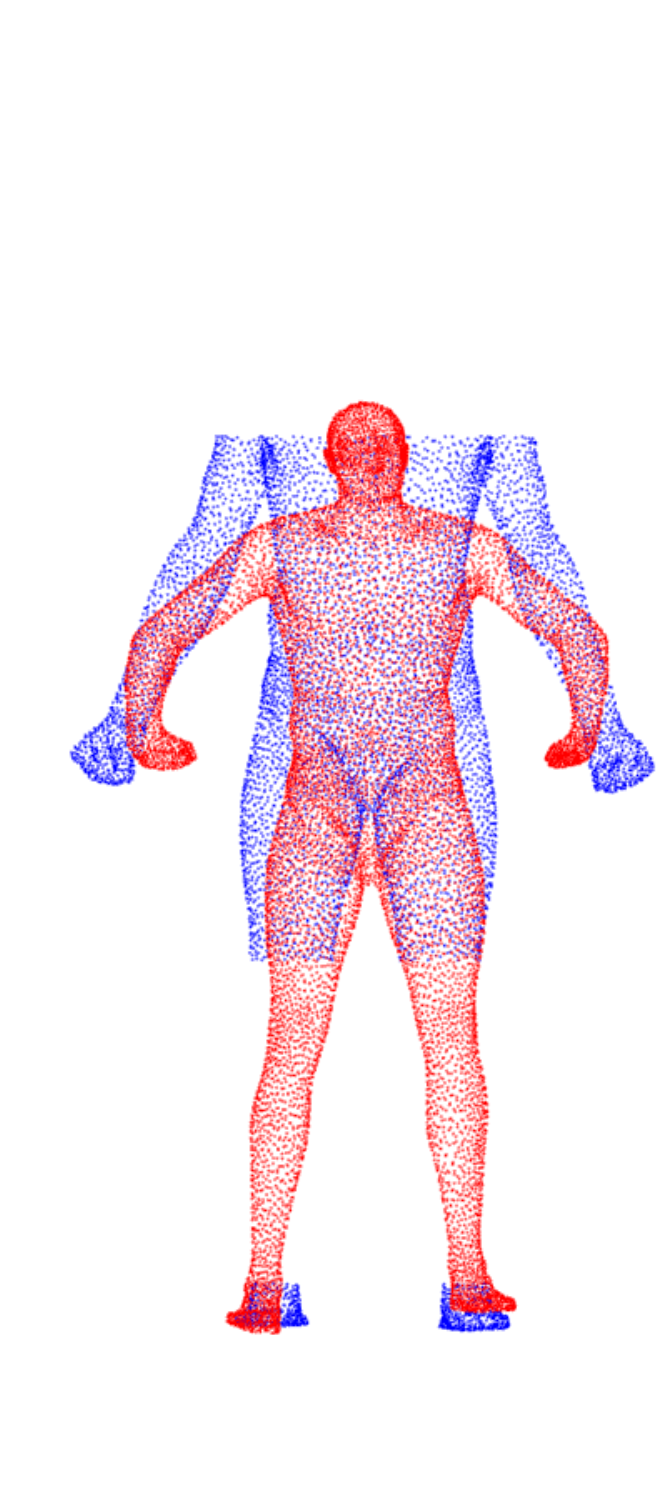}
       \includegraphics[width=0.160\textwidth]{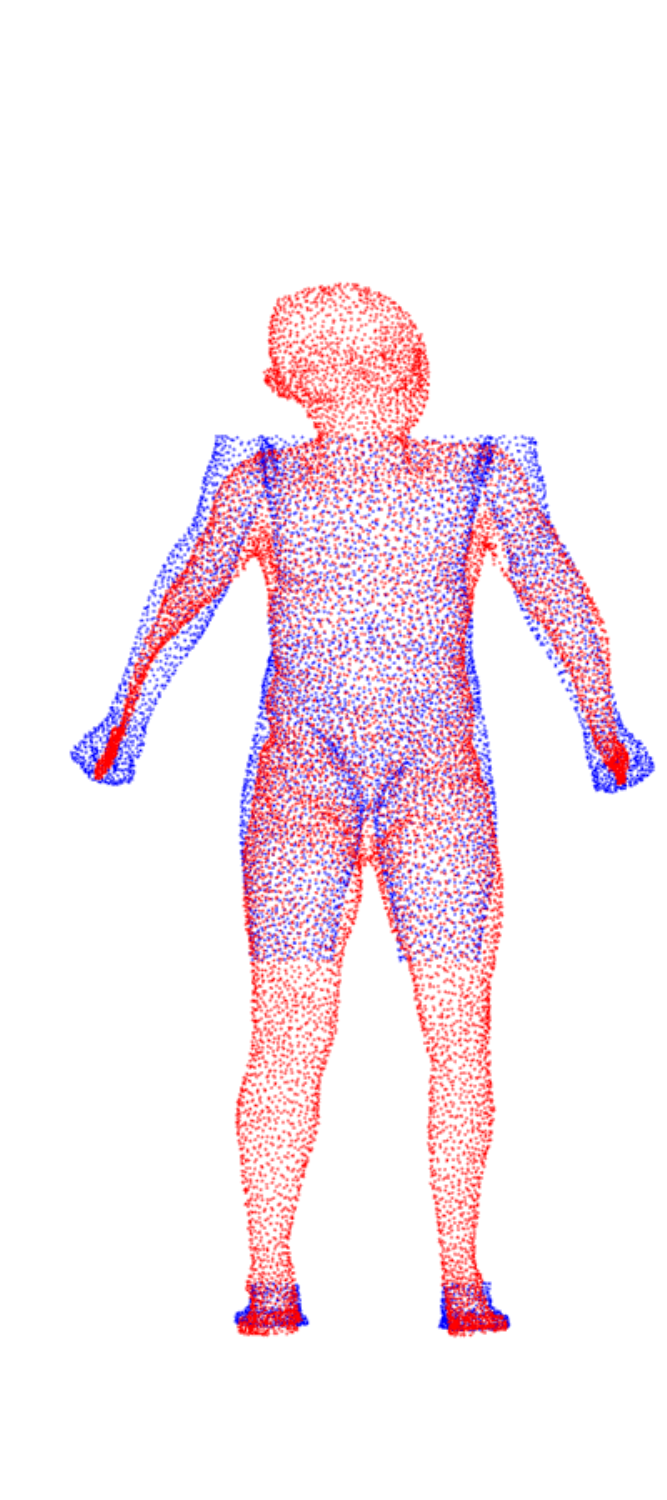}
       \includegraphics[width=0.160\textwidth]{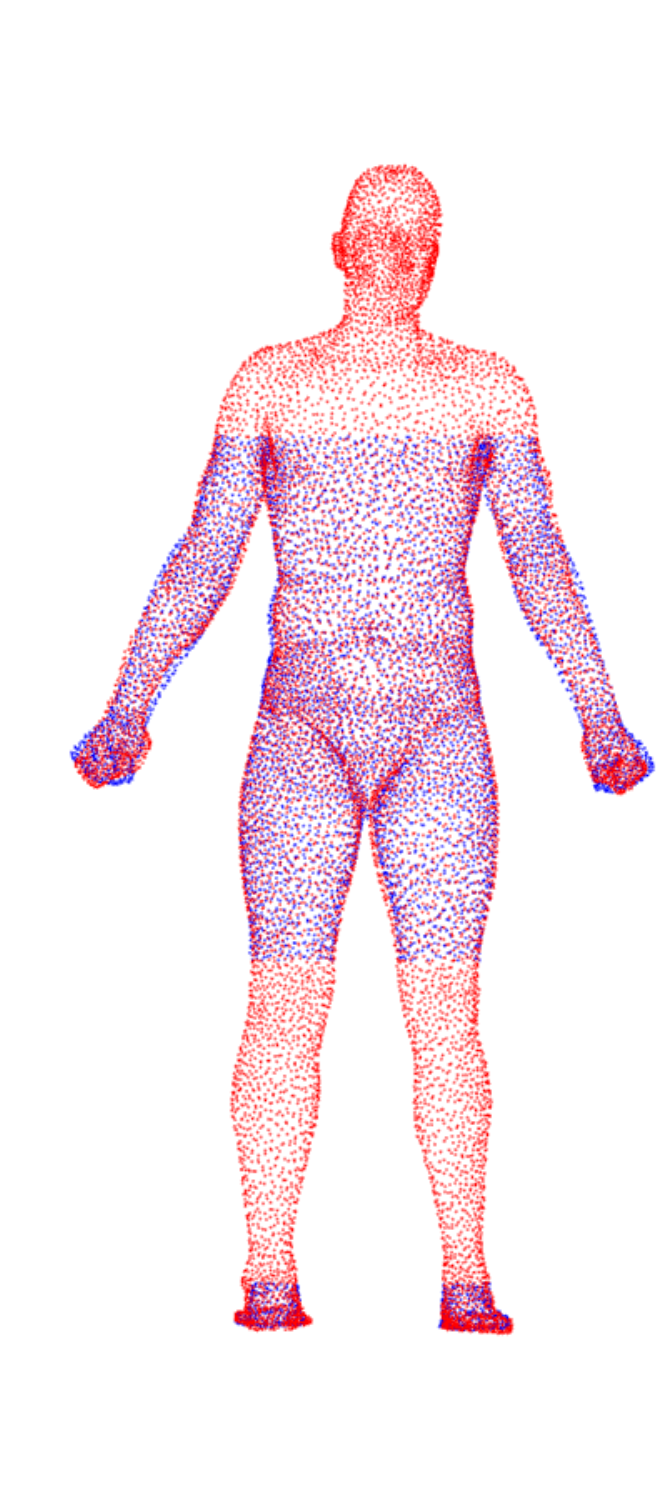} 
       \\ \vspace{-5mm} \hspace{2mm}
       \includegraphics[width=0.165\textwidth]{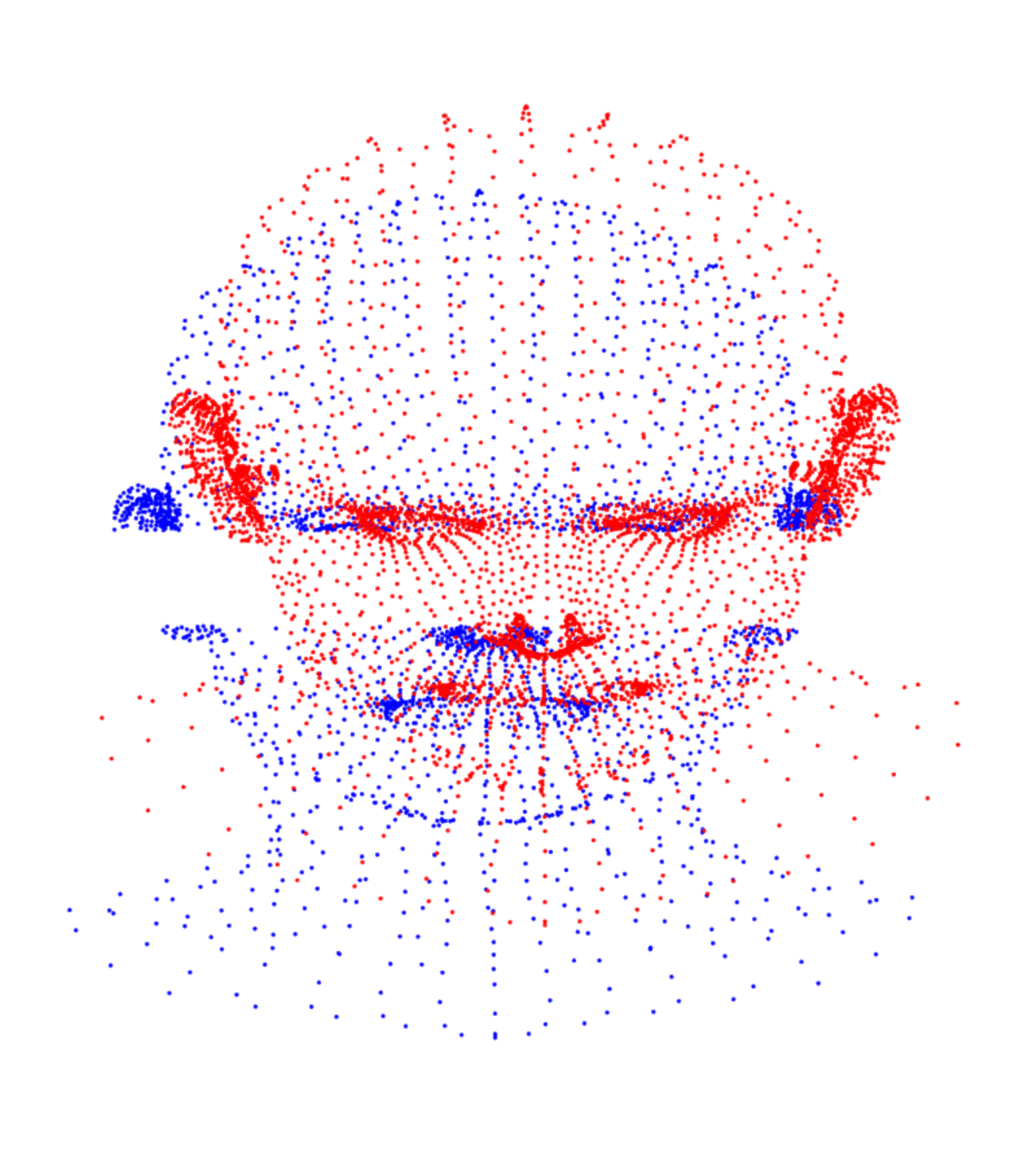}
       \includegraphics[width=0.165\textwidth]{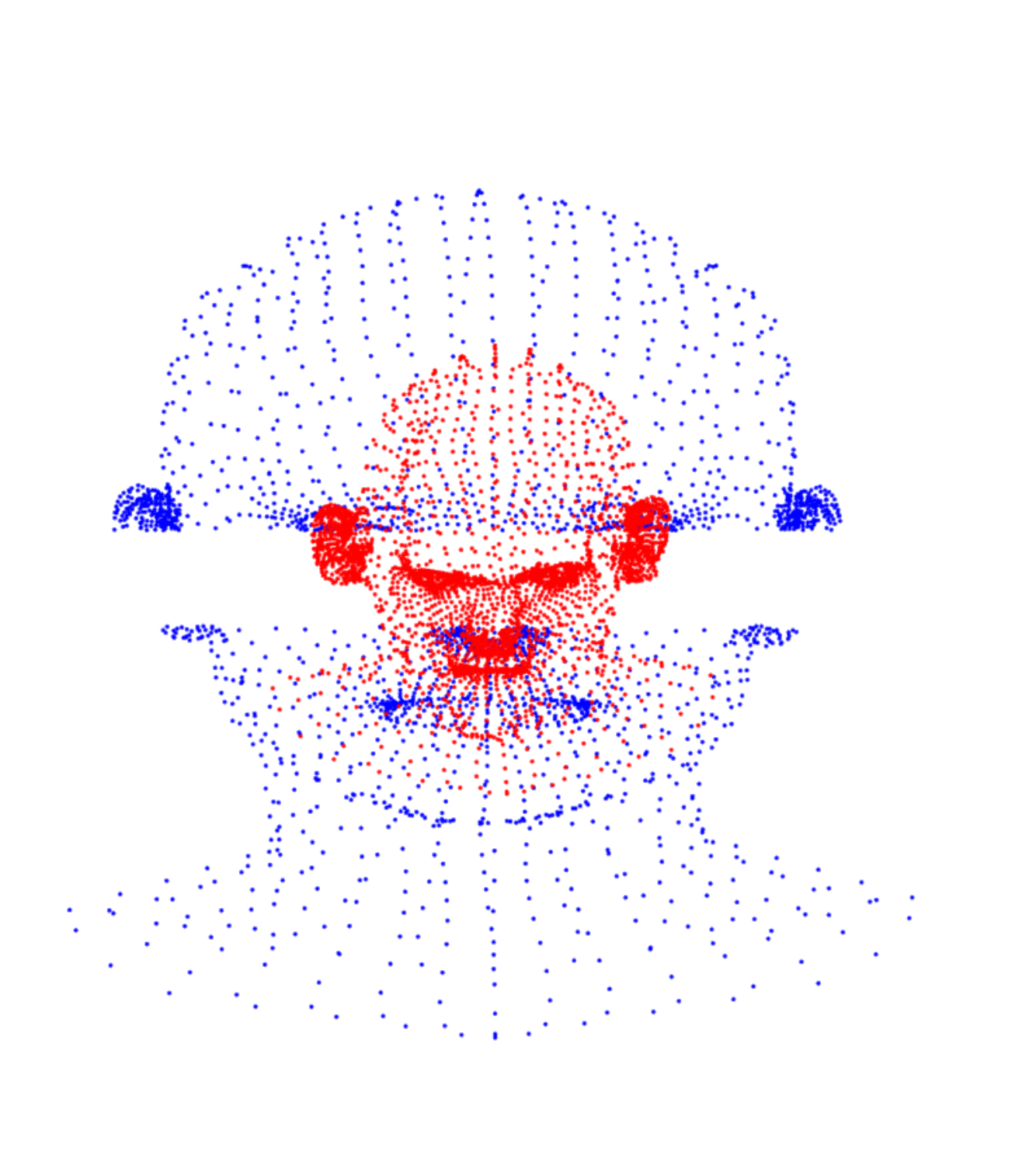}
       \includegraphics[width=0.165\textwidth]{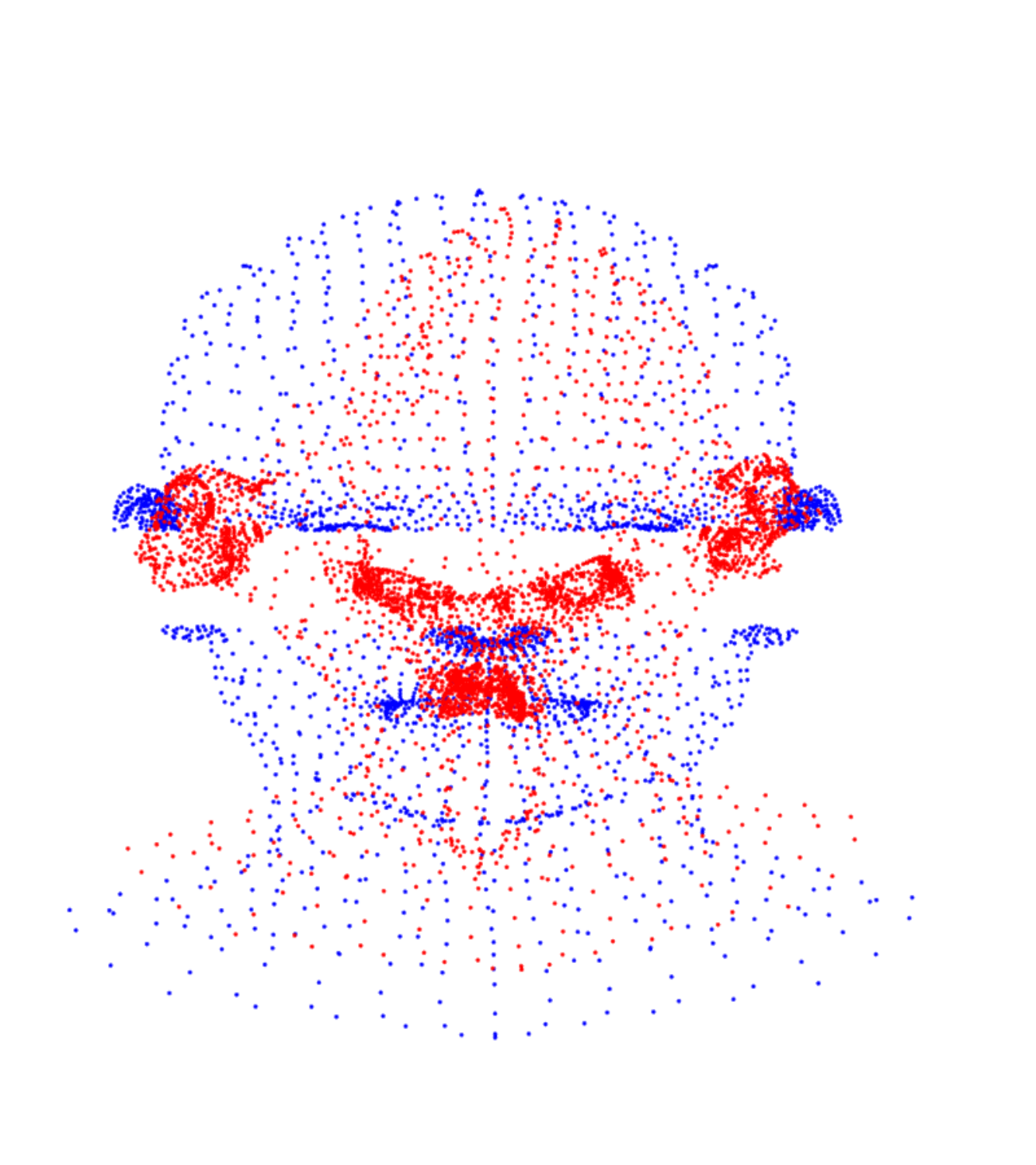}
       \includegraphics[width=0.165\textwidth]{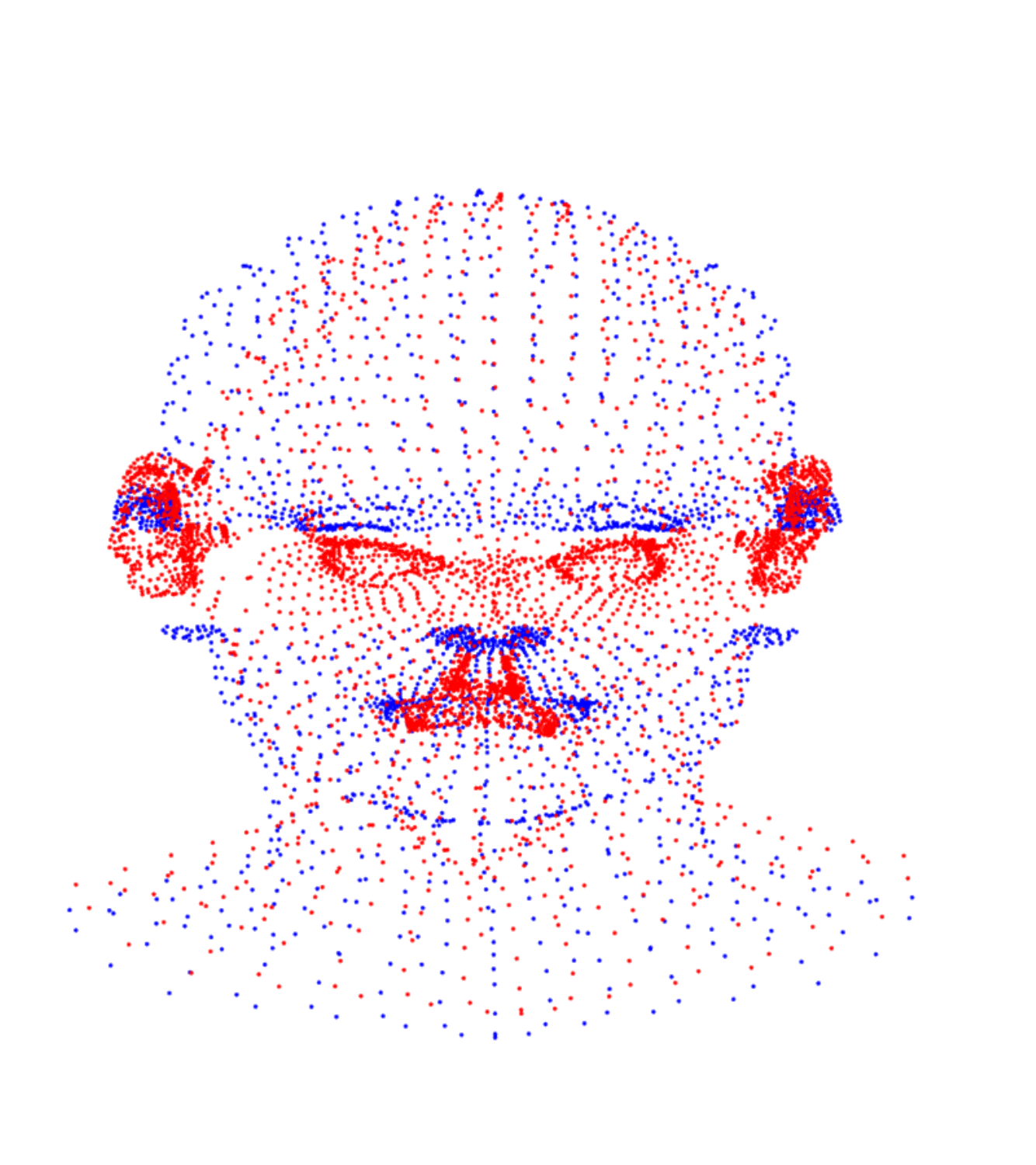}
       \includegraphics[width=0.165\textwidth]{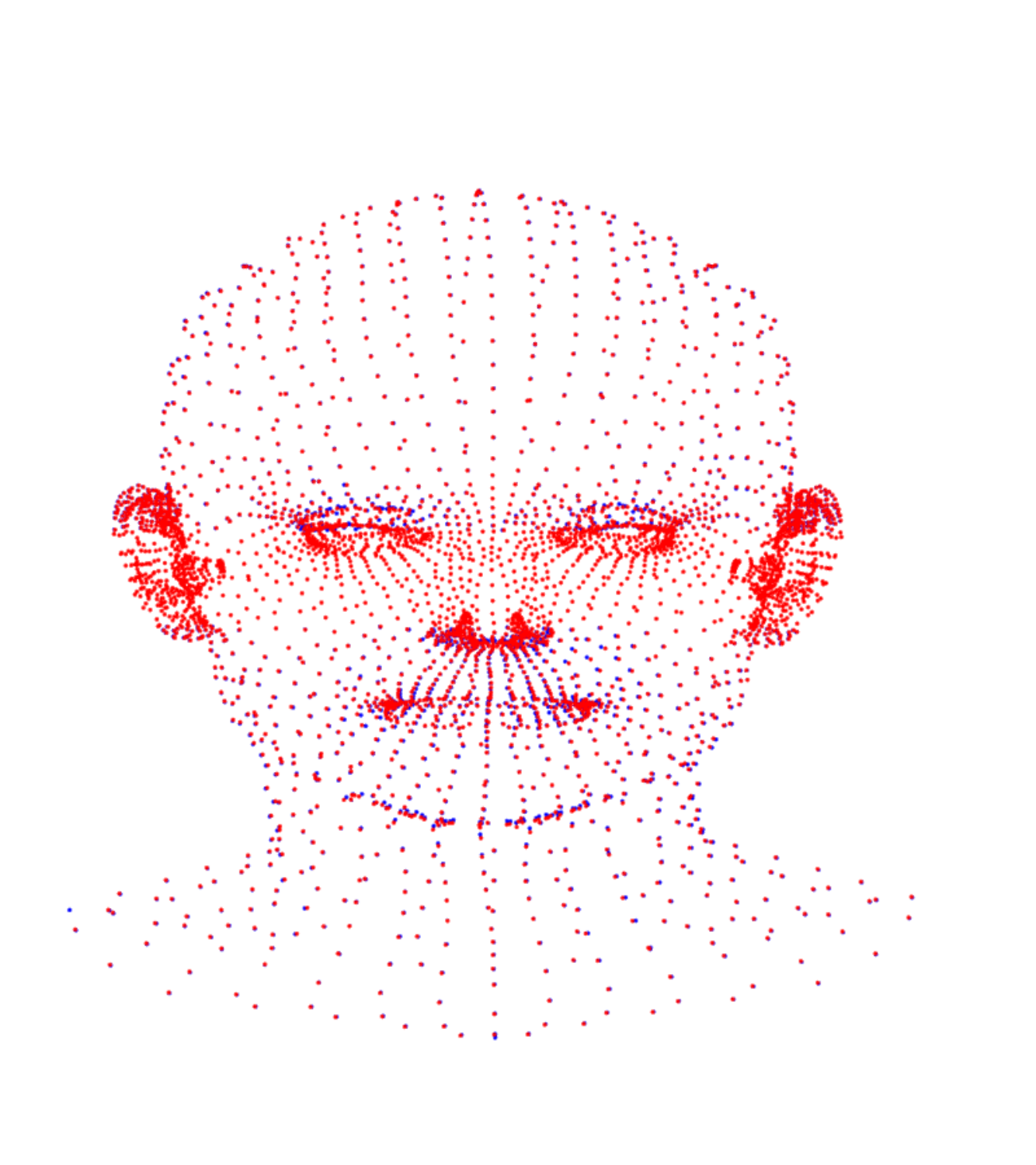} 
      \vspace{-2mm}
      \caption{
        Optimization trajectories for a body shape (top) and a face (bottom) with missing regions.
        The points in the target shapes and deformed shapes are colored blue and red,
        respectively. The leftmost column shows the initial point sets, and
        the optimization proceeds from left to right.
      }
      \label{fig:teaser}
  \end{teaserfigure}
\else
  \begin{figure}
     \centering \vspace{-1mm} \hspace{0mm}
       \\ \vspace{-8mm}
       \includegraphics[width=0.160\textwidth]{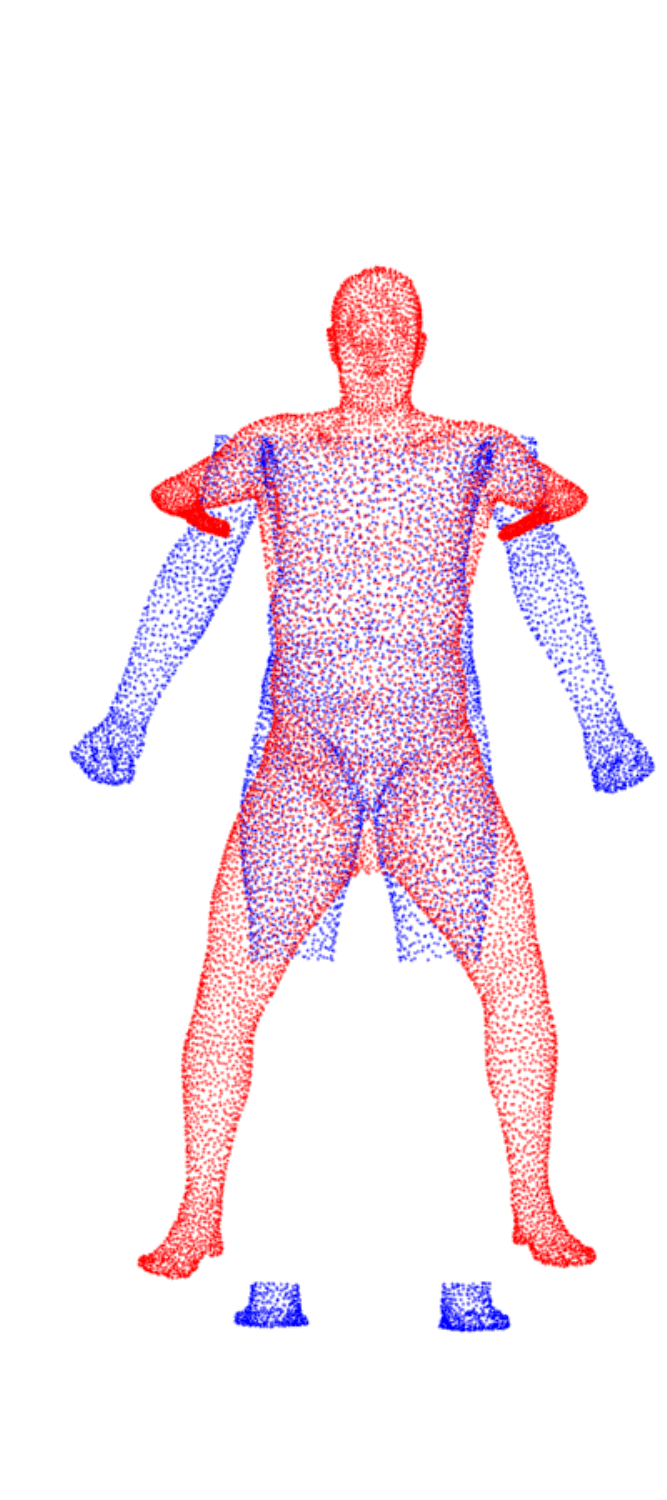}
       \includegraphics[width=0.160\textwidth]{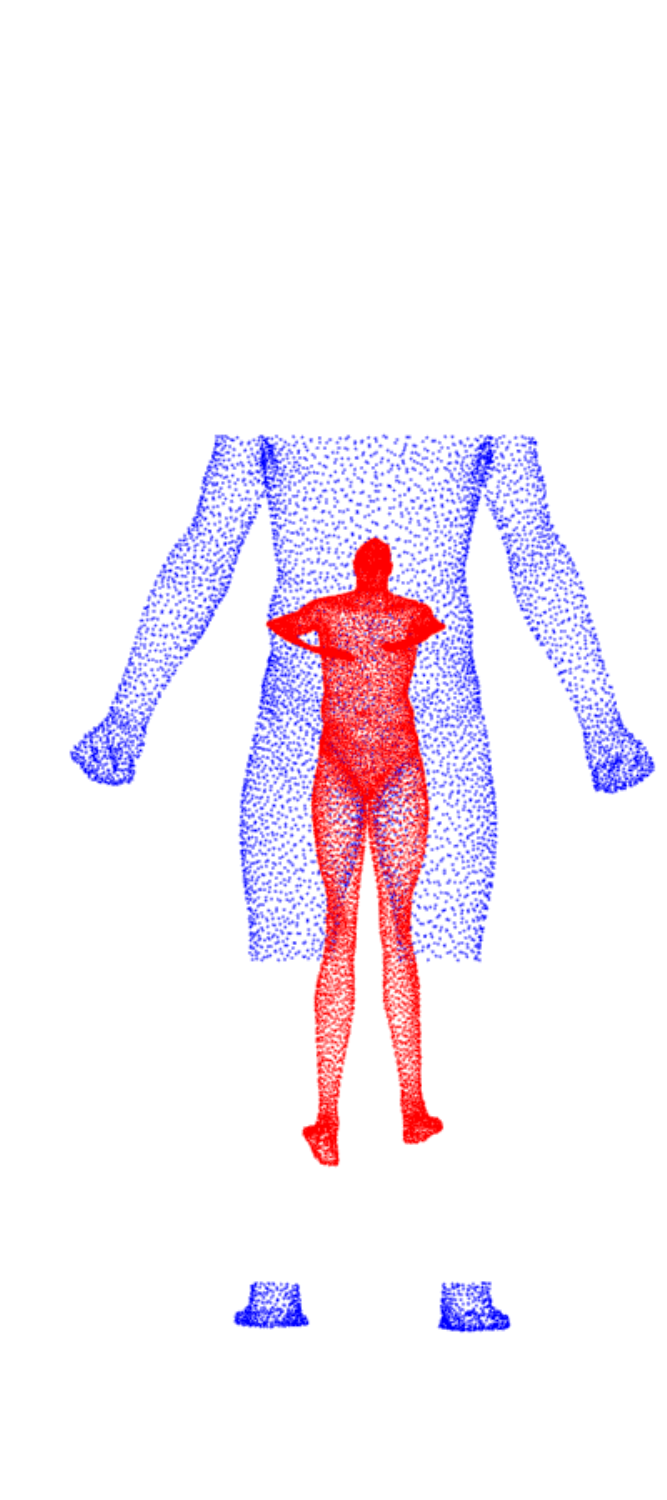}
       \includegraphics[width=0.160\textwidth]{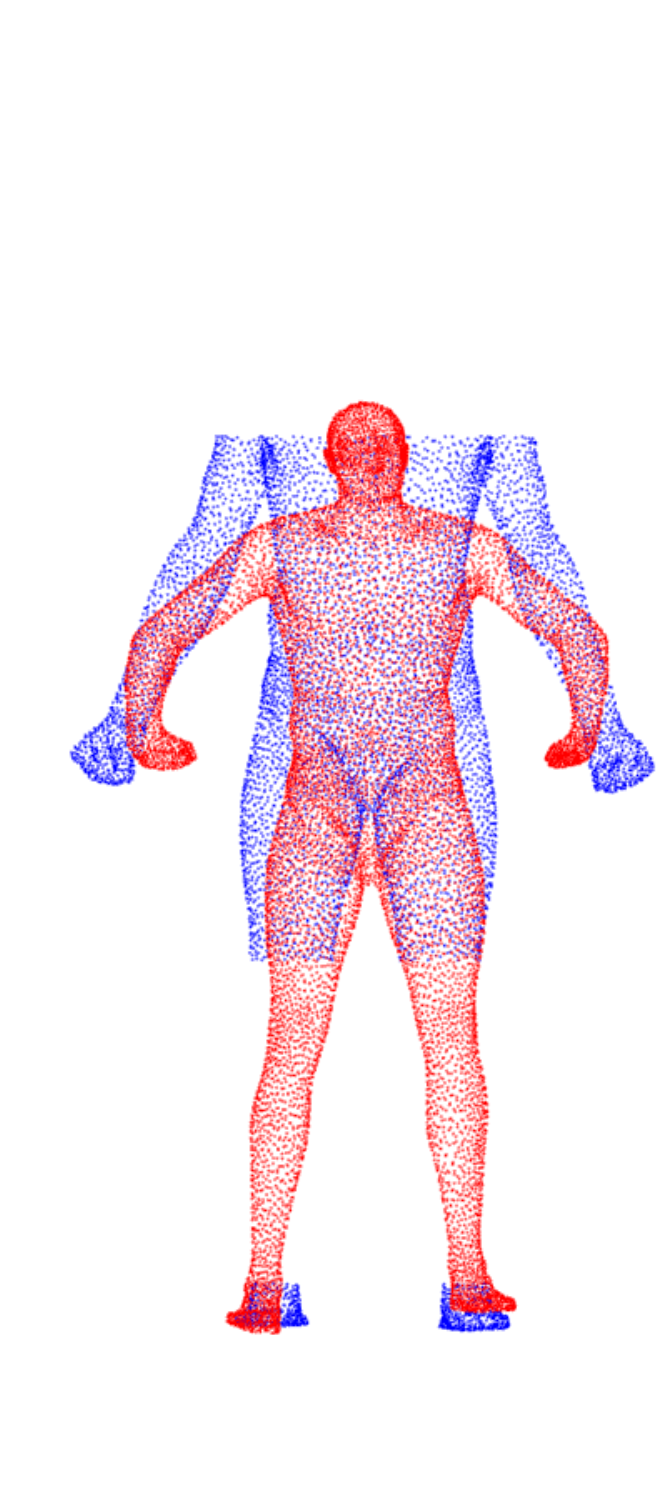}
       \includegraphics[width=0.160\textwidth]{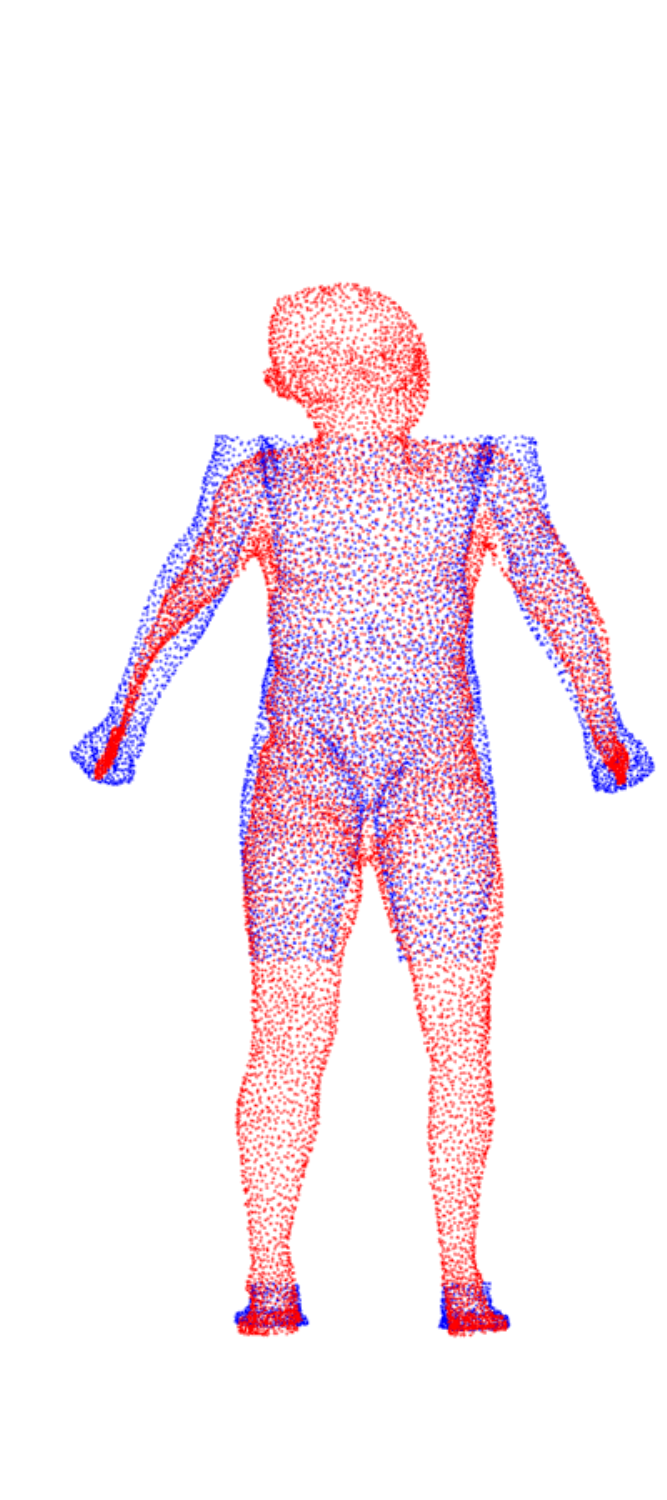}
       \includegraphics[width=0.160\textwidth]{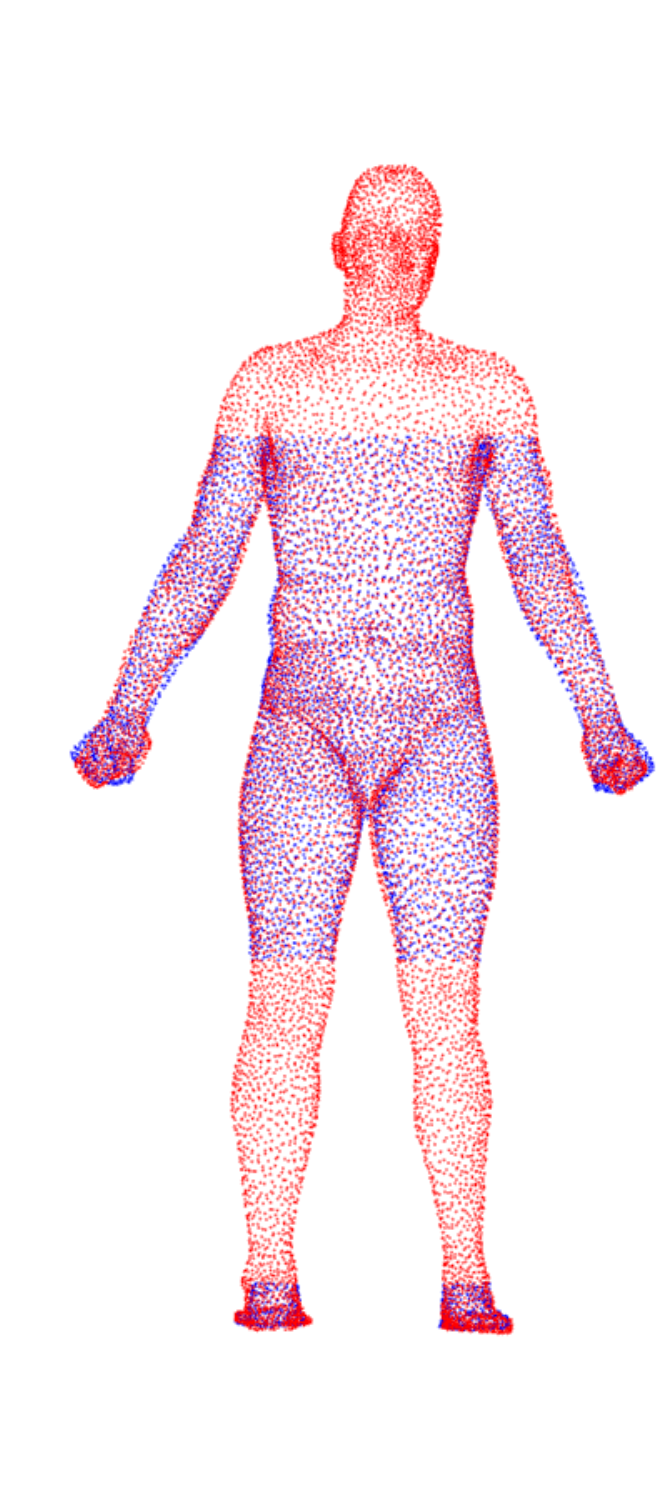} 
       \\ \vspace{-5mm} \hspace{2mm}
       \includegraphics[width=0.165\textwidth]{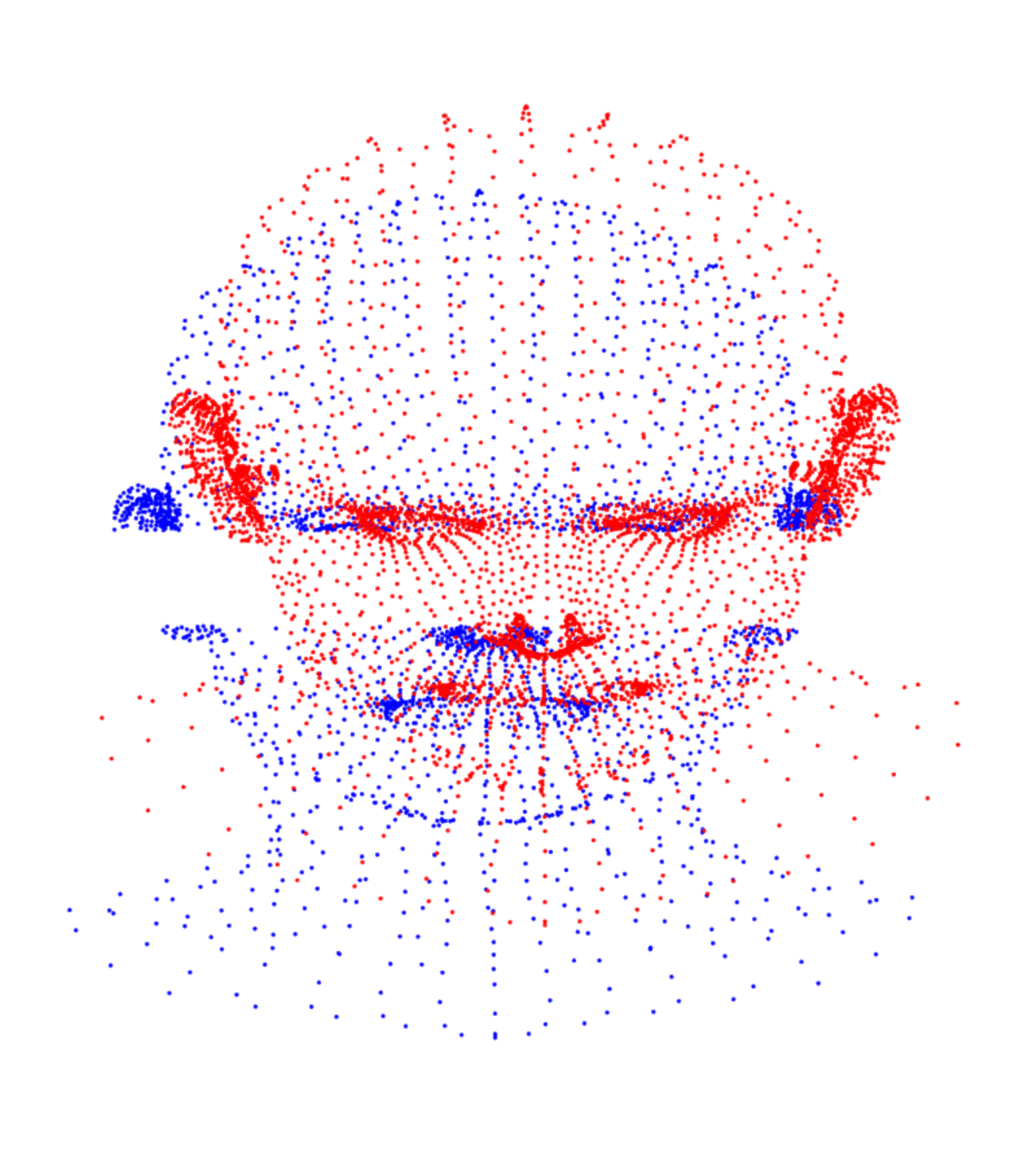}
       \includegraphics[width=0.165\textwidth]{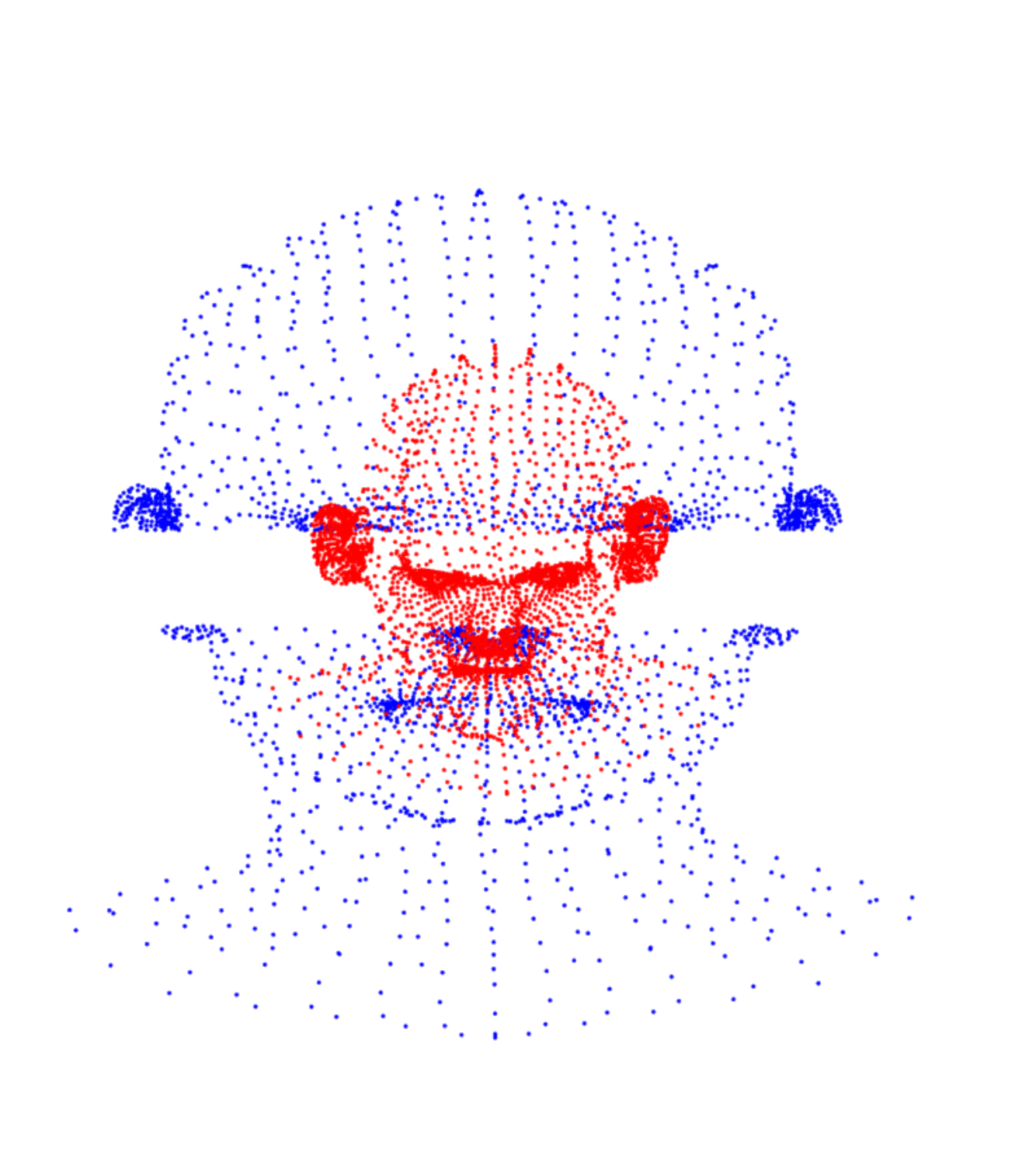}
       \includegraphics[width=0.165\textwidth]{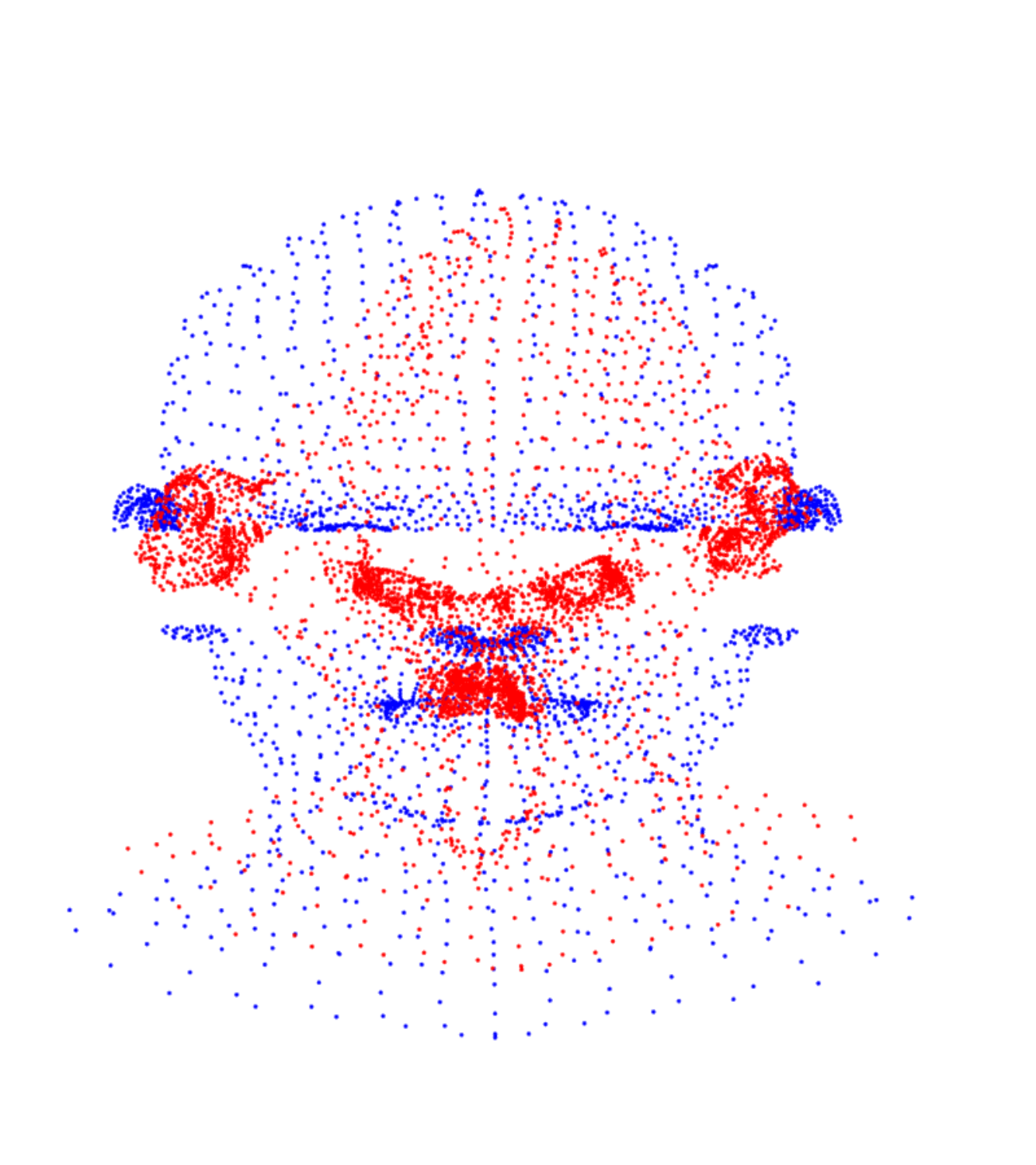}
       \includegraphics[width=0.165\textwidth]{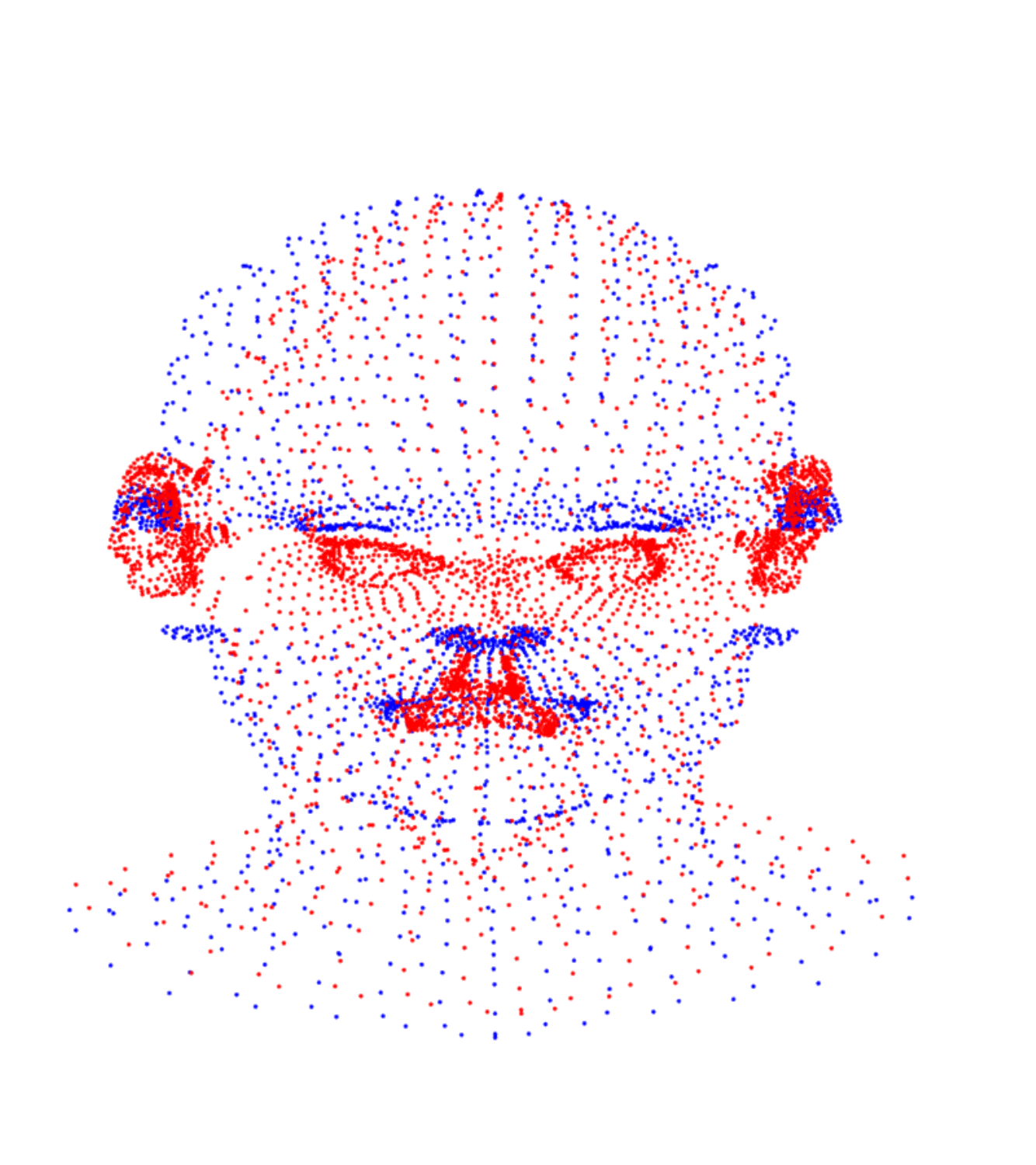}
       \includegraphics[width=0.165\textwidth]{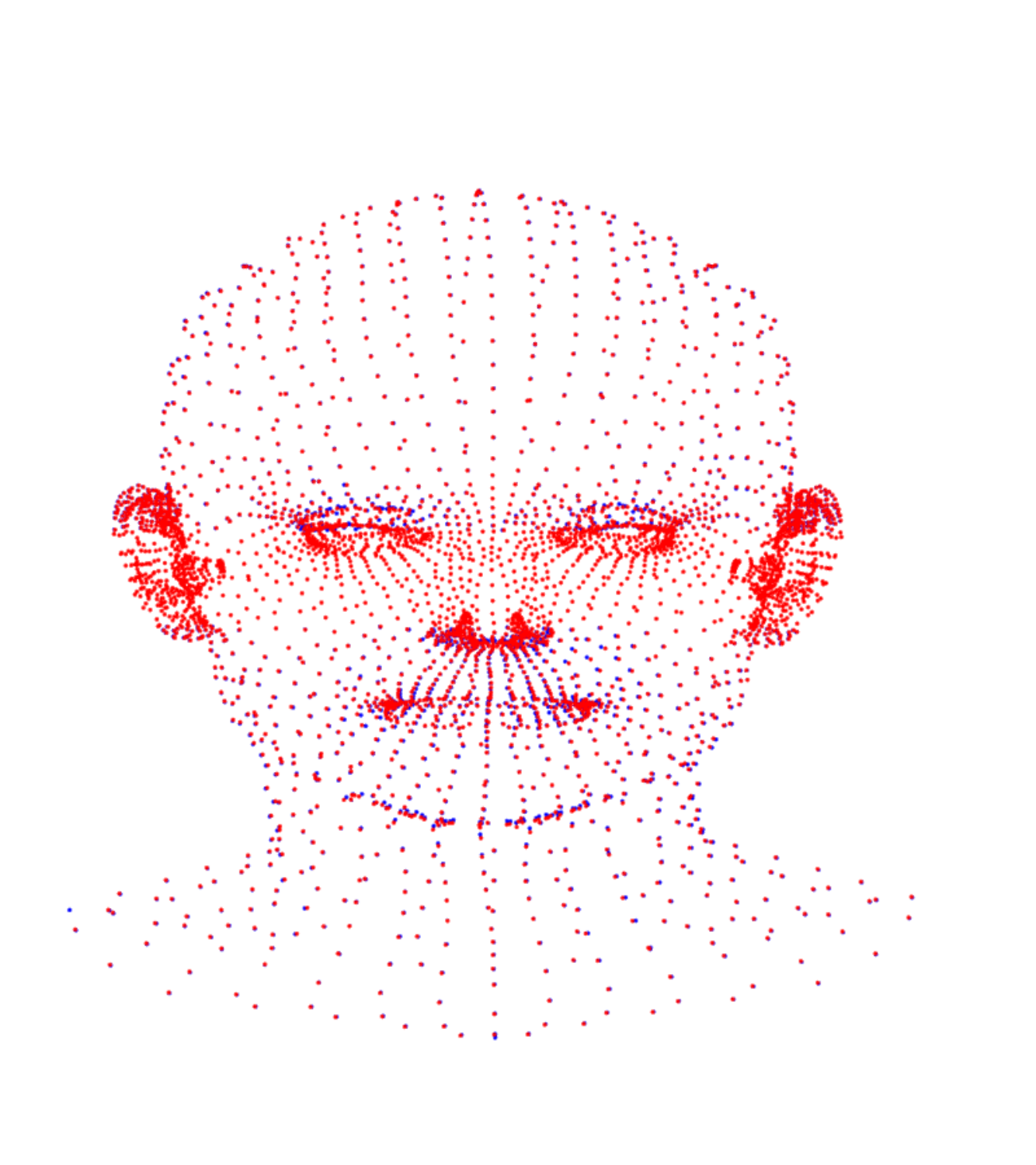} 
      \vspace{-2mm}
      \caption{
        Optimization trajectories for a body shape (top) and a face (bottom) with missing regions.
        The points in the target shapes and deformed shapes are colored blue and red,
        respectively. The leftmost column shows the initial point sets, and
        the optimization proceeds from left to right.
      }
      \label{fig:teaser}
  \end{figure}
\fi

\fi
\begin{figure*}
   \centering
     \hspace{-0.0cm}\includegraphics[width=0.15\textwidth]{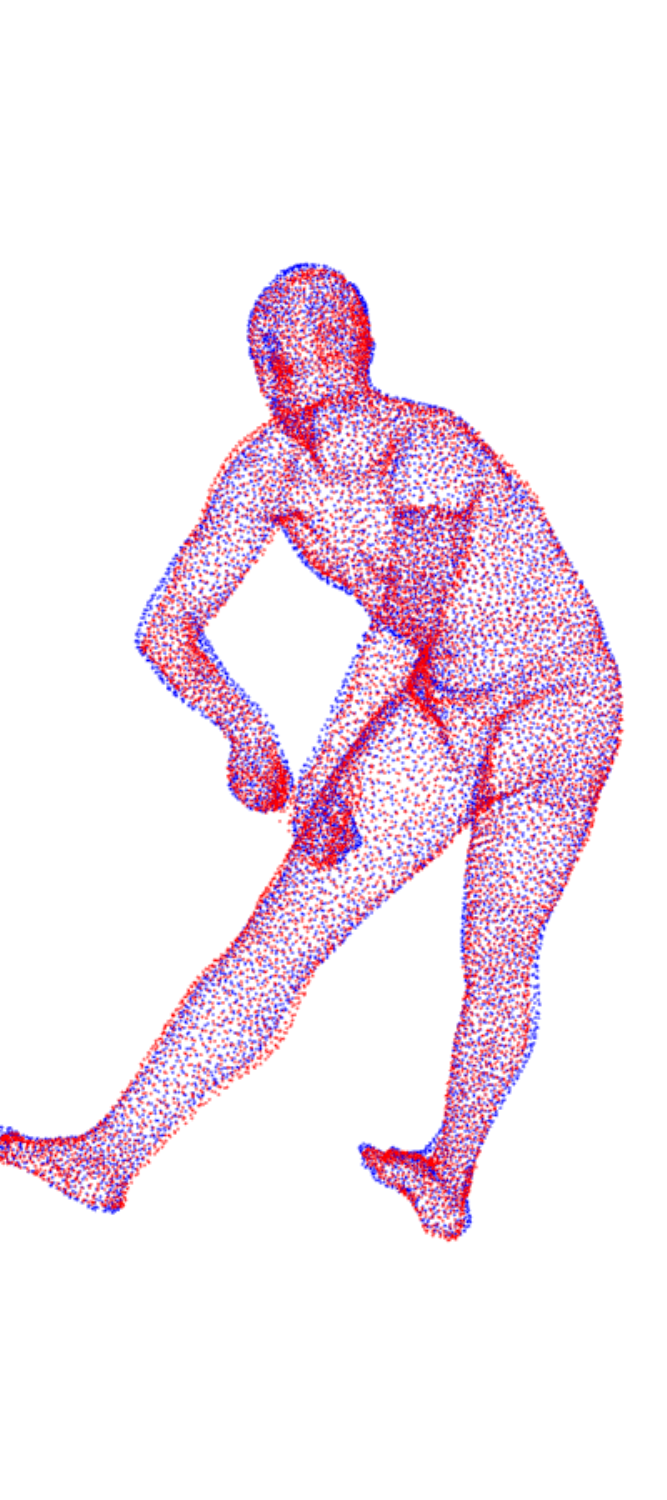}
     \hspace{-0.0cm}\includegraphics[width=0.15\textwidth]{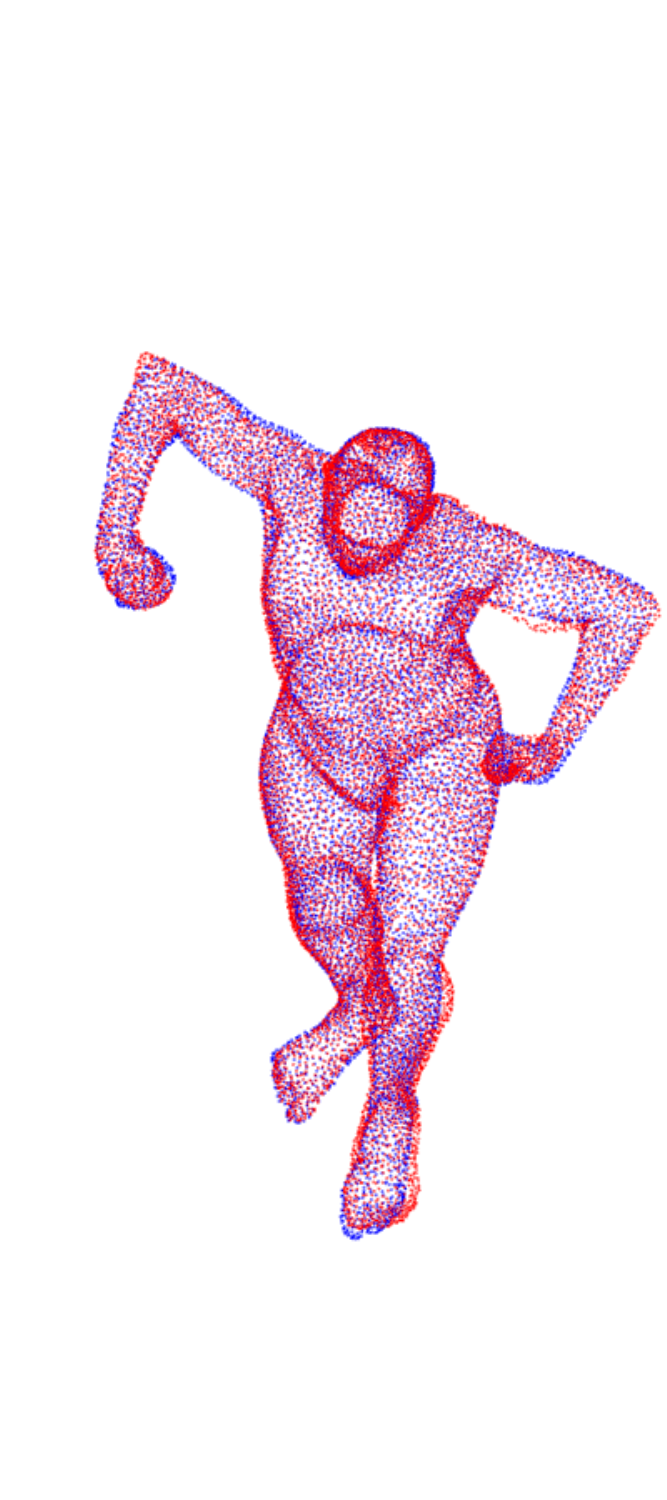}
     \hspace{-0.0cm}\includegraphics[width=0.15\textwidth]{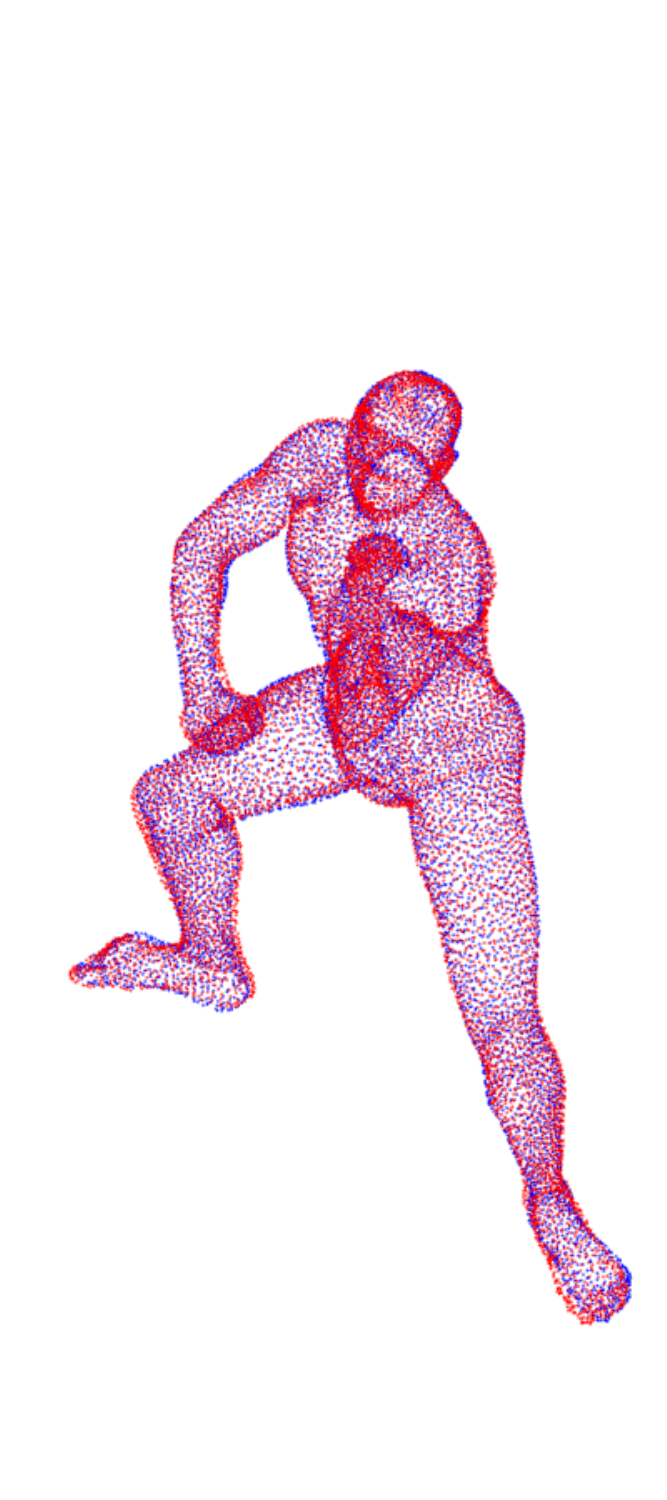}
     \hspace{-0.0cm}\includegraphics[width=0.15\textwidth]{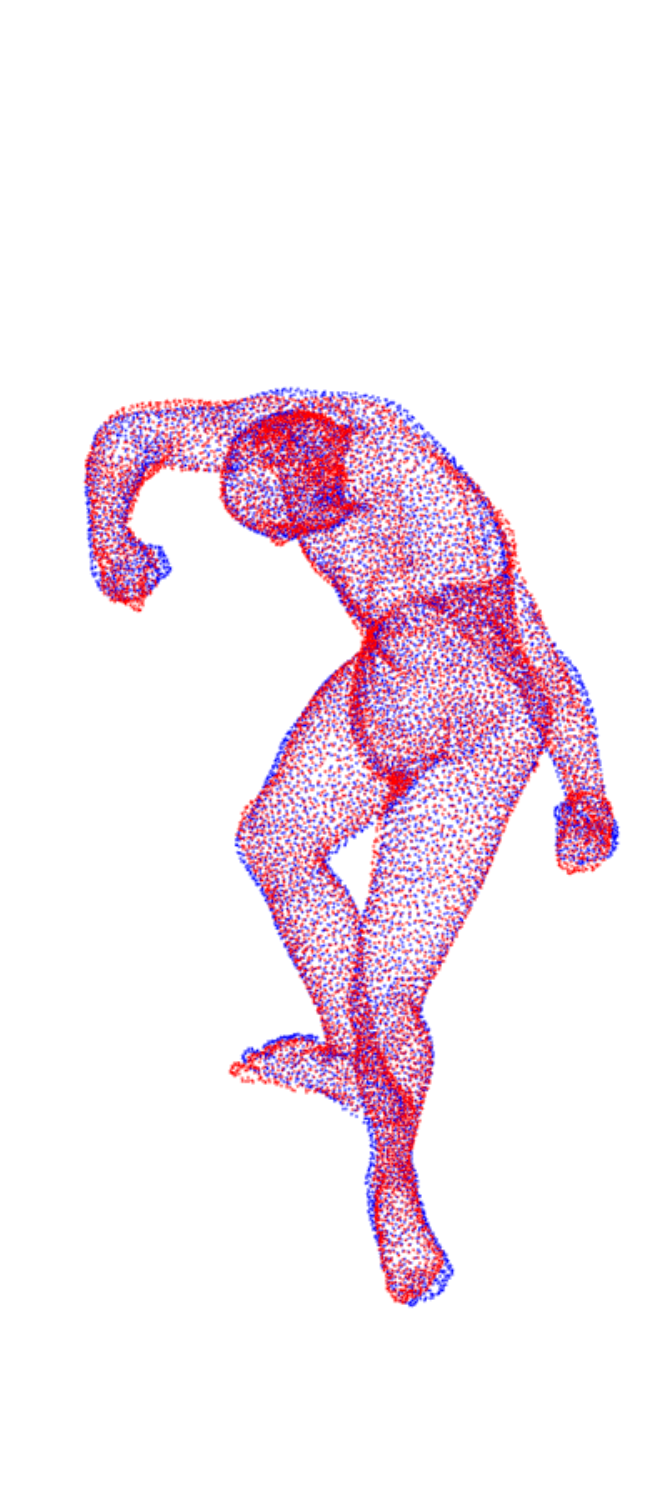}
     \hspace{-0.0cm}\includegraphics[width=0.15\textwidth]{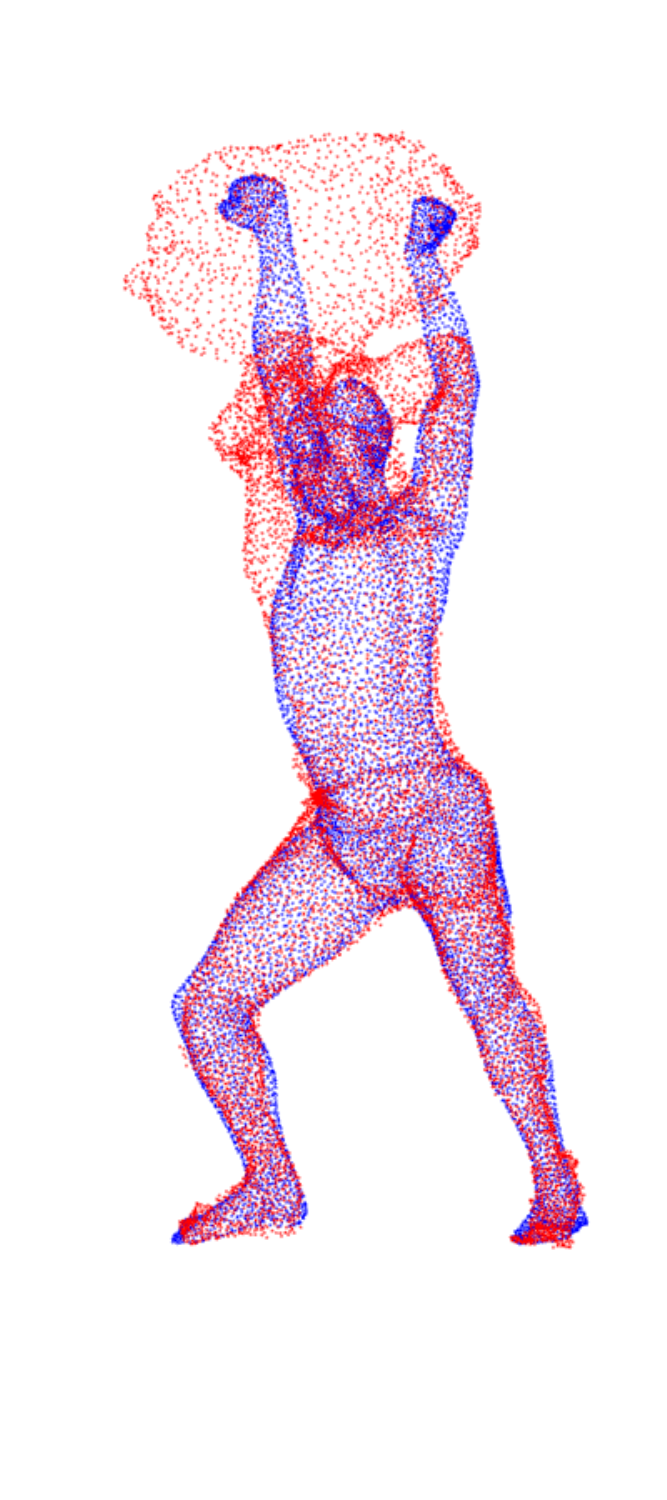}
   \ifDCOL
     \vspace{-0.5cm}
   \fi
    \caption{
      Resulting deformed shapes for five bodies with different body postures 
      obtained from the SCAPE dataset.
      %Point set registration between point sets using the proposed algorithm.
      The left four figures show success examples and the rightmost figure
      shows a failure example.
    }
    \label{fig:scape}
\end{figure*}

\begin{figure}
   \centering
     \hspace{-0.0cm}\includegraphics[width=0.40\textwidth]{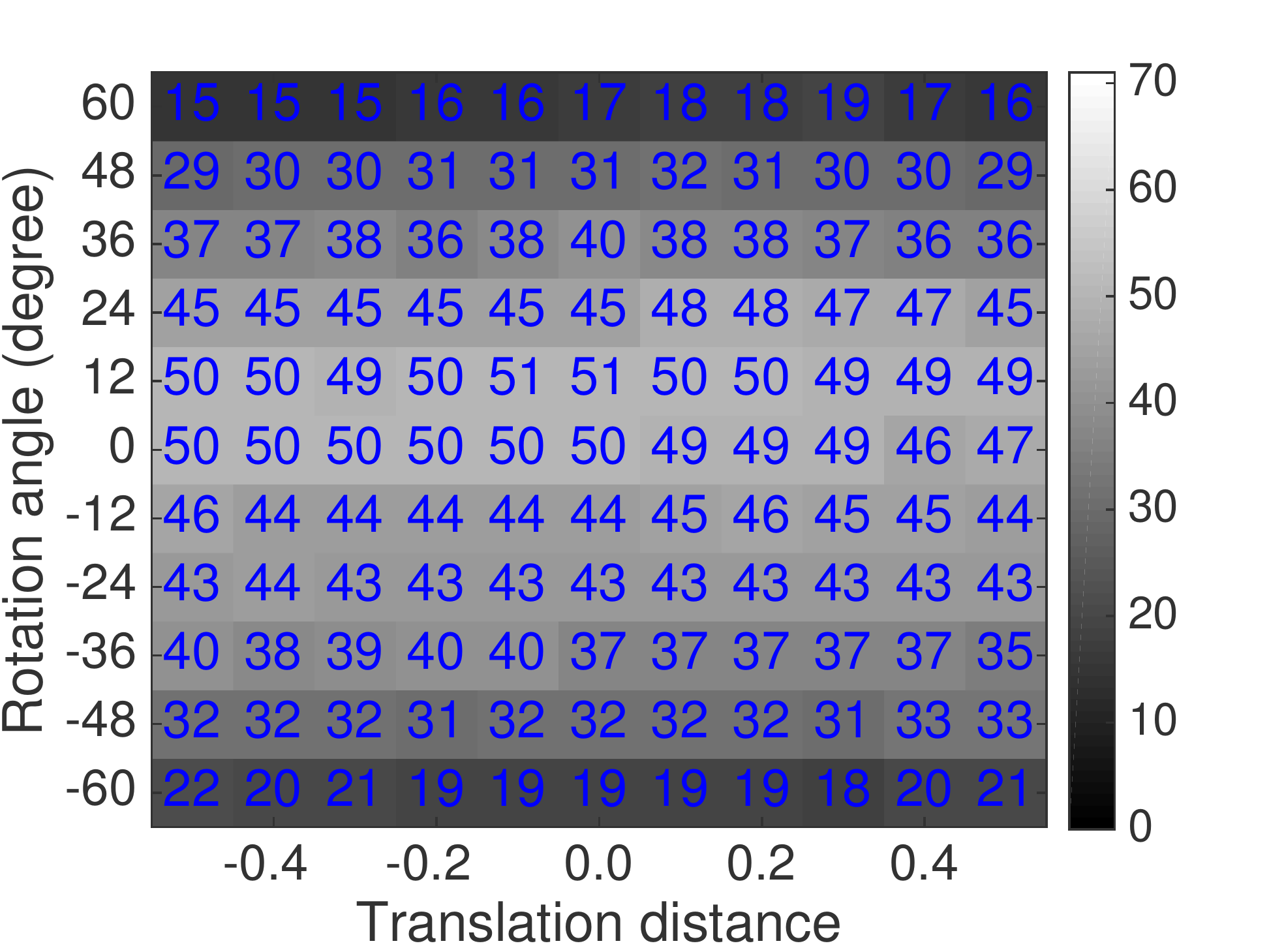} 
     \hspace{-0.0cm}\includegraphics[width=0.40\textwidth]{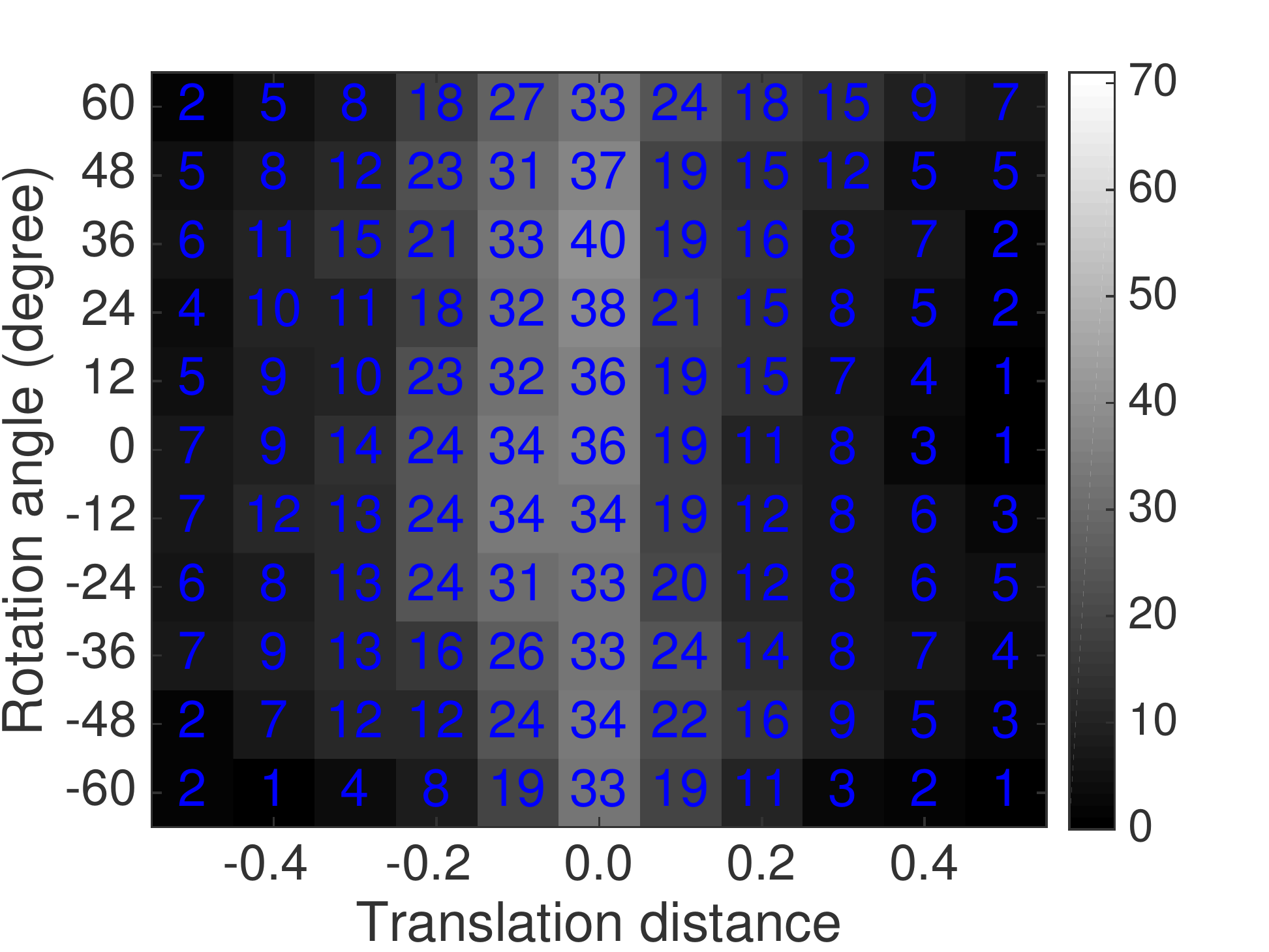}
    \caption{
      \blue{
      The performance comparison between the DLD without the pre-alignment (top) 
      and the GMM-ASM with the pre-alignment (bottom). 
      The number shown in each square represents the number of successes for 71 different body postures,
      obtained from the SCAPE dataset. Each trial was conducted in a leave-one-out manner. 
      The position of the square represents the rigid transformation added to the 71 body postures. 
      }
    }
    \label{fig:heatmap-scape}
\end{figure}

%A trial was defiend as a success if the maximum distance between a was less than 0.15
      %For GMM-ASM, the mean shape was initially aligned to each target body shape by the rigid CPD.

%% \input{fig-ehand.tex}
%% \input{fig-eface.tex}

In this section, we report the registration performance and the computational time of the proposed algorithm.
The registration performance will be tested through experimental comparisons
with supervised learning approaches for 3D examples and unsupervised learning approaches for 2D examples.
The supervised learning approaches we compared were GMM-ASM \cite{Rasoulian2012} and an ICP method 
motivated by \cite{Allen2003,Zhou2016,Hirshberg2012}, whereas the unsupervised ones were CPD \cite{Myronenko2010}
and TPS-RPM \cite{Chui2000} for which the authors of the papers distributed high-quality implementations.
The time taken for computing pairwise matching probabilities for the two approximation methods, the Nystr{\"o{}m method \cite{Williams2001} and 
the IFGT \cite{Yang2003,Raykar2005} will be compared with the 
direct computations. 

\subsection{Datasets and training statistical shape models}
We used and the FLAME dataset \cite{Li2017}  and the SCAPE dataset \cite{Anguelov2005} 
as 3D examples and the IMM hand and face datasets \cite{Stewart2003} as 2D examples.
The SCAPE dataset includes 71 human body shapes with different body postures, each of which
is composed of 12,500 points. The ground truth data for point-by-point correspondence were provided for all 71 point sets.
The FLAME dataset constitutes time-lapse 3D face models with various facial expressions for 10 people.
The number of face models included in the dataset is 46,905, and each face model is composed of 5,023 points
with point-by-point correspondence. Among them, we randomly extracted 950 faces without considering
the differences in facial expressions and human identities. For each face model, we used 3,929 points obtained by
removing 1,094 points corresponding to eyeballs to visualize the point sets around the eyes clearly.
For IMM datasets, each human hand and face is composed of 56 and 58 points. 
%% each shape was obtained from a 2D image of a human hand or a human face
%% by manually placing 56 and 58 landmarks, respectively. 
The points in the shapes were correspondent across all shapes for each dataset.
The number of shapes in the hand dataset was 40, and the face dataset consisted of 240 shapes, 
i.e., six faces with different angles per person for 40 different people.

As mentioned in the previous section, the effects of scale, rotation, and translation are typically removed 
from the training data to learn the statistical shape model. 
Therefore, we pre-processed the shapes in the original datasets as follows:
For each dataset, we first calculated the mean shape as the sample average of all shapes.
We then constructed training data by computing shapes with the best rigid alignment with
the mean shape. Note that, from Result 7.1 described in \cite{Dryden2016}, 
it is possible to find the best rigid transformation to match two shapes, i.e., two 
point sets with point-by-point correspondence. For the SCAPE and FLAME datasets, we repeated the computation process
of the mean shape three times to improve the quality of the mean shape. 
Finally, for each body shape, we multiplied the scale factor to normalize the size of the
mean shape for more interpretable experimental results. The scale factor was obtained by setting the volume of the cuboid that envelopes the mean shape to one.
For the SCAPE and IMM datasets,
the statistical shape model required for the registration was trained in a leave-one-out manner, 
i.e., one for a validation point set and the others for training shapes.
For the FLAME dataset, we used 900 shapes as a training dataset and the remaining 50 shapes
as a test dataset.

\subsection{Application to the SCAPE dataset}
Here, we report the applications to the SCAPE data set, which is composed of 71 human bodies
with various body postures. The primary purpose of the applications is to compare
(1) the registration accuracy of the DLD with that of the GMM-ASM \cite{Rasoulian2012},
and (2) the CPU times of the DLD with or without the acceleration by the approximation techniques.
Throughout the applications to the SCAPE dataset, we use the common parameters
$(K,\omega,\gamma)=(70,10^{-4},10^{-4})$. %and the adaptive control of $\gamma$ was turned on.

We first report the application to several human body shapes for the demonstration purpose. 
To demonstrate the robustness against missing regions, we generated a point set by 
removing 2,000 points from the legs and 2,000 points above the shoulder. 
The number of points used in the Nystr{\"o}m method was set to 500.
Figure \ref{fig:teaser} shows the progress of the optimization for human body No. 1 with missing regions.
In this figure, each optimization proceeds from the left to the right. %For both cases,
The DLD algorithm succeeded in the registration for the body shape with missing regions, 
suggesting that it has robustness against missing regions. % and outliers.
Figure \ref{fig:scape} shows typical examples of success trials and a failure trial
for different body postures. These examples suggested that our algorithm worked effectively for
different body postures without any extra information such as colors, surface normal, and
human body articulation. However, the examples also suggested that the registration error 
could not always be removed.
\ifDCOL
The optimization trajectories for this shape are shown in Supplementary Video 1.
\fi

\subsubsection*{Performance comparison}

We compared the registration accuracy of the DLD algorithm with that of the GMM-ASM \cite{Rasoulian2012}.
By this comparison, we aim at quantifying the improvement in the registration performance,
originating in two differences between the methods: (1) the simultaneous optimization 
of the shape and location parameters and (2) the adaptive control of the smoothness
of the displacement field.
To this aim, the comparison with the GMM-ASM is reasonable 
because the DLD algorithm is the same as their method except for the above two differences
and the acceleration by the Nystr\"om method.

To evaluate the robustness against the similarity transformation, we generated target point sets 
by adding combinations of the rotation and translation to each body shape.
We changed the rotation angle from $-\pi/3$ to $\pi/3$ at intervals 
of $\pi/15$ and changed the translation distance from $-0.5$ to $0.5$ at intervals of $0.1$.
The maximum translation distance corresponds to roughly one-fourth of the height for the mean shape.
For the GMM-ASM, the mean shape was initially aligned to each body shape by using the rigid CPD
\cite{Myronenko2010} to minimize the disadvantage in the definition of the transformation model.
Here, we used the IFGT with error bound $10^{-6}$ as an alternative accelerating method for
both methods to avoid the effect of randomness, originating in the random sampling 
scheme in the Nystr\"om method. 
Figure \ref{fig:heatmap-scape} shows the number of successes for each combination
of a rotation and a translation. A trial was defined as a success if the maximum distance between 
a point in a deformed shape and the corresponding point in a target shape was less than 0.15.
%The distance 0.15 is roughly estimated as 10 cm in the real world. 
In comparison with the GMM-ASM, the DLD was robust against both the rotation and translation; further, 
the effect of the translation was quite small.
On the contrary, the GMM-ASM was less robust
especially against the translation despite of the fact that the pre-alignment by the rigid CPD was conducted 
before fitting the SSM. This result shows that the two differences between these methods 
contributed to the improvement in the registration accuracy.
%

%% This suggest that the pre-alignment by the rigid CPD were not always cons
%% might be because the inaccuracy of the pre-alignment by the rigid CPD

%% The maximum
%% success rate of the DLD and the GMM-ASM was $72\%$

\subsubsection*{Computational time}
\ifDCOL
  \begin{figure}
     \centering
      \vspace{-2cm}
      \includegraphics[width=0.5\textwidth]{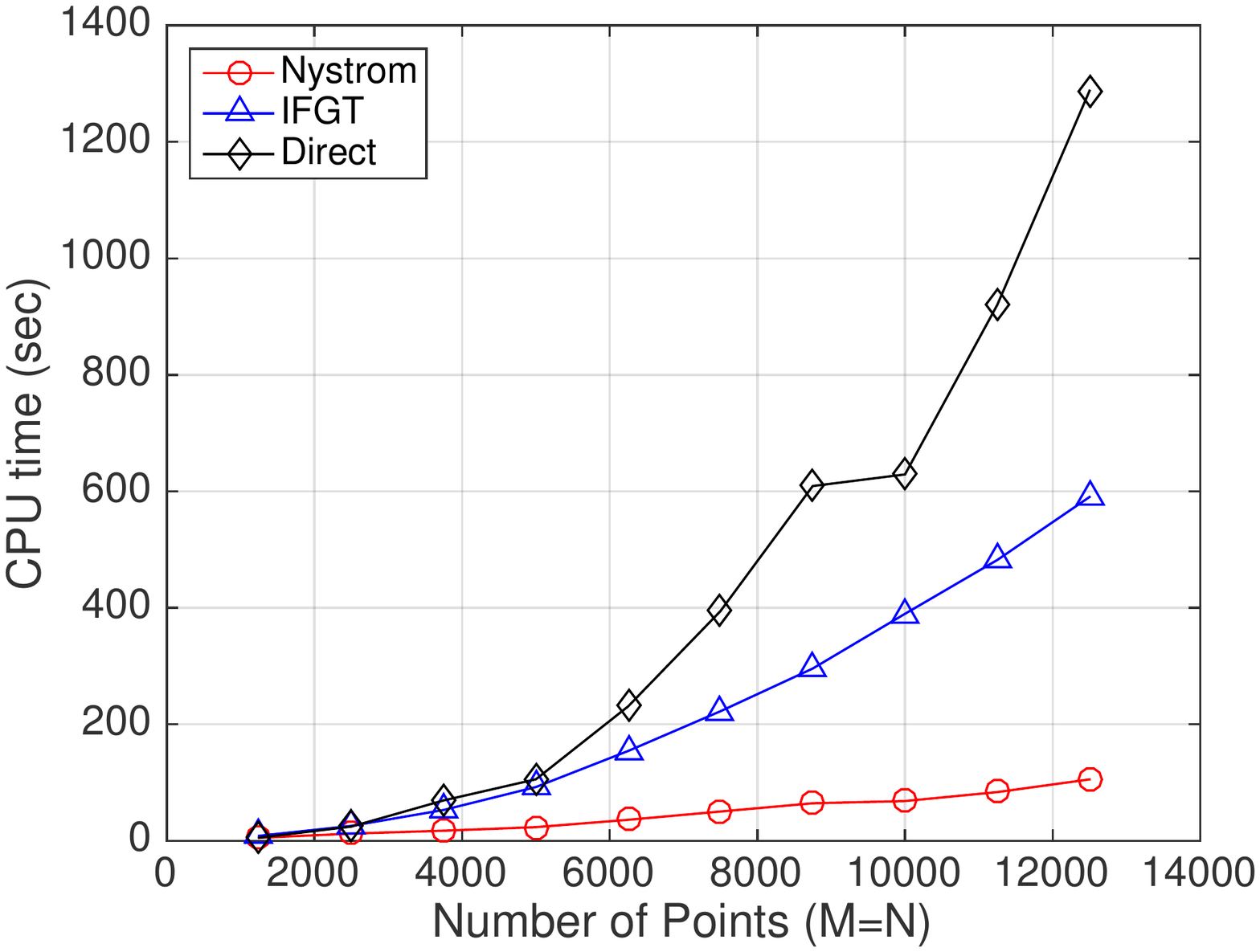}
      \vspace{-3cm}
      \caption{
        Comparison of the CPU time among the direct computation (black),
        the IFGT (blue), and the Nystr{\"o}m method (red).
      }
      \label{fig:cputime}
  \end{figure}
\else
  \begin{figure}
     \centering
      \vspace{-3cm}
      \includegraphics[width=0.6\textwidth]{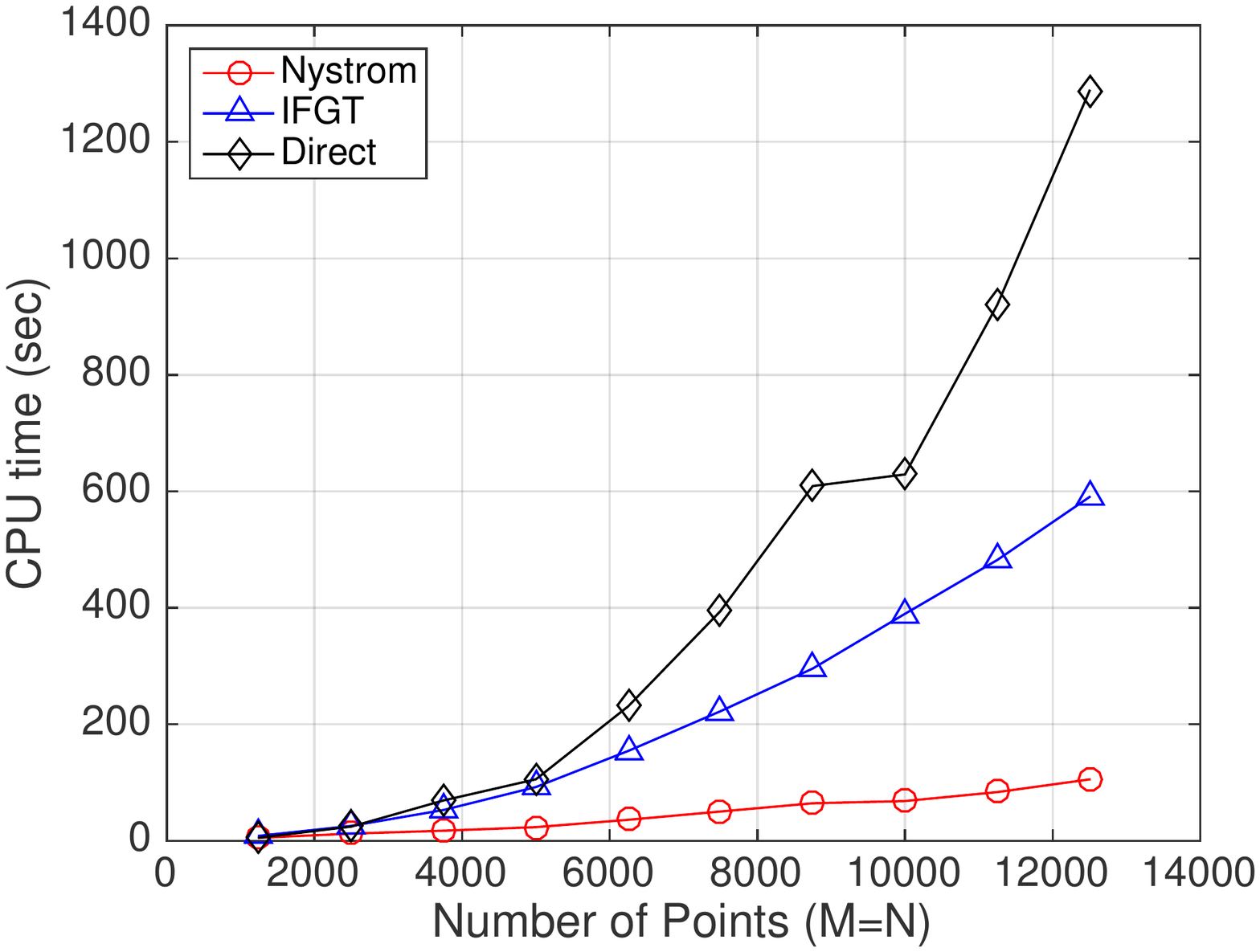}
      \vspace{-3cm}
      \caption{
        Comparison of the CPU time among the direct computation (black),
        the IFGT (blue), and the Nystr{\"o}m method (red).
      }
      \label{fig:cputime}
  \end{figure}
\fi

We measured the CPU time of the proposed algorithm using the human body No. $1$ in the SCAPE dataset
and the average shape calculated for the dataset. To evaluate the scalability of the DLD algorithm, 
we generated 10 pairs of point sets with the same numbers of points by randomly sampling 1250, 2500,
$\cdots$, 12,500 points without replacement from the target point set and the average shape. 
We used a MacBook Pro (Retina 15-inch, Early 2013, OS X El Capitan 10.11.6) with a 2.4 GHz Intel Core i7 
and 16 GB RAM as our computational environment. We implemented the DLD algorithm in the C language, 
and used GCC 6.0 as a C compiler. We compared the direct method for computing the pairwise
probability with two approximation methods: the IFGT \cite{Yang2003,Raykar2005} and the Nyst\"om method
\cite{Williams2001}. The IFGT was implemented according to the technical paper \cite{Raykar2005}, and
the IFGT error bound was set to $10^{-6}$. The number of points used for the Nystr{\"o}m method was
set to 500. We stopped the computation if the improvement of the log-likelihood during one iteration
of the EM algorithm was below $10^{-4}$.
Figure \ref{fig:cputime} shows the resulting CPU times for the three methods.
For the point sets with 1,250 points, the CPU times for the Nystr{\"o}m method, the IFGT, and the
direct computation were 5.19, 8.56, and 5.41 s, respectively.
For the point sets with 12,500 points, i.e., the most severe condition, 
the CPU time for the Nystr{\"o}m method
was 105.4 s, which was 12.2 times faster than the direct computation and 5.6 times faster than the IFGT.
From the figure, we observe that the direct computation was over-linear. We also observe that the IFGT was slightly
over-linear, suggesting that the direct computation included in the IFGT framework was used during the optimization.
The computational time of the Nystr{\"o}m method was linear, which is consistent with the analysis of 
the computational cost described in the previous section. 

\subsection{Application to the FLAME dataset}

\begin{figure}
   \centering
     \hspace{-0.0cm}\includegraphics[width=0.45\textwidth]{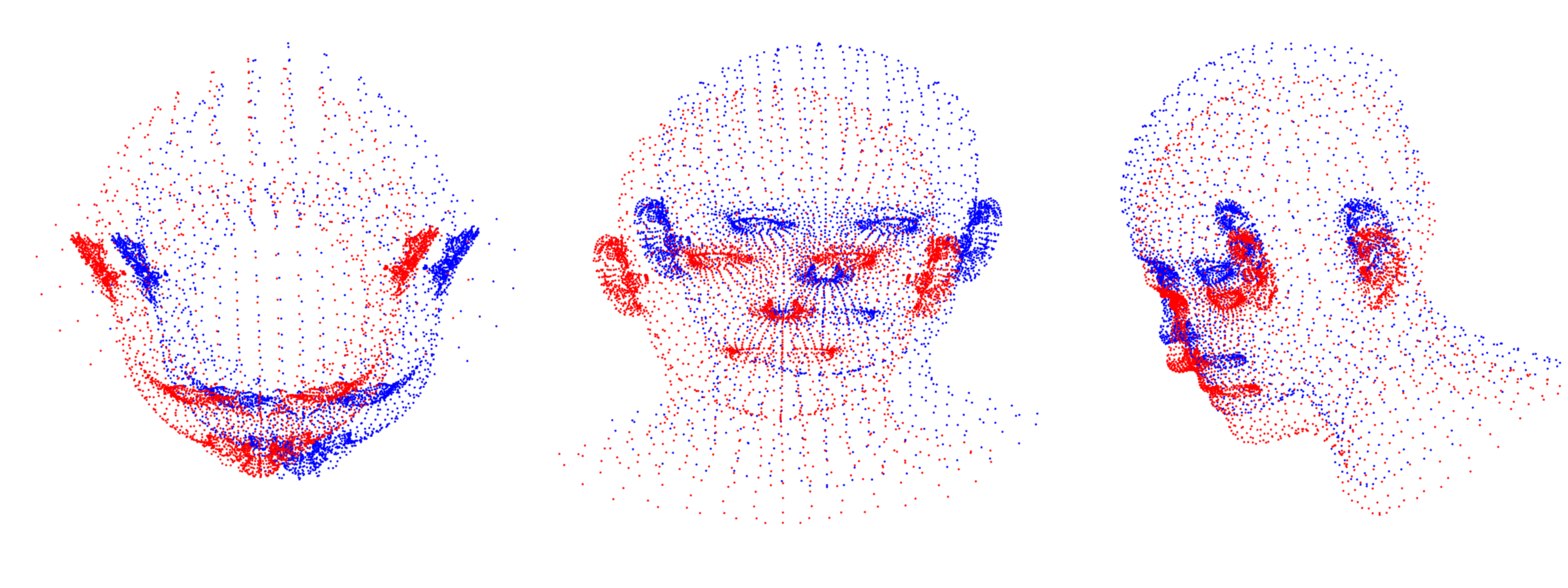} \\
     \hspace{-0.0cm}\includegraphics[width=0.45\textwidth]{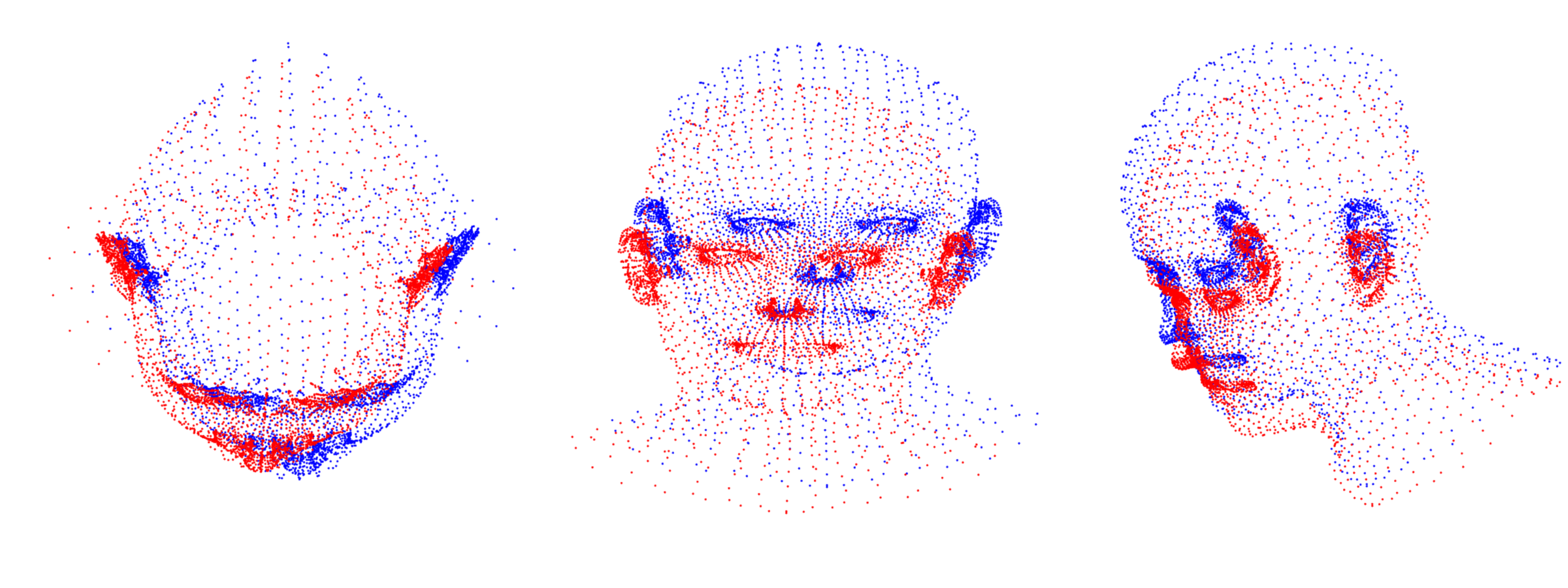}   \\
     \hspace{-0.0cm}\includegraphics[width=0.45\textwidth]{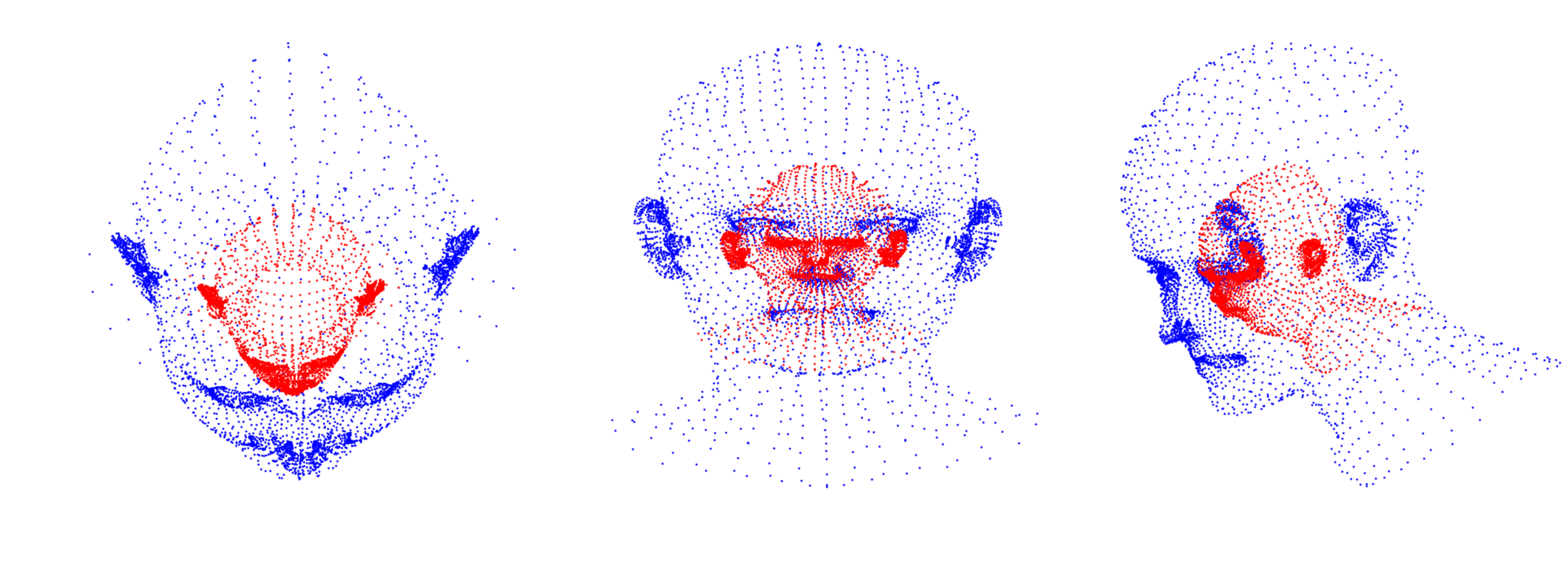}
    \caption{
      \blue{
      The difference in the optimization trajectory between the DLD and the Geman-McClure estimator:
      the face models before the registration (top), the deformed shapes after one step of 
      the optimization by the Geman-McClure estimator (middle) and the DLD (bottom). 
      The target point sets and the source point sets are colored blue and red, respectively.
      The columns of the figure show the face models rendered at different camera positions.
      }
    }
    \label{fig:flame}
\end{figure}

\begin{figure}
   \centering
     \hspace{-0.0cm}\includegraphics[width=0.40\textwidth]{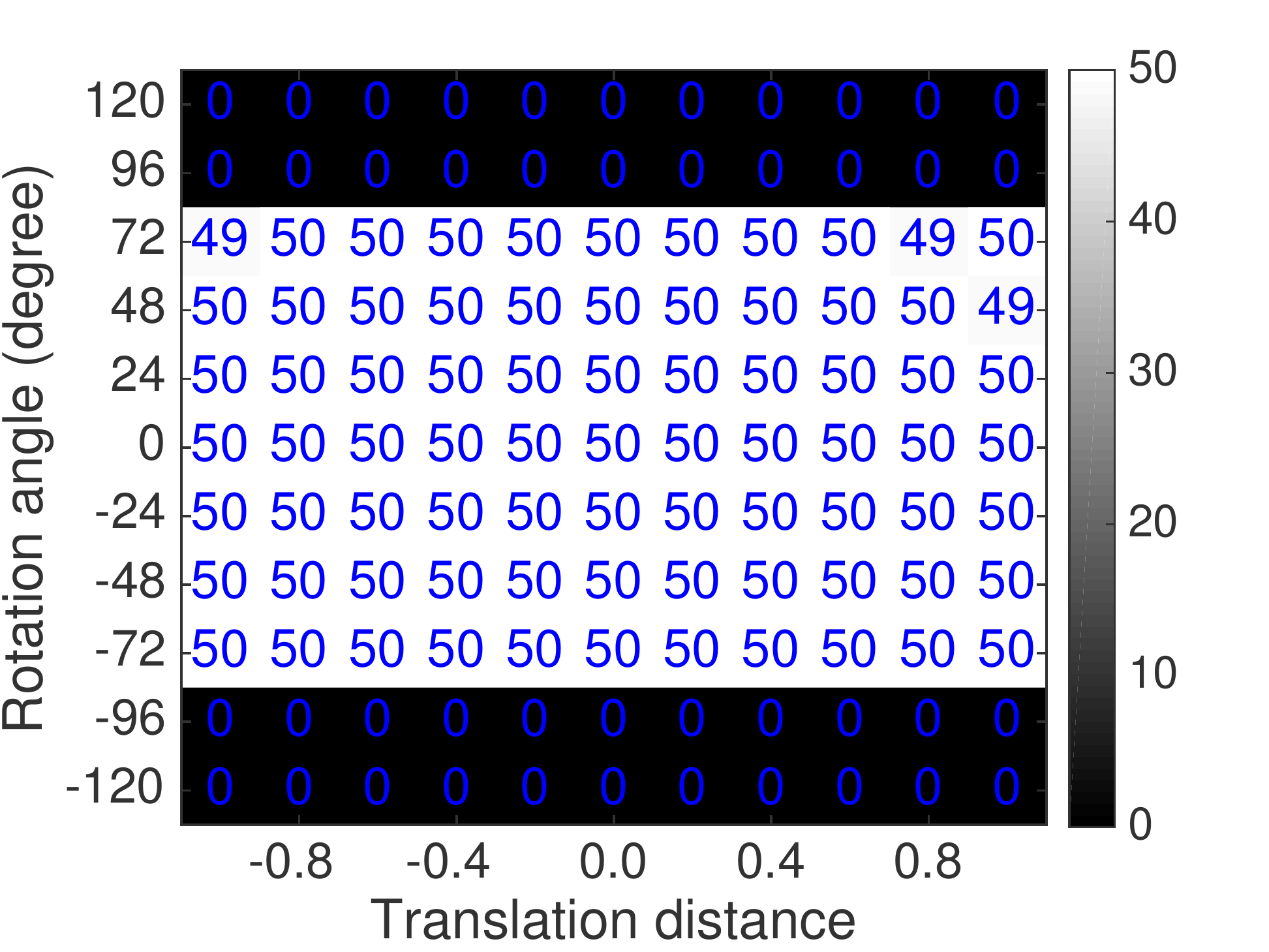} 
     \hspace{-0.0cm}\includegraphics[width=0.40\textwidth]{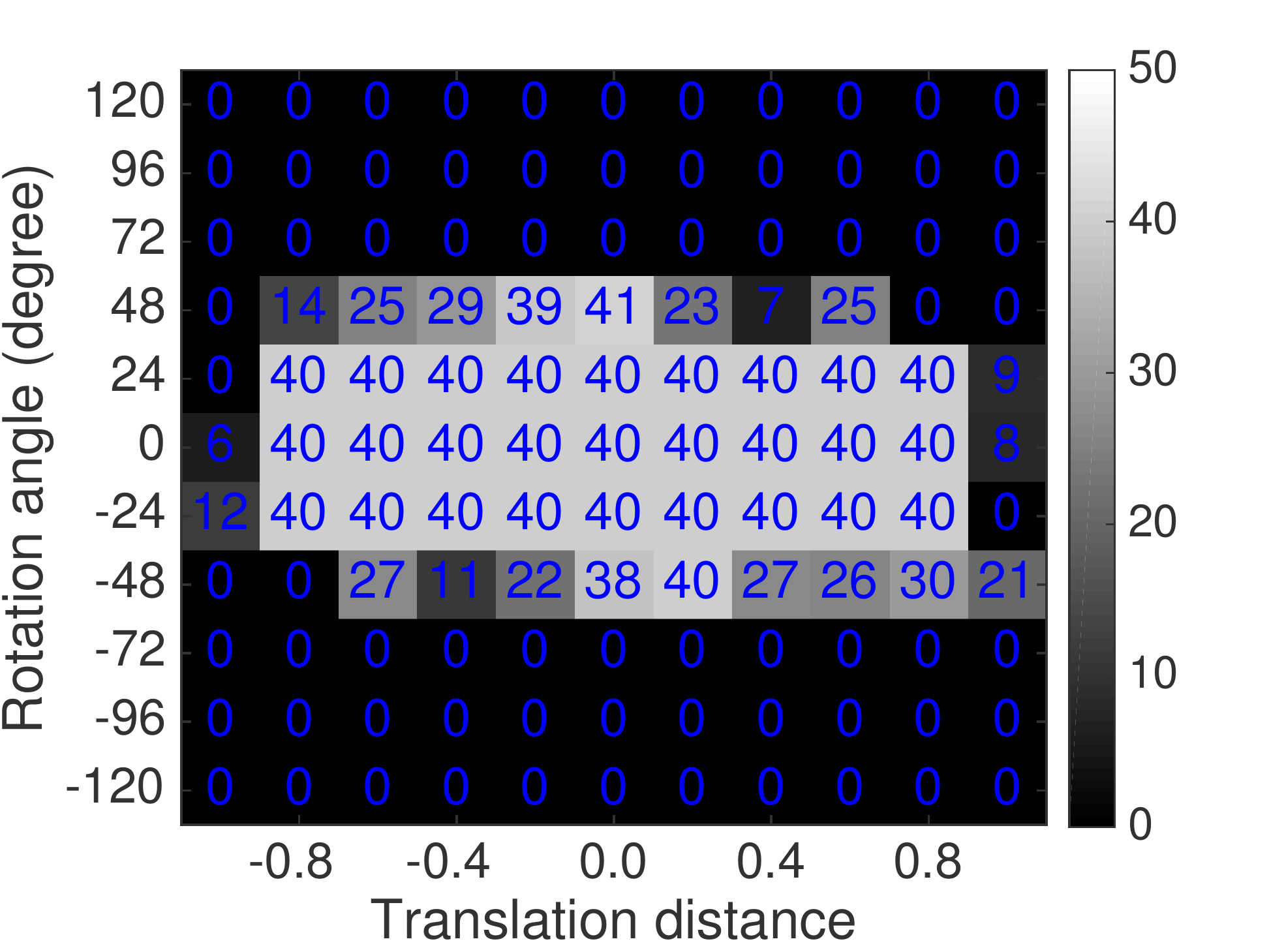}
    \caption{
      \blue{
      The performance comparison between the DLD (top) and the Geman-McClure estimator (bottom).
      The number shown in each square represents the number of successes for 50 different 3D face models,
      extracted from the FLAME dataset.
      The position of the square represents the rigid transformation added to the 50 face models. 
      }
    }
    \label{fig:heatmap-flame}
\end{figure}

We report the applications to the FLAME dataset, which constitutes
time-lapse 3D face models with various facial expressions.
%We compare the DLD with an ICP-based method motivated by Allen et al. \cite{??}.
To evaluate the registration accuracy of the proposed method, we implemented an ICP-based 
method motivated by \cite{Allen2003,Zhou2016,Hirshberg2012} as a competitor.
The method we implemented as a competitor is summarized as follows: 
(1) the energy function is defined on the basis of the theory of a scaled Geman-McClure estimator, 
(2) the transformation model is the same as that of the proposed method,
i.e., $\mathcal{T}_\text{DLD}$, and
%the SSM with the similarity transformation, and
(3) the shape and location parameters are optimized by updating them and the matching weight
alternatively. 
The energy function is formally defined as follows:
\begin{align}
    E= \sum_{n=1}^{N} \sum_{m=1}^M g_{mn} ||x_n-\mathcal{T}_\text{DLD}(u_m;\theta)||^2 
       + \mu (\sqrt{g_{mn}}-1)^2  \nonumber
\end{align}
where %$y_m=\mathcal{T}_\text{DLD}(u_m;\theta)$ is the $m$th  point in the deformed shape,
$g_{mn}=\big(\frac{\mu}{\mu+||x_n-y_m||^2}\big)^2$ is the matching weight between $y_m$ and $x_n$,
and $\mu$ is a parameter that controls the range within which
residuals have a significant effect on the energy function. %The matching weight $g_{mn}$ is
Figure \ref{fig:flame} shows the apparent difference in the optimization trajectories for the DLD and 
the Geman-McClure estimator. For the DLD, the mean shape was shrunk and largely translated 
after one step of the optimization because of the automatic radius control.
On the contrary, for the Geman-McClure estimator, the mean shape was not shrunk, and the translation
distance was considerably small after one step of the optimization.
The complete optimization trajectories are available in Supplementary Video 2.

To evaluate the robustness against the similarity transformation of target point sets,
we added the combinations of the rotation and translation in a similar manner as the applications
to the SCAPE dataset. We changed the rotation angle from $-2\pi/3$ to $2\pi/3$ at intervals of $2\pi/15$
and the translation distance from $-1.0$ to $1.0$ at intervals of $0.2$. The maximum translation distance
$1.0$ roughly corresponds to the length of the average face.
We used the parameters $(K,\omega,\gamma)=(20,10^{-8},10^{-4})$ for the DLD and $(K,\mu)=(20,10^{-4})$
for the Geman-McClure estimator.
 
Figure \ref{fig:heatmap-flame} shows the result of the comparison. Here, a trial was defined as
a success if the maximum distance between a point in a deformed shape and the correspondent point
in the target shape was less than 0.05.
%, meaning that roughly 1.5 cm in the real world.
The DLD was more robust against both the rotation and translation than the Geman-McClure estimator, 
and the success rate was nearly 100 \% if the rotation angle was less than or equal to $2\pi/5$. 
We note that the success rate sharply dropped to 0 \%. This was because the DLD tried to register
point sets upside-down when the rotation angle was greater than or equal to $3\pi/5$.
The Geman-McClure estimator was also robust against the rotation and translation. 
The registration accuracy was, however, relatively less than the DLD.

%% By this experiment, we compare the 
%% registration accuracy originating in the difference
%% between the loss function of the two m
%% between the GMM and the Geman-McClure estimator.

\subsection{Illustrative results for 2D datasets}

\ifDCOL
  \begin{figure*}
    \begin{center} %\vspace{-3.5cm}
       %\begin{tabular}{ccccc}  & Amplification & Missing & Outlier & Rotation  \end{tabular}
       \begin{tabular}{ll}
         \raisebox{ 10.0\totalheight}{\bf True shape} 
           & \hspace{-0.5cm} \includegraphics[width=0.26\textwidth]{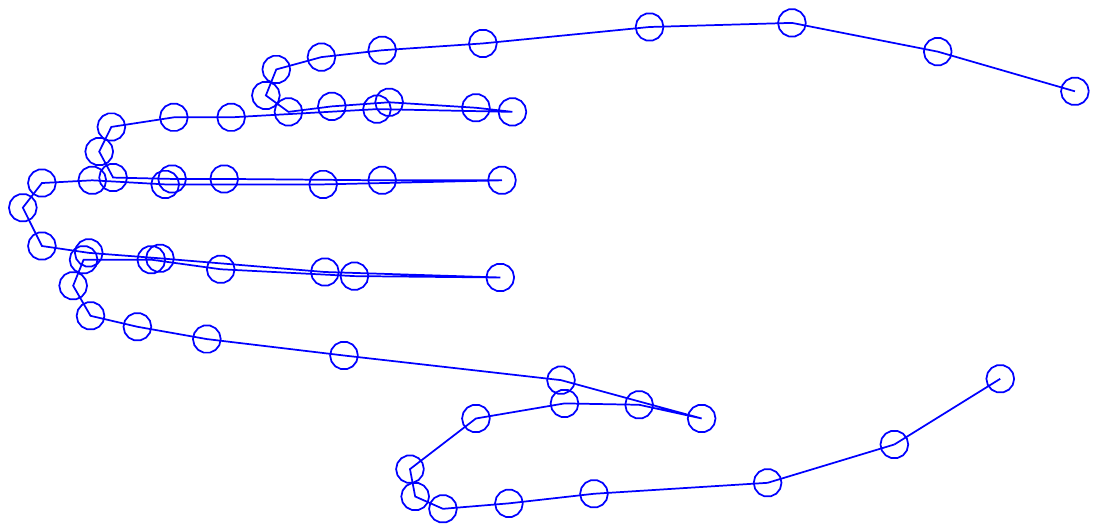}    
             \hspace{-1.2cm} \includegraphics[width=0.26\textwidth]{figure/R1/Demo/hand006.pdf}    
             \hspace{-1.2cm} \includegraphics[width=0.26\textwidth]{figure/R1/Demo/hand006.pdf}    
             \hspace{-1.2cm} \includegraphics[width=0.26\textwidth]{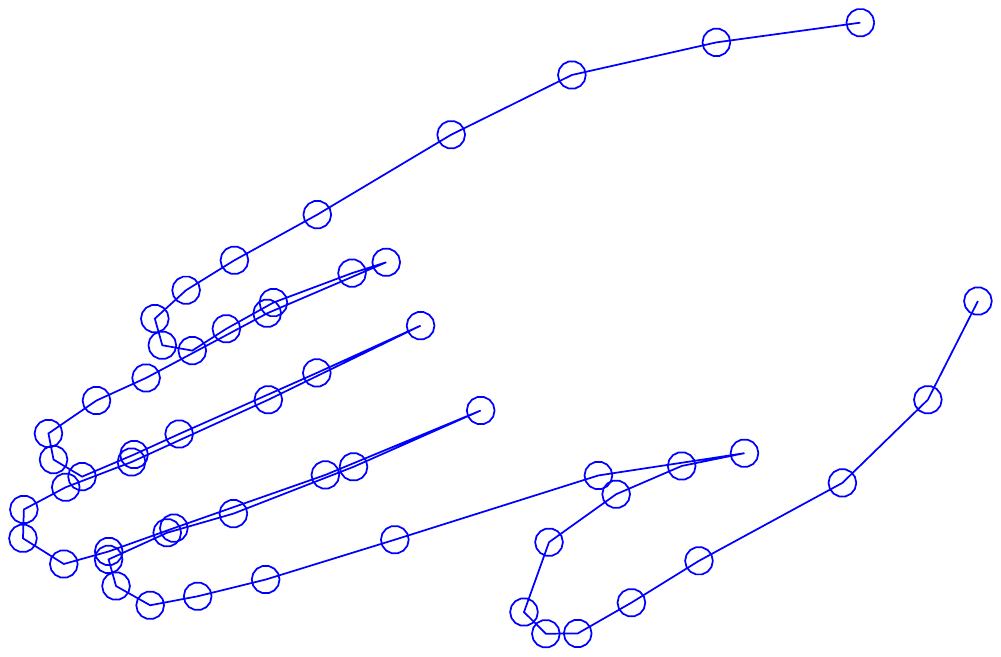}
             \vspace{-3.0cm}\\% &
         \raisebox{ 10.0\totalheight} {\bf Target data}   
           & \hspace{-0.5cm} \includegraphics[width=0.26\textwidth]{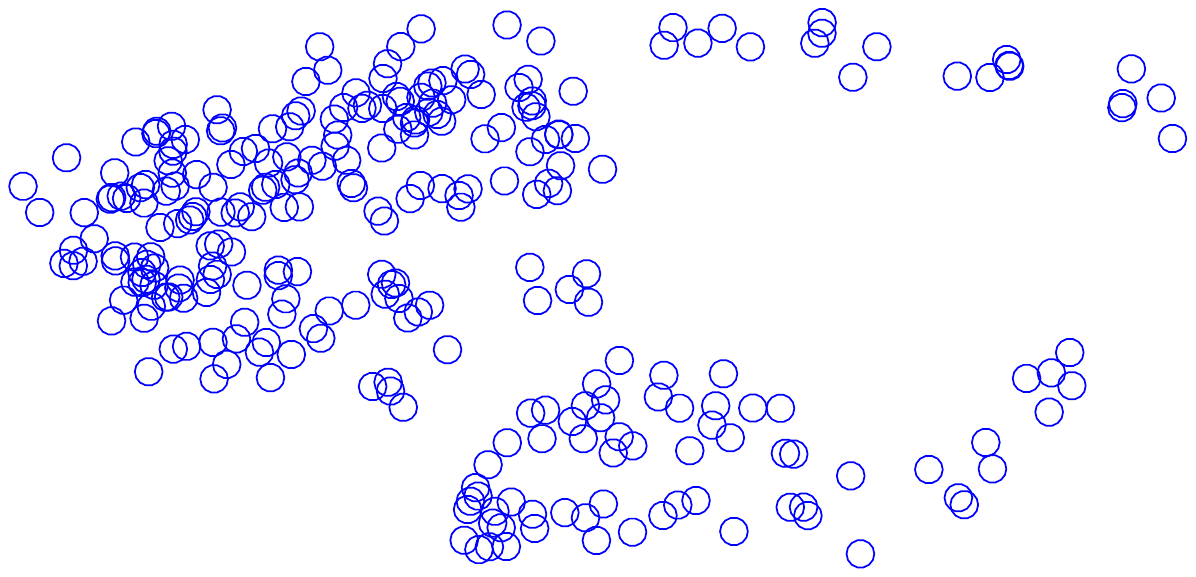}    
             \hspace{-1.2cm} \includegraphics[width=0.26\textwidth]{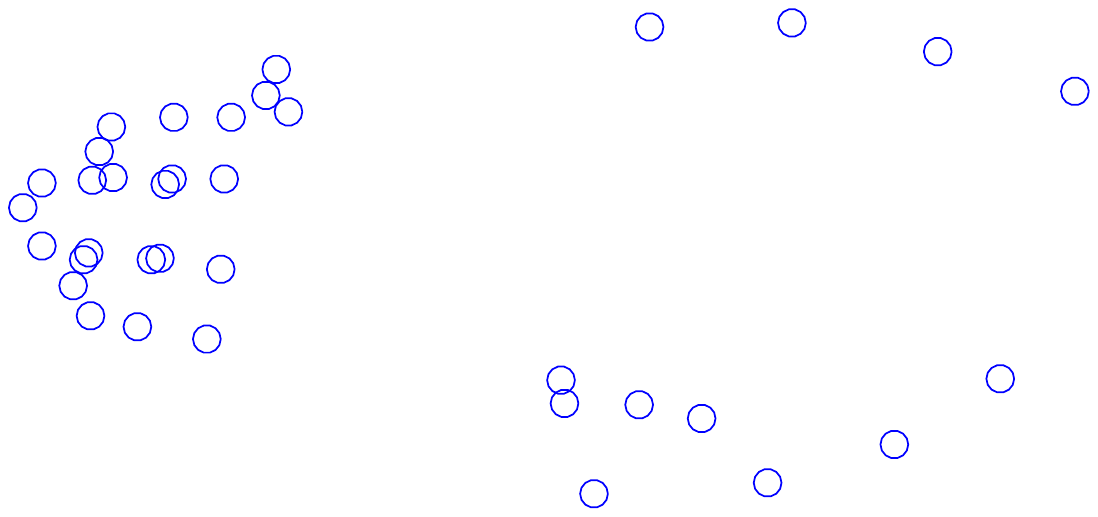}    
             \hspace{-1.2cm} \includegraphics[width=0.26\textwidth]{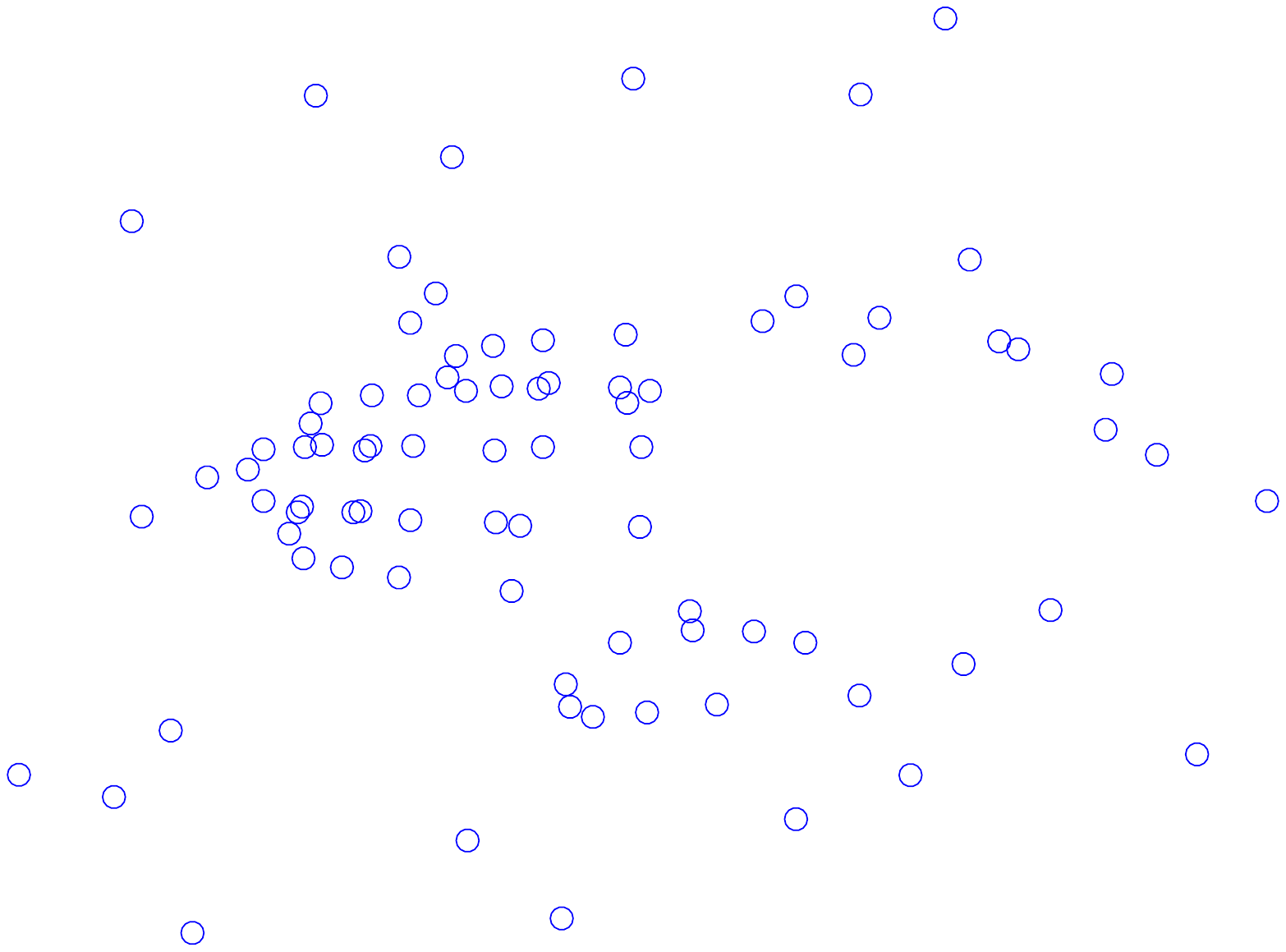}    
             \hspace{-1.2cm} \includegraphics[width=0.26\textwidth]{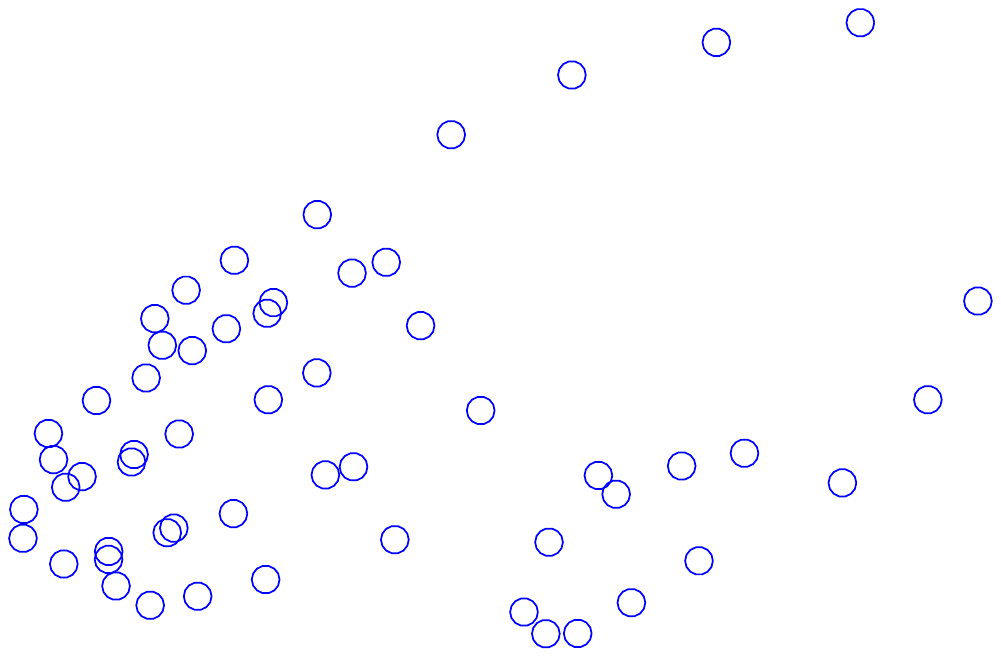}      
             \vspace{-3.0cm}\\% &
         \raisebox{ 12.0\totalheight} {\bf Initial Shape}  
           & \hspace{-0.5cm} \includegraphics[width=0.26\textwidth]{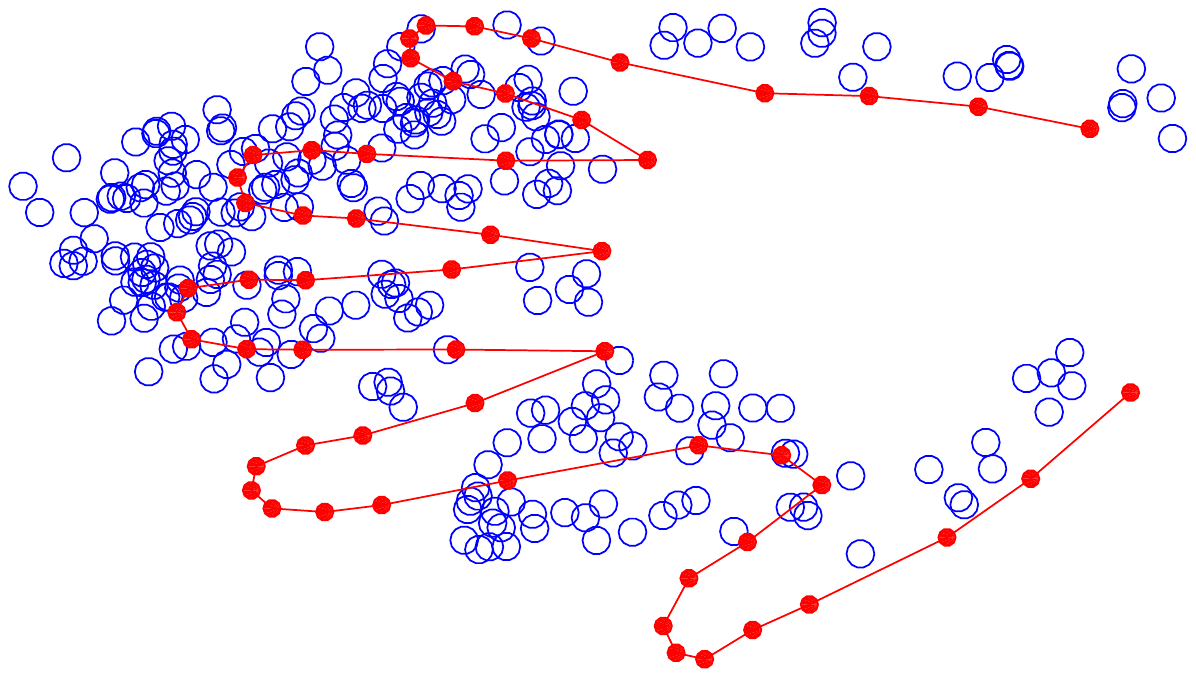} 
             \hspace{-1.2cm} \includegraphics[width=0.26\textwidth]{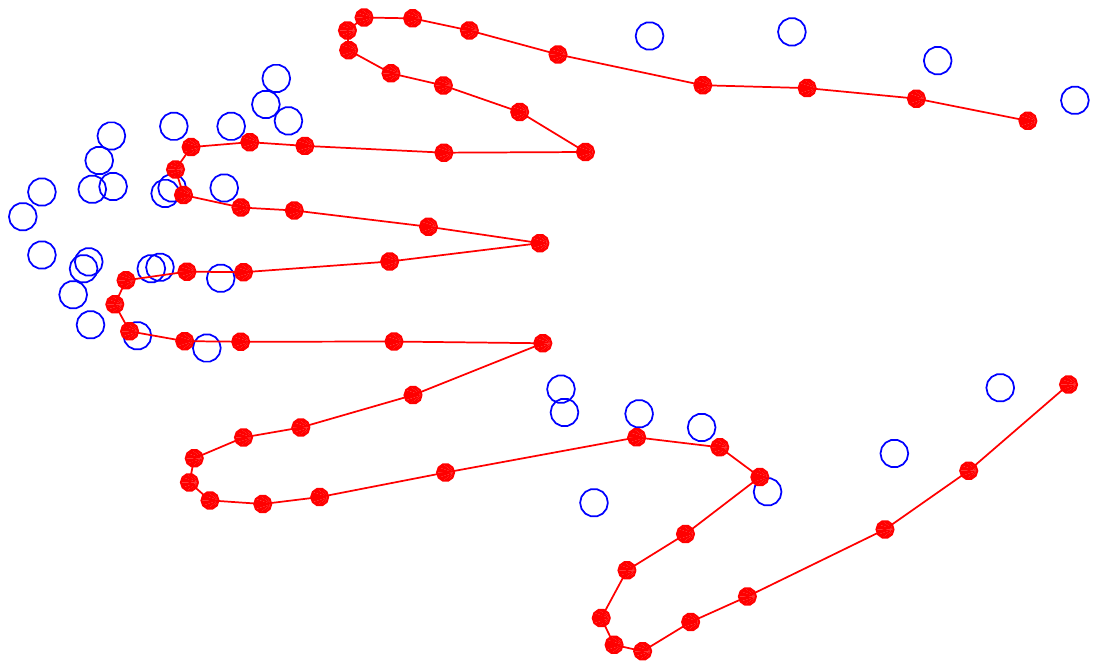} 
             \hspace{-1.2cm} \includegraphics[width=0.26\textwidth]{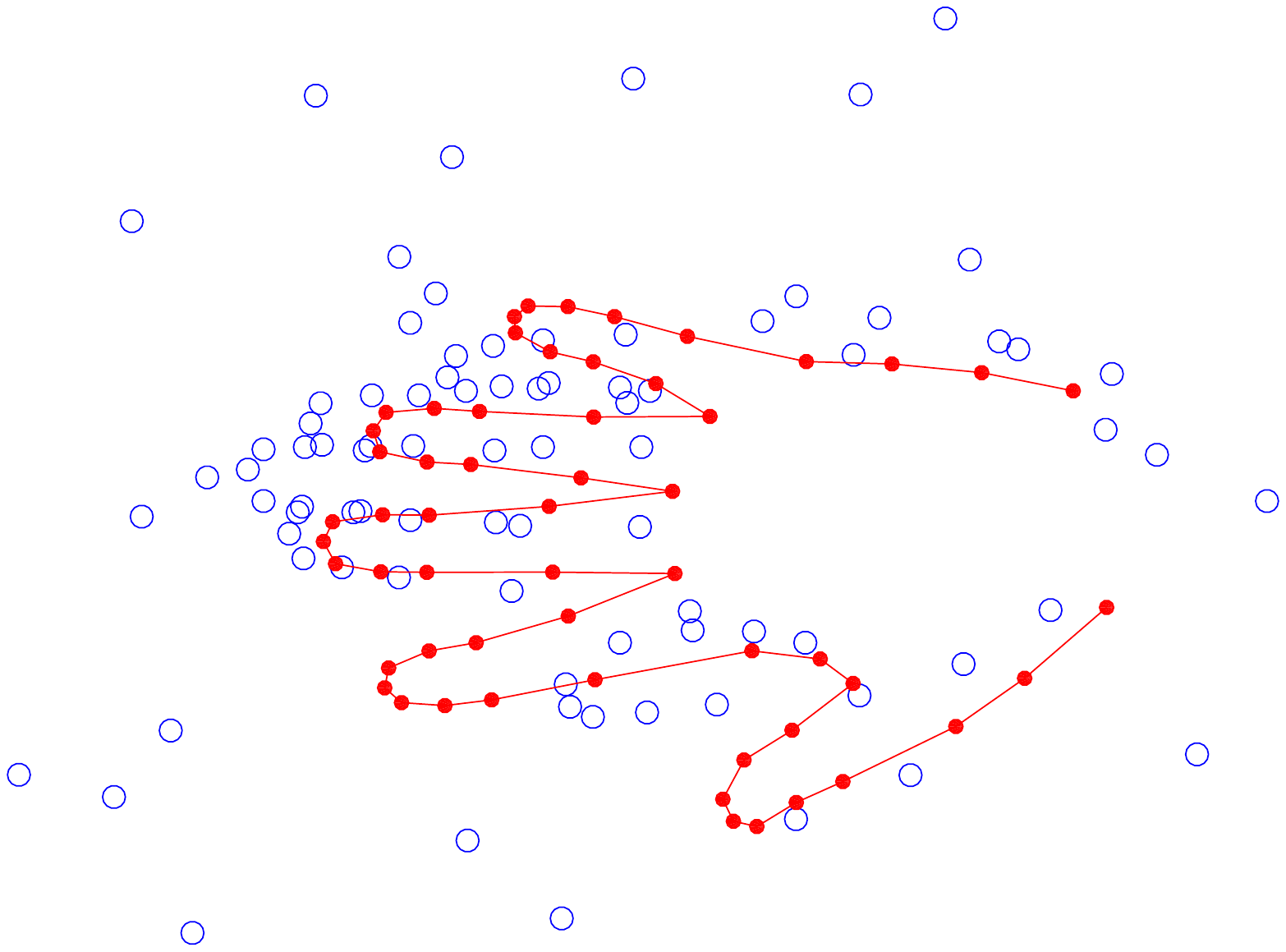} 
             \hspace{-1.2cm} \includegraphics[width=0.26\textwidth]{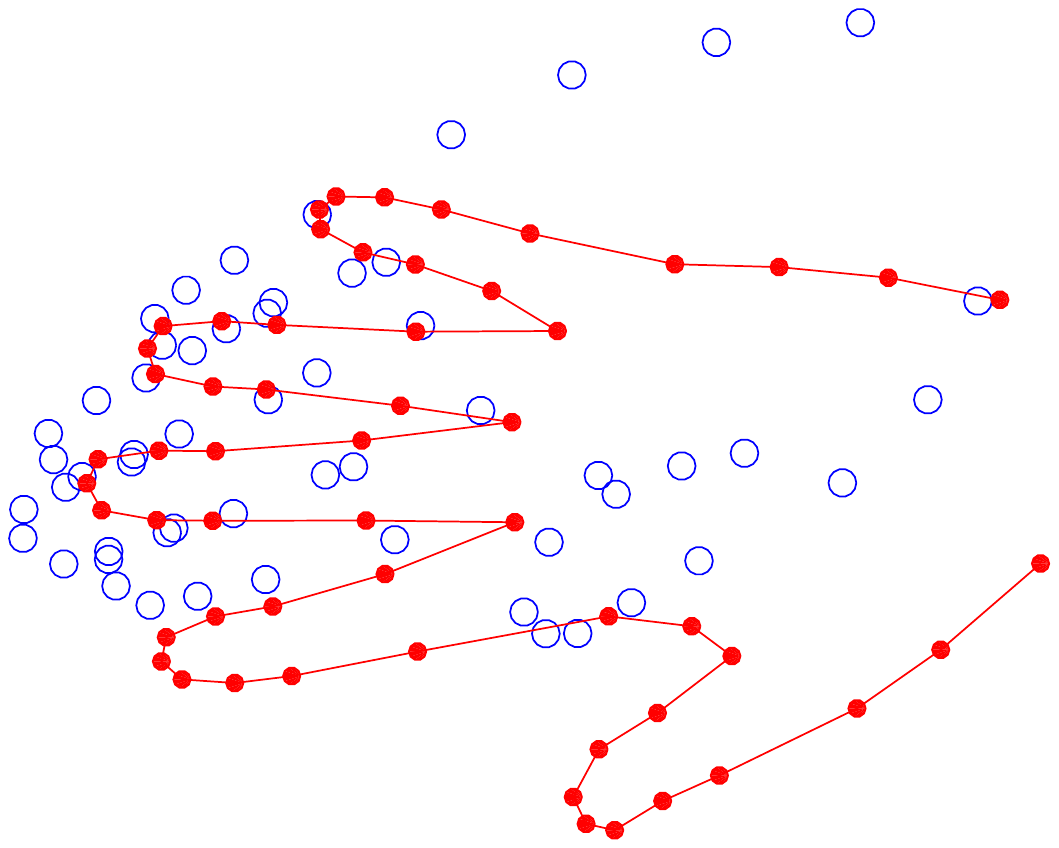}    
             \vspace{-3.0cm}\\ % &
         \raisebox{ 12.0\totalheight} {\bf DLD         }  
           & \hspace{-0.5cm} \includegraphics[width=0.26\textwidth]{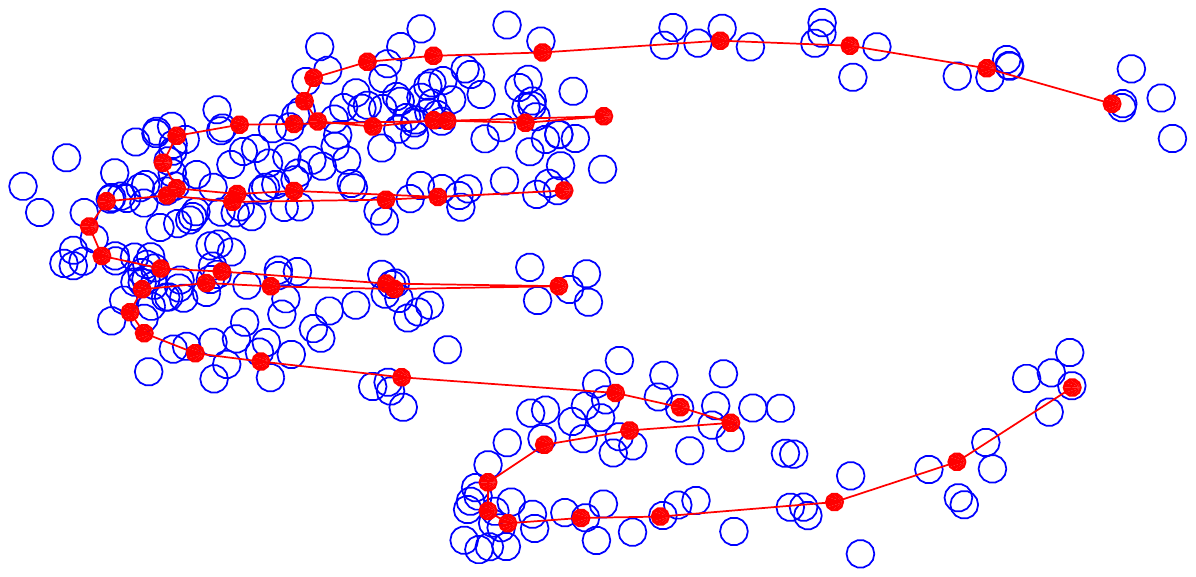}
             \hspace{-1.2cm} \includegraphics[width=0.26\textwidth]{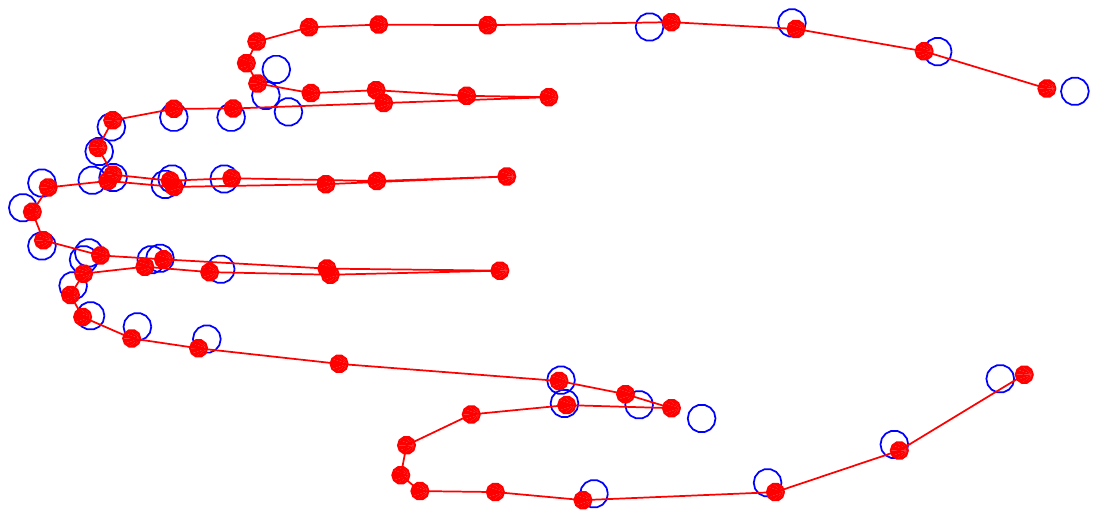}
             \hspace{-1.2cm} \includegraphics[width=0.26\textwidth]{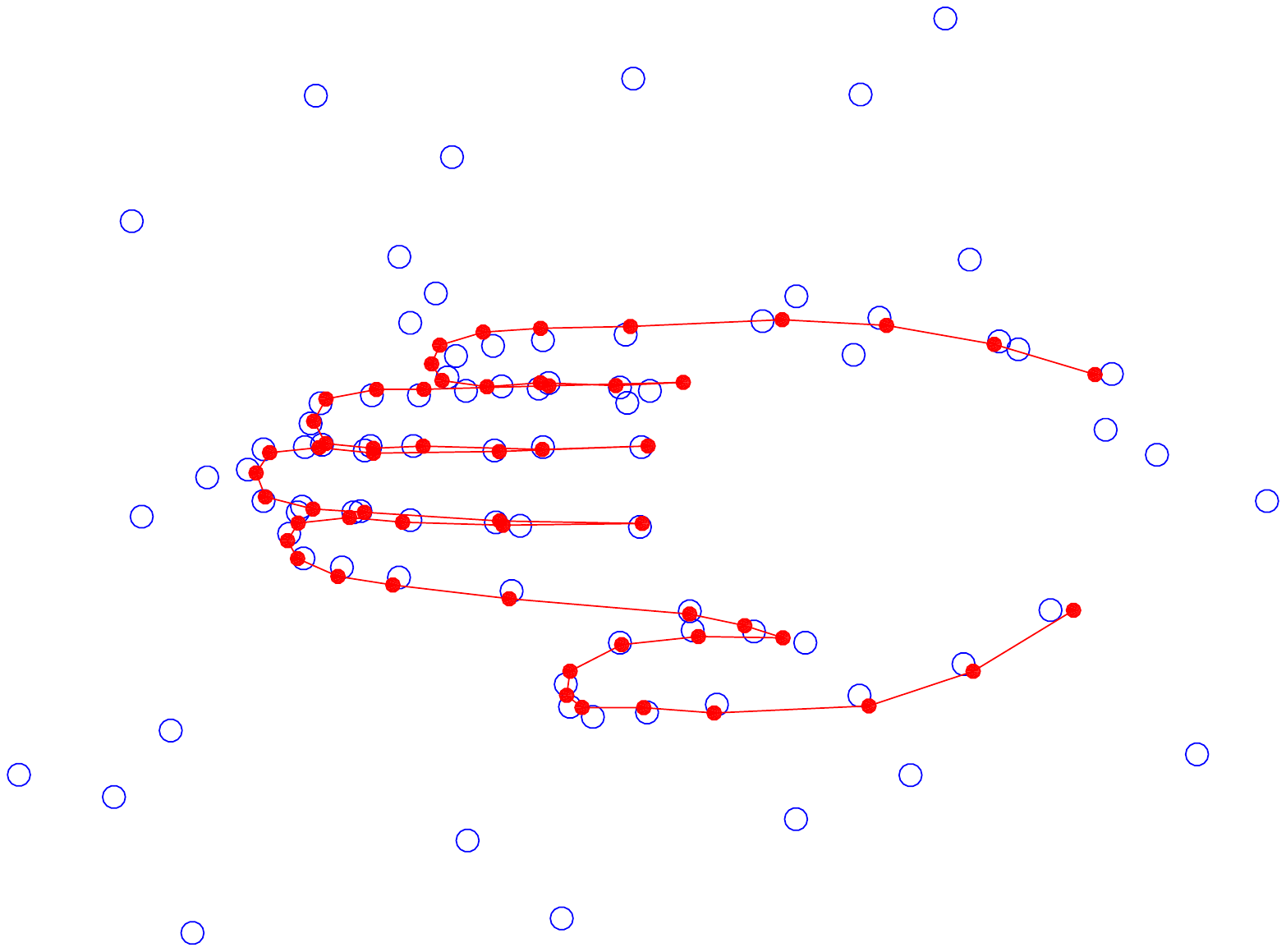}
             \hspace{-1.2cm} \includegraphics[width=0.26\textwidth]{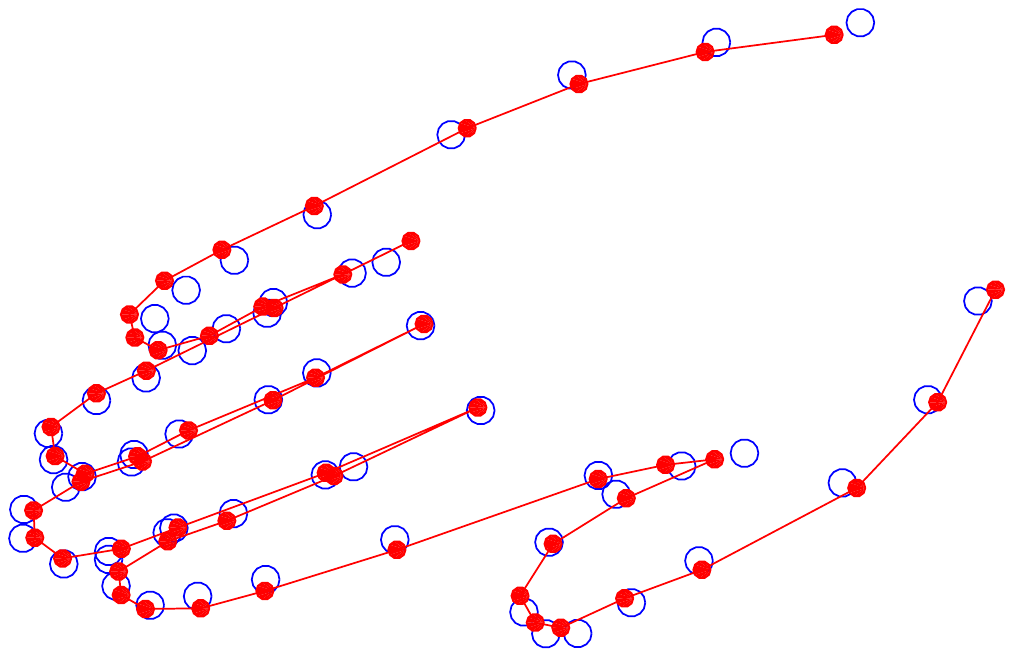} 
             \vspace{-3.0cm}\\% &
         \raisebox{ 12.0\totalheight} {\bf CPD         }  
           & \hspace{-0.5cm} \includegraphics[width=0.26\textwidth]{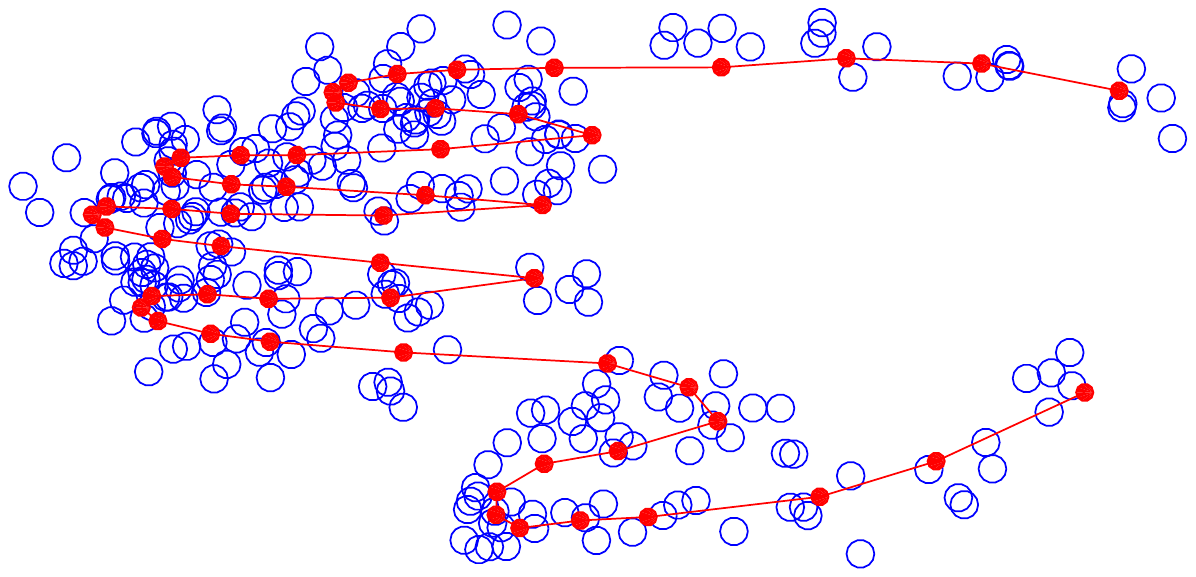} 
             \hspace{-1.2cm} \includegraphics[width=0.26\textwidth]{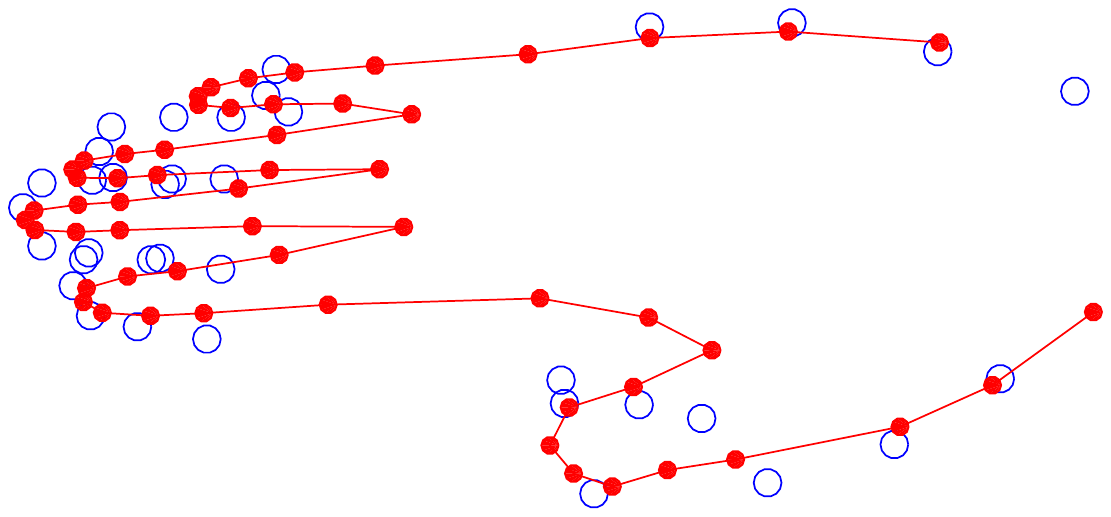} 
             \hspace{-1.2cm} \includegraphics[width=0.26\textwidth]{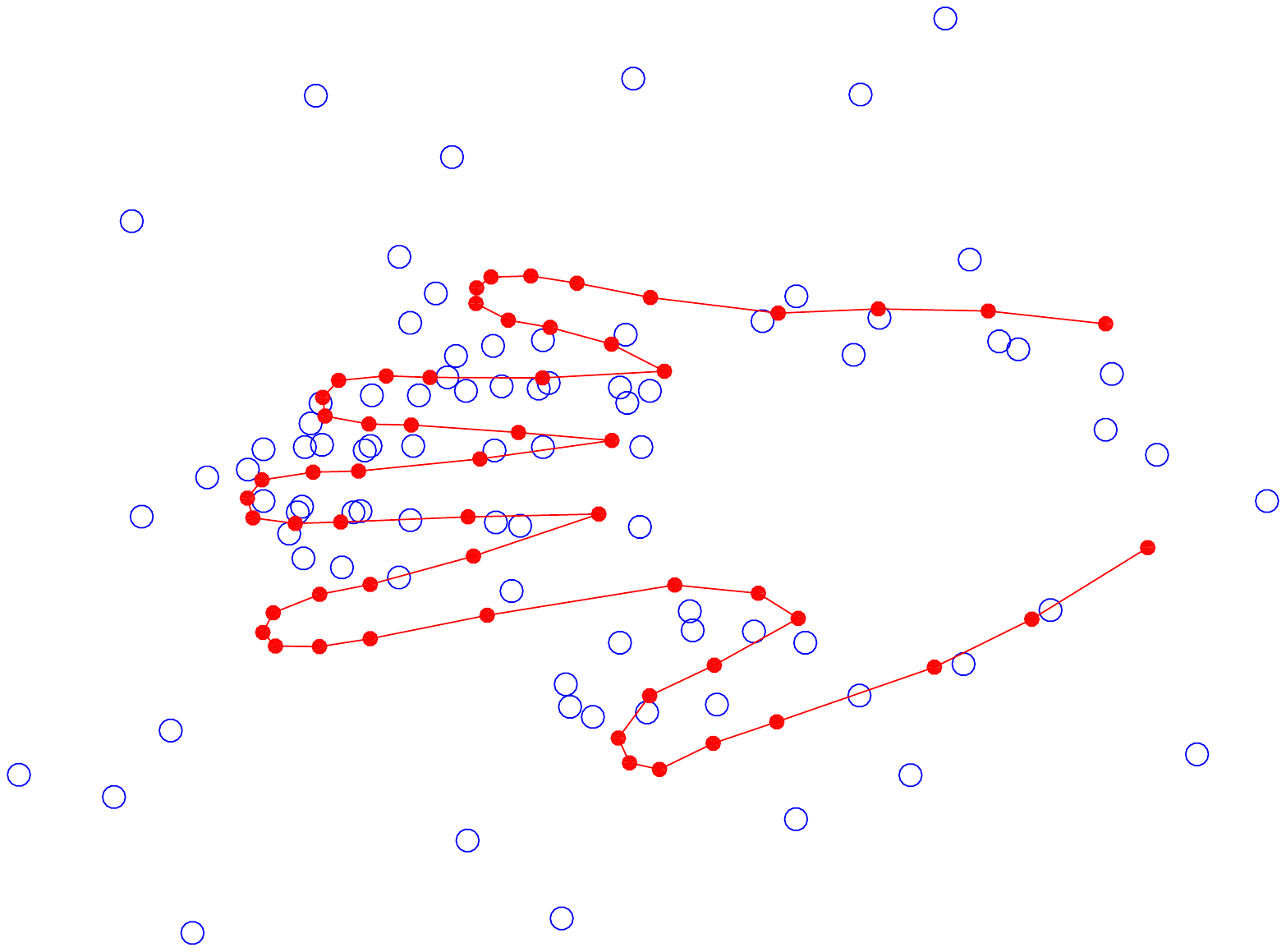} 
             \hspace{-1.2cm} \includegraphics[width=0.26\textwidth]{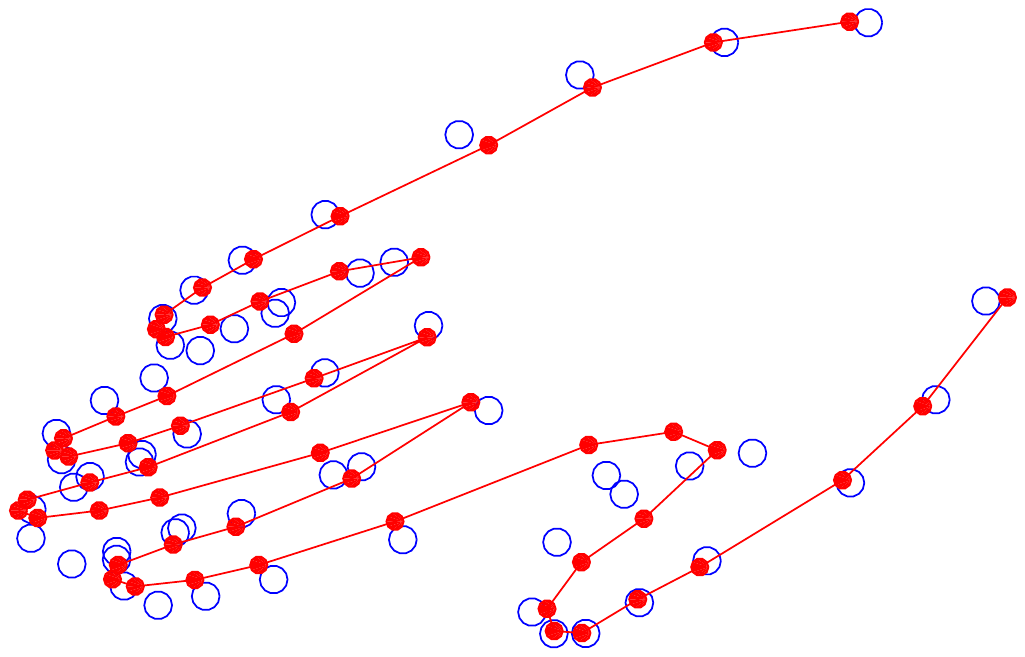} 
             \vspace{-3.0cm}\\% % 
         \raisebox{ 12.0\totalheight} {\bf TPS-RPM     }  
           & \hspace{-0.5cm} \includegraphics[width=0.26\textwidth]{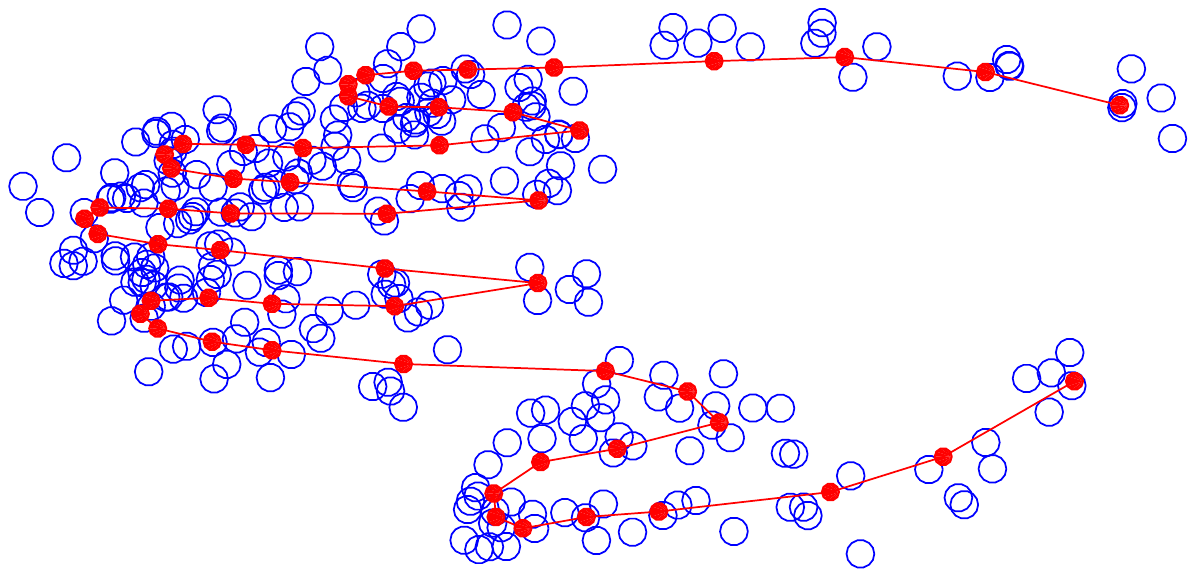} 
             \hspace{-1.2cm} \includegraphics[width=0.26\textwidth]{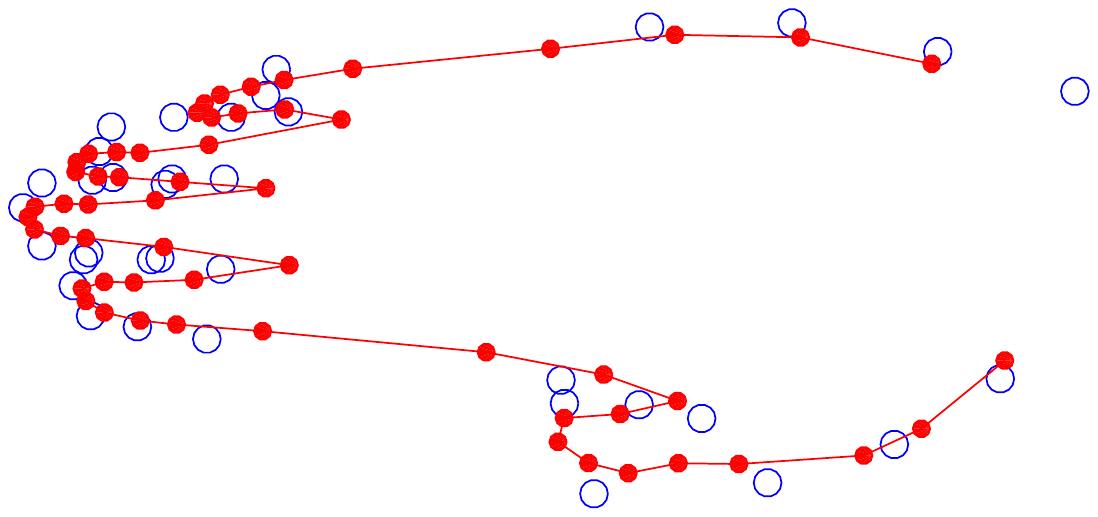} 
             \hspace{-1.2cm} \includegraphics[width=0.26\textwidth]{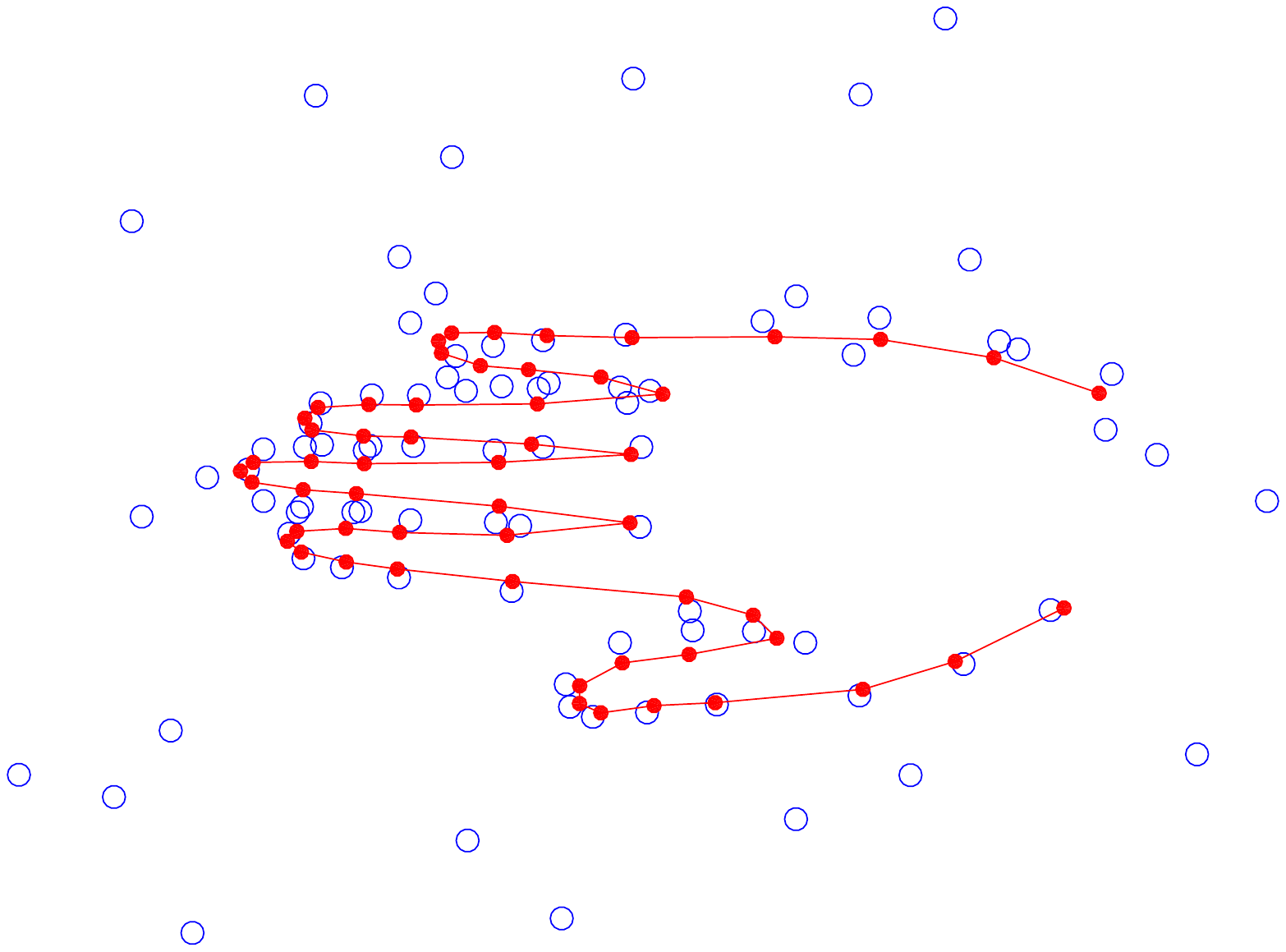} 
             \hspace{-1.2cm} \includegraphics[width=0.26\textwidth]{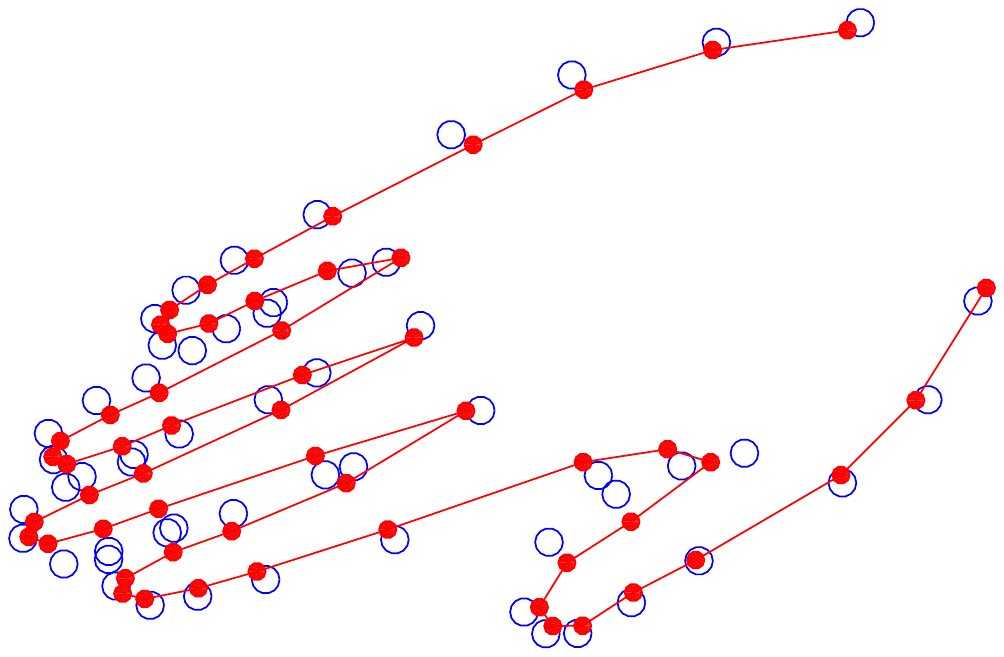} 
       \end{tabular}
    \end{center} \vspace{-1.5cm}
    \caption{ 
      Applications of DLD, CPD, and TPS-RPM to hand No.$6$ with four types of modification: 
      (1) replication of points with dispersion, (2) deletion of points, (3) addition of outliers, 
      and (4) rotation of the whole shape.  
    }
    \label{fig:demo}
  \end{figure*}
\else
  \begin{figure*}
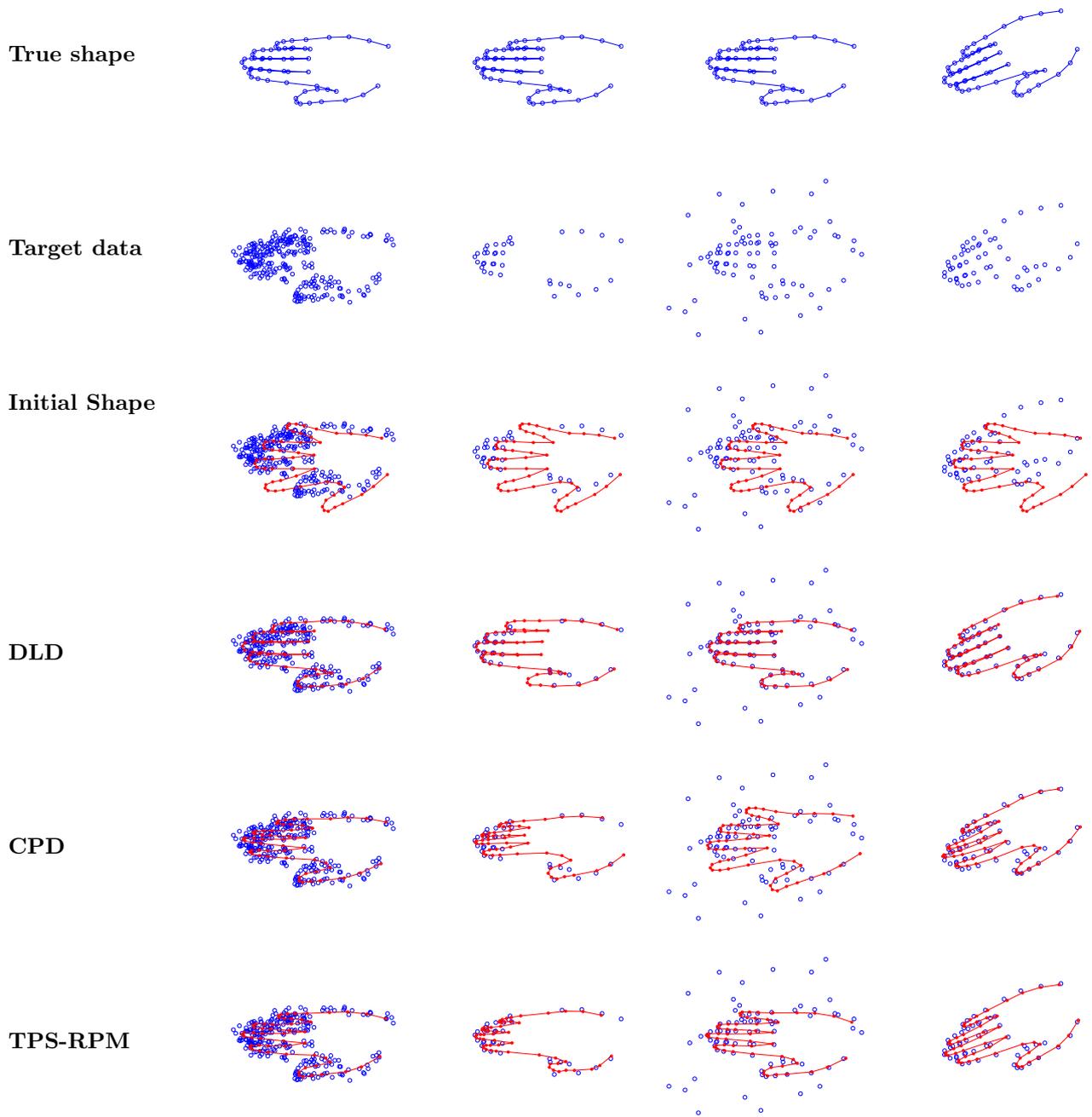

    \begin{center} %\vspace{-3.5cm}
       %\begin{tabular}{ccccc}  & Amplification & Missing & Outlier & Rotation  \end{tabular}
       \begin{tabular}{ll}
         \raisebox{ 10.0\totalheight}{\bf True shape} 
           & \hspace{-0.5cm} \includegraphics[width=0.27\textwidth]{figure/R1/Demo/hand006.pdf}    
             \hspace{-1.2cm} \includegraphics[width=0.27\textwidth]{figure/R1/Demo/hand006.pdf}    
             \hspace{-1.2cm} \includegraphics[width=0.27\textwidth]{figure/R1/Demo/hand006.pdf}    
             \hspace{-1.2cm} \includegraphics[width=0.27\textwidth]{figure/R1/Demo/hand006-rotation030-true.pdf}
             \vspace{-3.0cm}\\% &
         \raisebox{ 10.0\totalheight} {\bf Target data}   
           & \hspace{-0.5cm} \includegraphics[width=0.27\textwidth]{figure/R1/Demo/hand006-amplify-x5-ID001-data.pdf}    
             \hspace{-1.2cm} \includegraphics[width=0.27\textwidth]{figure/R1/Demo/hand006-missing020-ID001-data.pdf}    
             \hspace{-1.2cm} \includegraphics[width=0.27\textwidth]{figure/R1/Demo/hand006-outlier028-ID001-data.pdf}    
             \hspace{-1.2cm} \includegraphics[width=0.27\textwidth]{figure/R1/Demo/hand006-rotation030-ID001-data.pdf}      
             \vspace{-3.0cm}\\% &
         \raisebox{ 12.0\totalheight} {\bf Initial Shape}  
           & \hspace{-0.5cm} \includegraphics[width=0.27\textwidth]{figure/R1/Demo/hand006-amplify-x5-ID001-before.pdf} 
             \hspace{-1.2cm} \includegraphics[width=0.27\textwidth]{figure/R1/Demo/hand006-missing020-ID001-before.pdf} 
             \hspace{-1.2cm} \includegraphics[width=0.27\textwidth]{figure/R1/Demo/hand006-outlier028-ID001-before.pdf} 
             \hspace{-1.2cm} \includegraphics[width=0.27\textwidth]{figure/R1/Demo/hand006-rotation030-ID001-before.pdf}    
             \vspace{-3.0cm}\\ % &
         \raisebox{ 12.0\totalheight} {\bf DLD         }  
           & \hspace{-0.5cm} \includegraphics[width=0.27\textwidth]{figure/R1/Demo/hand006-amplify-x5-ID001-after-dld.pdf}
             \hspace{-1.2cm} \includegraphics[width=0.27\textwidth]{figure/R1/Demo/hand006-missing020-ID001-after-dld.pdf}
             \hspace{-1.2cm} \includegraphics[width=0.27\textwidth]{figure/R1/Demo/hand006-outlier028-ID001-after-dld.pdf}
             \hspace{-1.2cm} \includegraphics[width=0.27\textwidth]{figure/R1/Demo/hand006-rotation030-ID001-after-dld.pdf} 
             \vspace{-3.0cm}\\% &
         \raisebox{ 12.0\totalheight} {\bf CPD         }  
           & \hspace{-0.5cm} \includegraphics[width=0.27\textwidth]{figure/R1/Demo/hand006-amplify-x5-ID001-after-cpd.pdf} 
             \hspace{-1.2cm} \includegraphics[width=0.27\textwidth]{figure/R1/Demo/hand006-missing020-ID001-after-cpd.pdf} 
             \hspace{-1.2cm} \includegraphics[width=0.27\textwidth]{figure/R1/Demo/hand006-outlier028-ID001-after-cpd.pdf} 
             \hspace{-1.2cm} \includegraphics[width=0.27\textwidth]{figure/R1/Demo/hand006-rotation030-ID001-after-cpd.pdf} 
             \vspace{-3.0cm}\\% % 
         \raisebox{ 12.0\totalheight} {\bf TPS-RPM     }  
           & \hspace{-0.5cm} \includegraphics[width=0.27\textwidth]{figure/R1/Demo/hand006-amplify-x5-ID001-after-tps.pdf} 
             \hspace{-1.2cm} \includegraphics[width=0.27\textwidth]{figure/R1/Demo/hand006-missing020-ID001-after-tps.pdf} 
             \hspace{-1.2cm} \includegraphics[width=0.27\textwidth]{figure/R1/Demo/hand006-outlier028-ID001-after-tps.pdf} 
             \hspace{-1.2cm} \includegraphics[width=0.27\textwidth]{figure/R1/Demo/hand006-rotation030-ID001-after-tps.pdf} 
       \end{tabular}
    \end{center} \vspace{-1.5cm}
    \caption{ 
      Applications of DLD, CPD, and TPS-RPM to hand No.$6$ with four types of modification: 
      (1) replication of points with dispersion, (2) deletion of points, (3) addition of outliers, 
      and (4) rotation of the whole shape.  
    }
    \label{fig:demo}
  \end{figure*}
\fi

To test the robustness of our approach, we generated four types of target point data
for hand No. $6$ in the IMM hand data.
The top row in Figure \ref{fig:demo} shows the correct target shapes, i.e., hand No. $6$ in the IMM dataset,
and the second row shows target point sets, which represent hand No. $6$ with four types of modifications:
(a) replication of target points with dispersion,
(b) deletion of target points, 
(c) addition of outliers, and 
(d) rotation of the whole shape.
Here, we used hand No. $6$ selected from the original dataset as a validation point set
instead of the pre-processed one for training a statistical shape model, 
to prevent the test problem from becoming too easy. 
The red points shown in figures of the 
third row are the initial hand shapes to be deformed, i.e., the mean shape of the hands, 
before optimization. 

\subsubsection*{Choice of parameters}
For all four target point sets, we fixed $K$ and $\gamma$, which are the parameters of the DLD, 
to $K=10$ and $\gamma=10^{-3}$.
On the contrary, we used the remaining parameter $\omega=0.01$ for (a), (b), and (d)
and $\omega=0.30$ for (c) as outliers were included in (c), but not included in (a), (b), and (d).
For CPD and TPS-RPM, we used the software distributed by the authors of CPD and TPS-RPM with 
their default parameters. For the DLD, we used 39 hand shapes as training data with target hand shape
No. $6$ removed. 

\subsubsection*{Results}

The fourth, fifth, and sixth rows in Figure \ref{fig:demo} show the results of registrations 
using the DLD, CPD, and TPS-RPM algorithms, respectively. The DLD yielded the best registration for all data.
For all hand shapes deformed by CPD and TPS-RPM, images of all fingers except the thumb 
were rendered thinner than those of the true target shape, suggesting the weakness in assuming only
a smooth displacement field as prior shape information:
a point was forced to be displaced coherently with its neighboring points, even for a different finger.
For data (a), all methods roughly recovered the true target shape,
whereas thinner fingers were observed for the CPD and TPS-RPM.
For data (b), the DLD recovered the missing points correctly. On the contrary, the CPD and TPS-RPM did not recover 
the missing points, and the thumb and all fingers were rendered shorter in their results 
than in the true target shapes. This suggests a weakness in the computation of the missing regions. 
This weakness originates from assuming only a smooth displacement field: 
source points moved much more coherently as the target points did not exist near them.
For data (c), the DLD found the true target shape,
whereas the deformed shapes generated by CPD and TPS-RPM were close to the mean shape rather than the 
true target shape.
For data (d), all methods roughly recovered the true target shape, whereas thinner fingers were
observed in results for the CPD and TPS-RPM. The DLD succeeded in recovering the true hand shape.

\subsection{Comparison of registration performance in 2D cases}

\ifDCOL
  \begin{figure*}
    \begin{center} \vspace{-1.5cm}
       %\begin{tabular}{ccccc}  & Amplification & Missing & Outlier & Rotation  \end{tabular}
       \begin{tabular}{ll}
         \raisebox{ 10.0\totalheight} {\bf Target data} 
           & \hspace{-0.7cm} \includegraphics[width=0.30\textwidth]{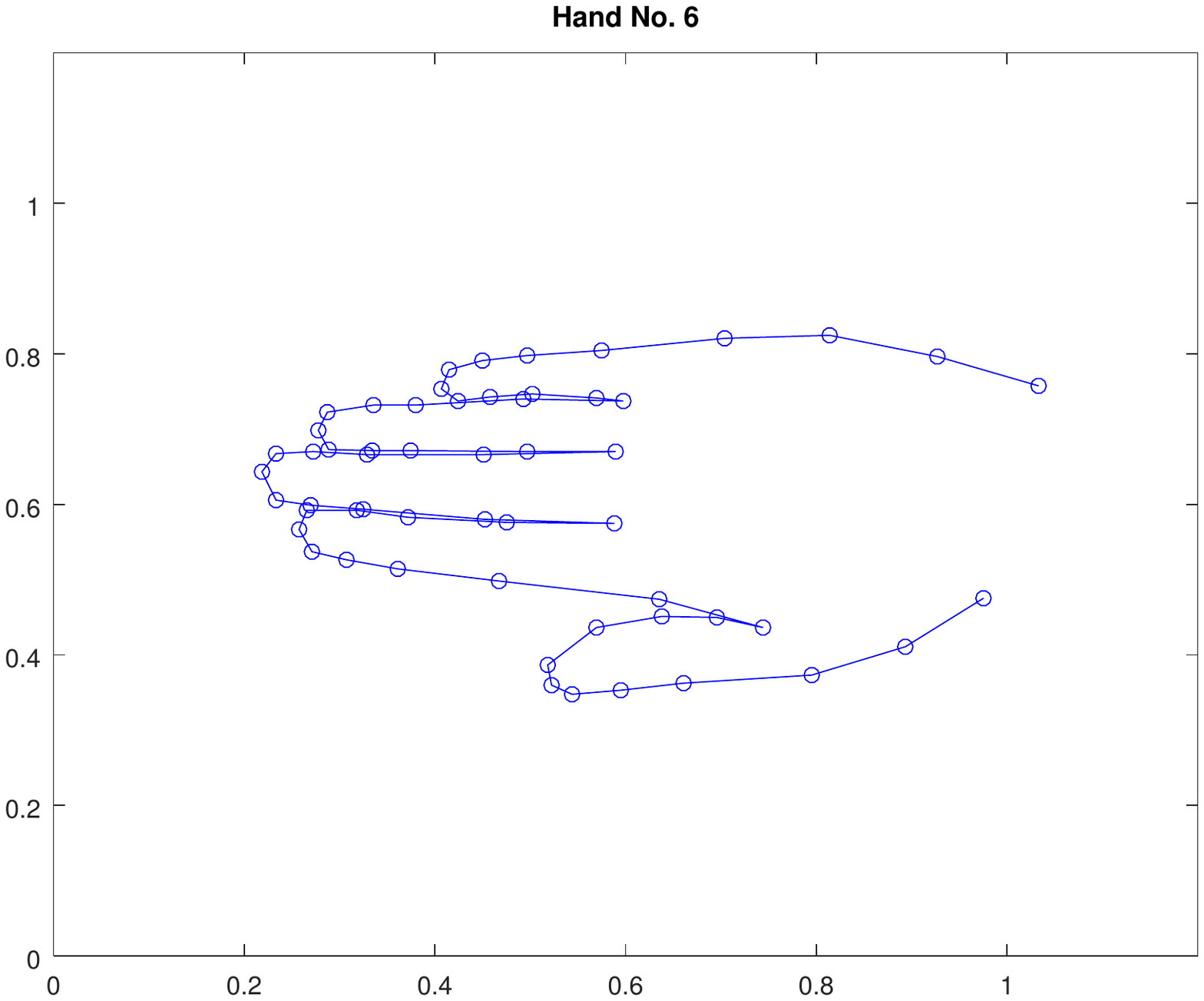}    
             \hspace{-0.7cm} \includegraphics[width=0.30\textwidth]{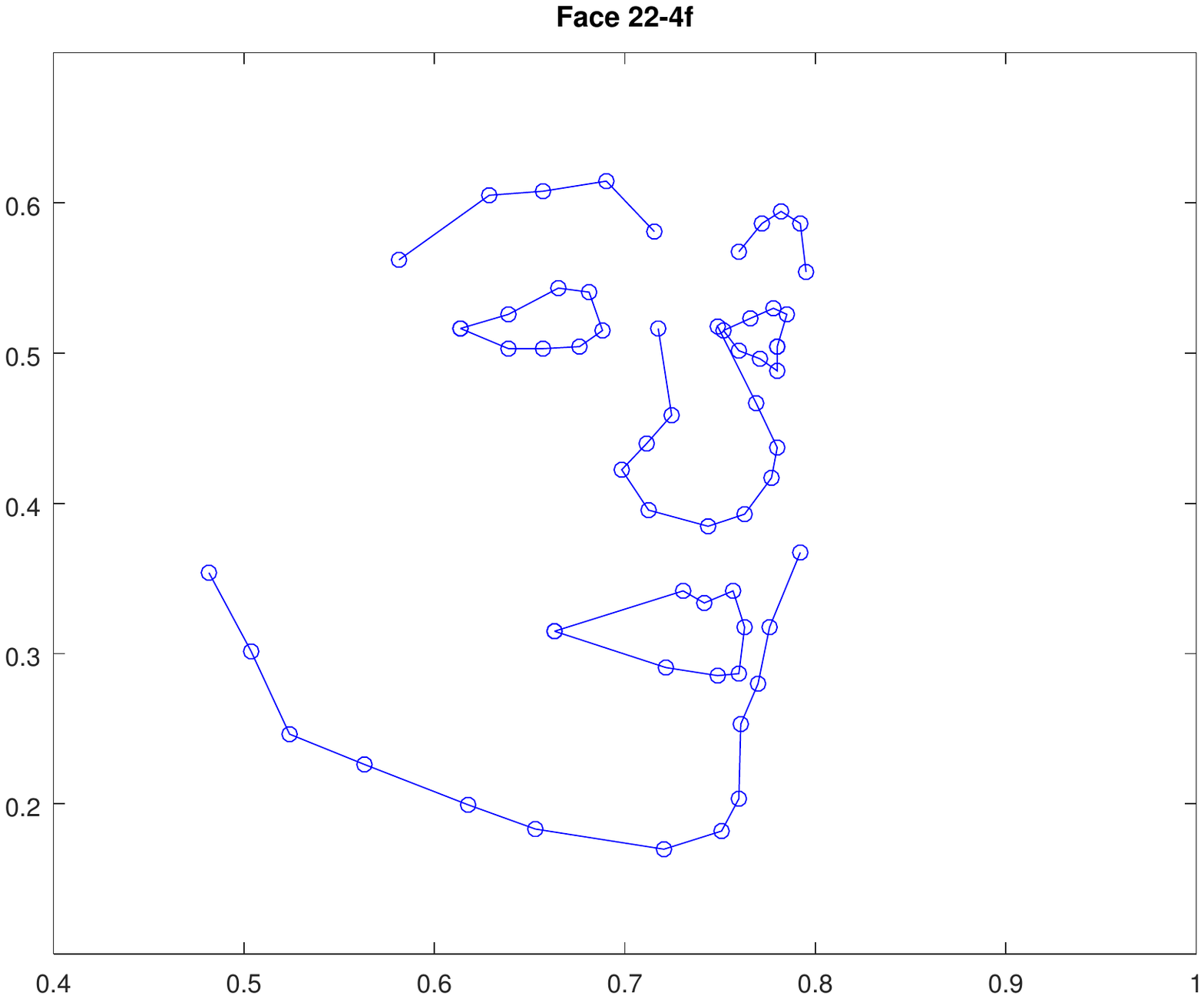}    
             \vspace{-3.0cm}\\% &
         \raisebox{ 10.0\totalheight} {\bf Replication}   
           & \hspace{-0.7cm} \includegraphics[width=0.30\textwidth]{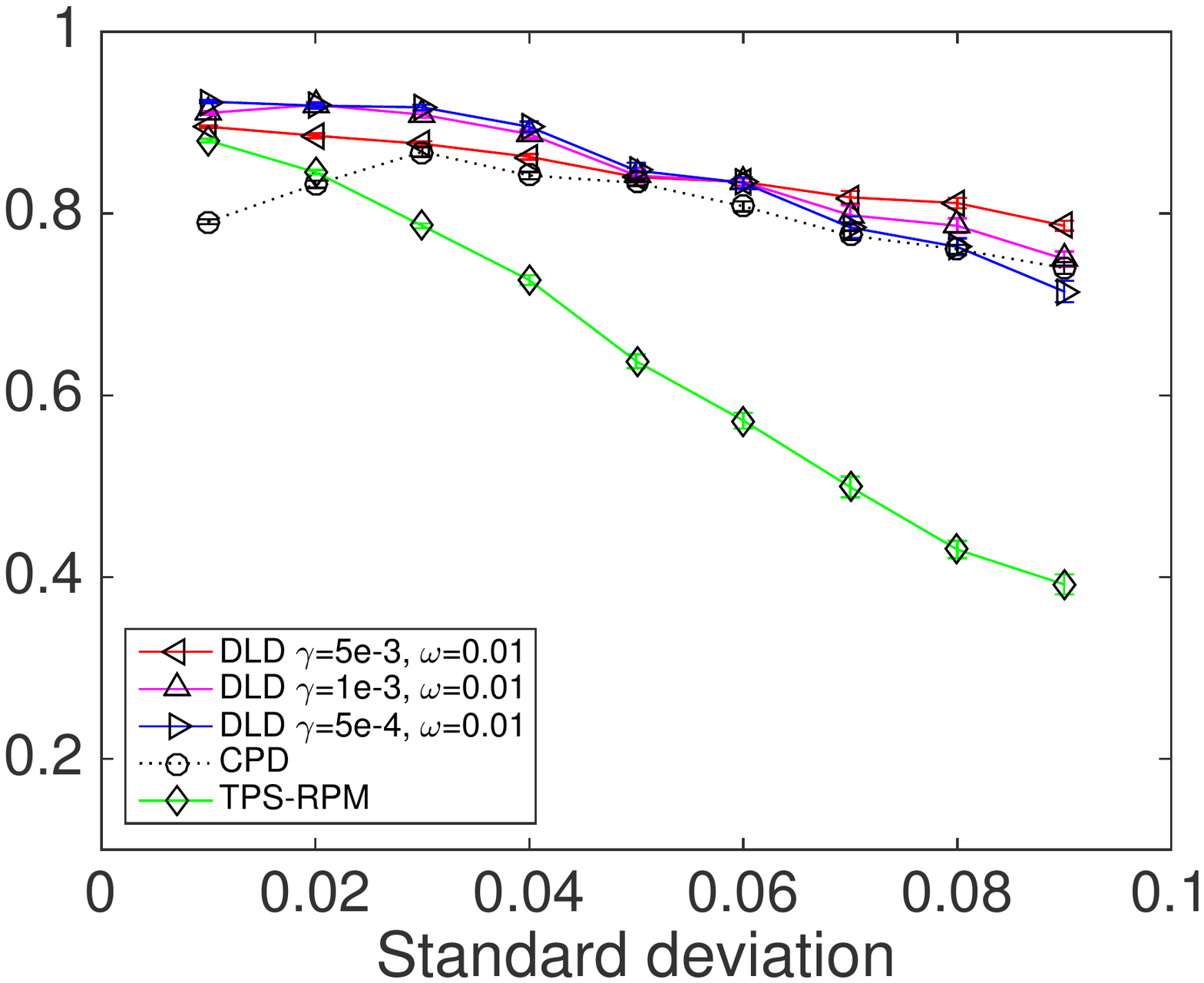}    
             \hspace{-0.7cm} \includegraphics[width=0.30\textwidth]{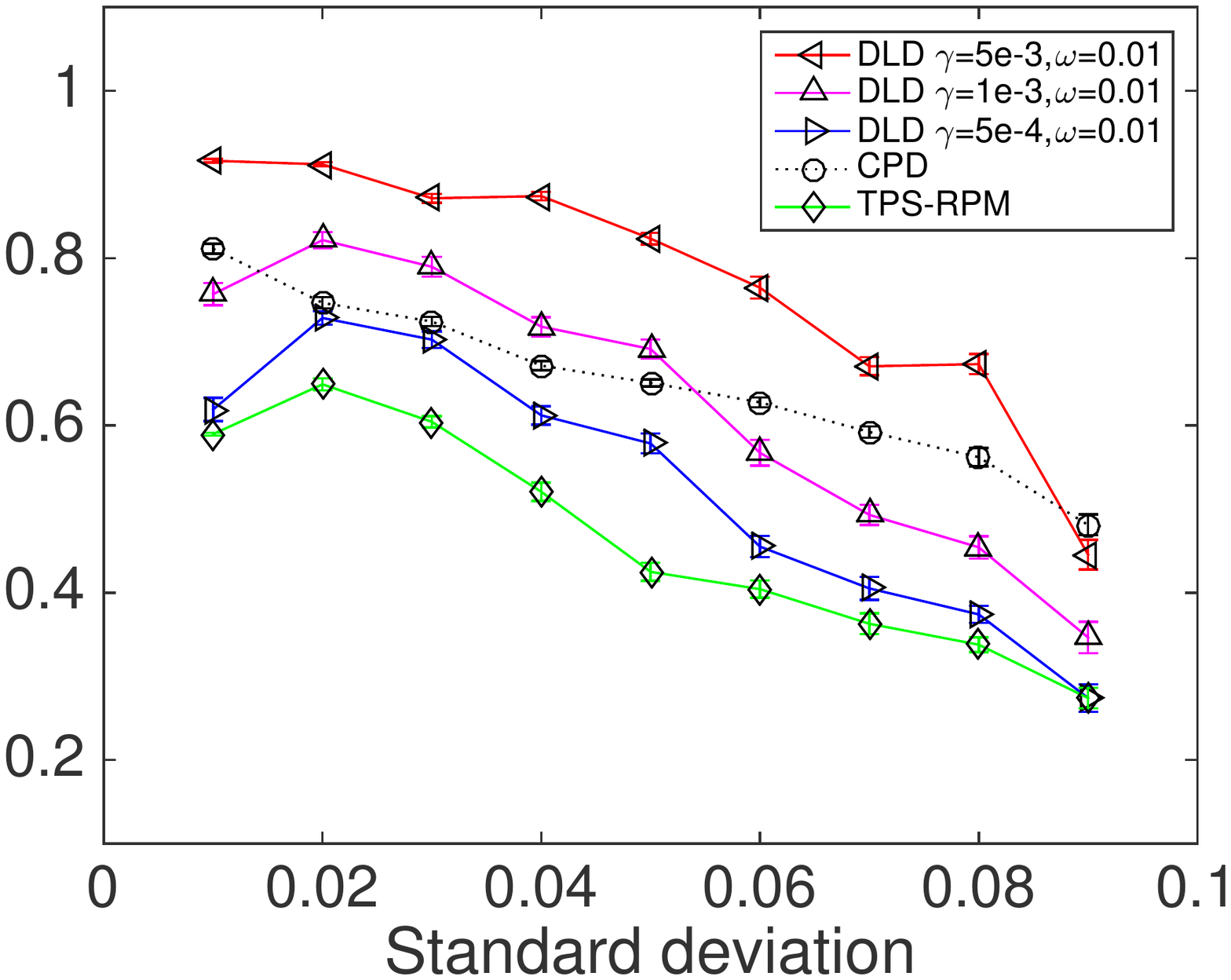}    
             \vspace{-3.0cm}\\% &
         \raisebox{ 12.0\totalheight} {\bf Deletion}  
           & \hspace{-0.7cm} \includegraphics[width=0.30\textwidth]{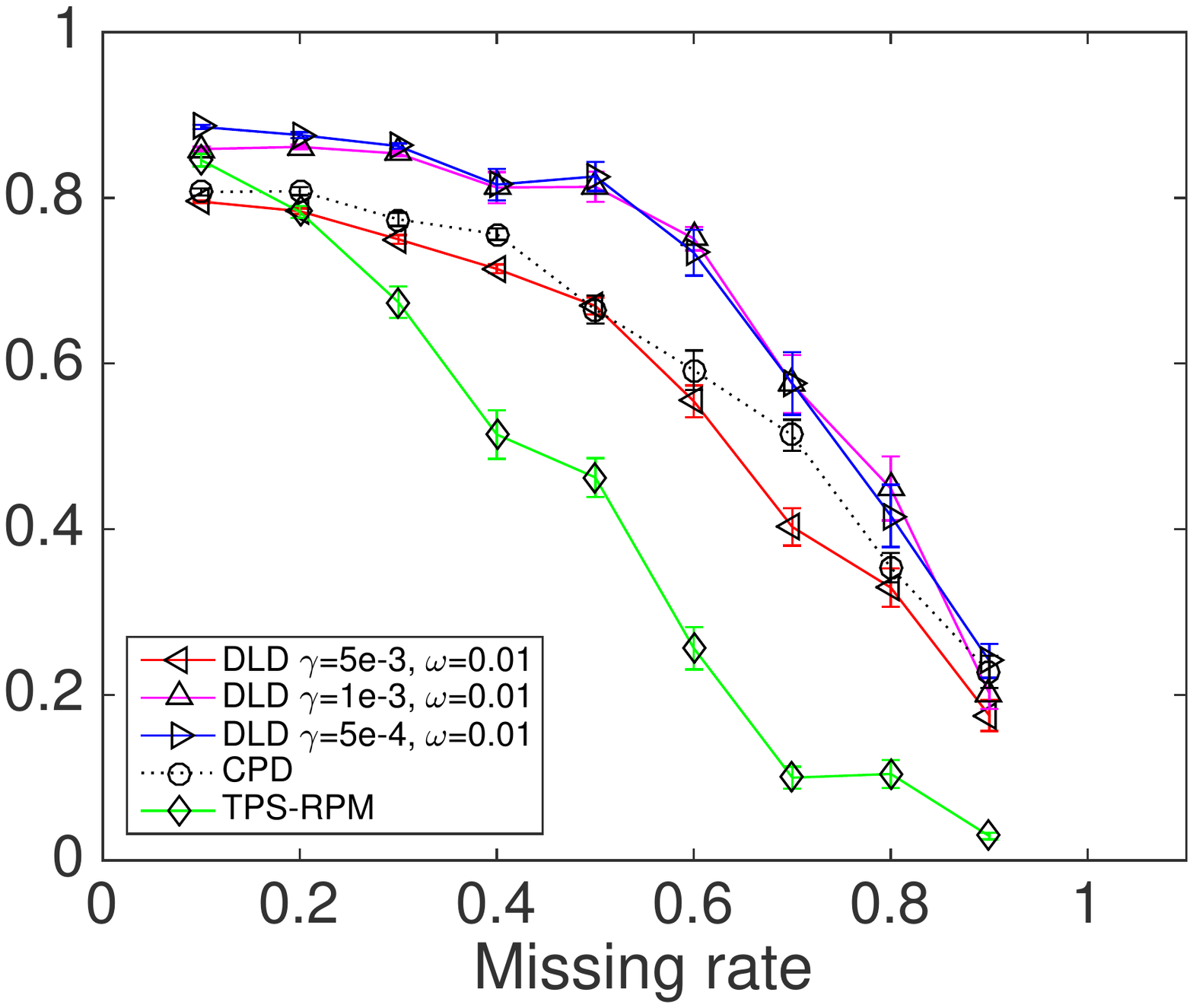} 
             \hspace{-0.7cm} \includegraphics[width=0.30\textwidth]{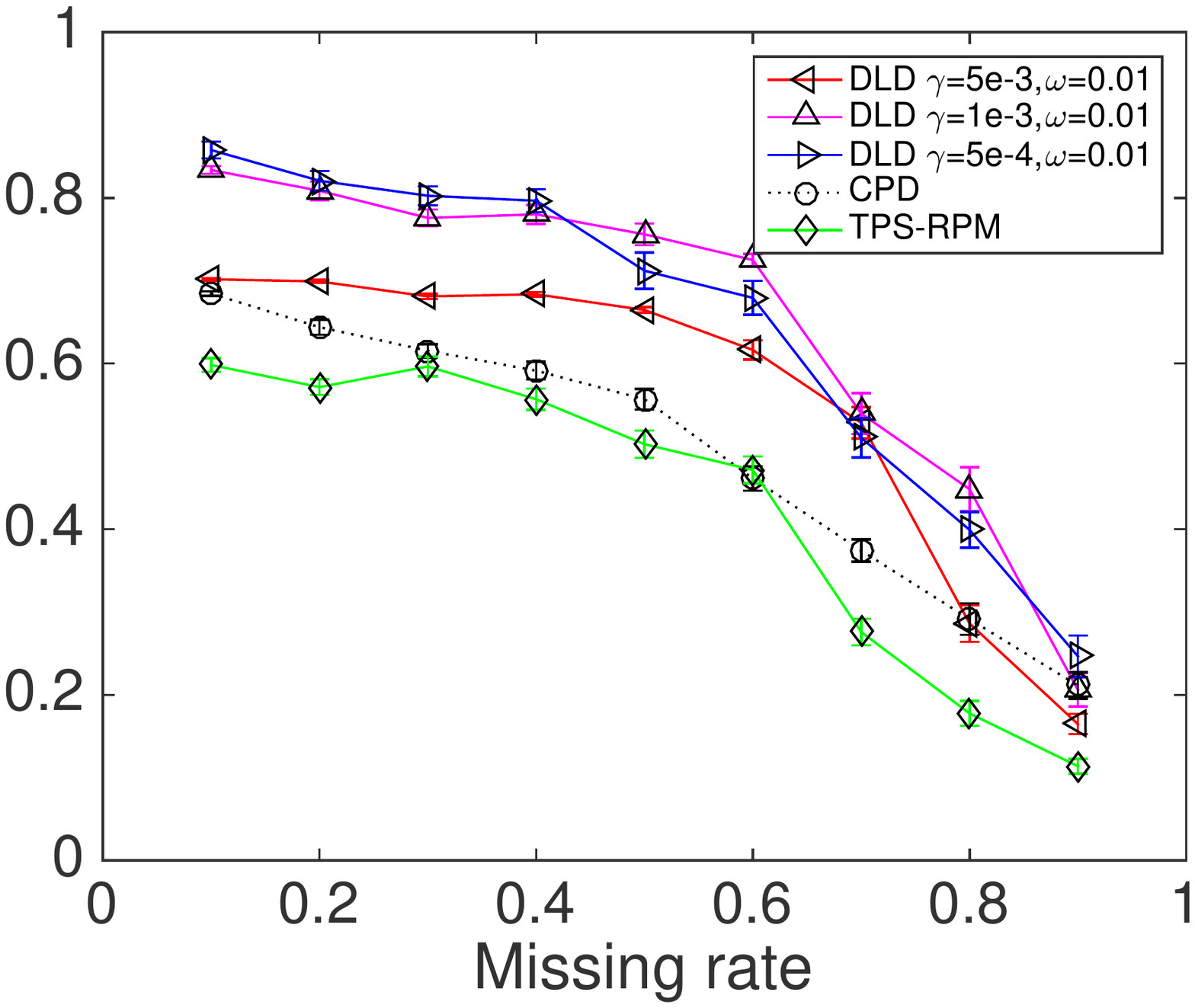} 
             \vspace{-3.0cm}\\ % &
         \raisebox{ 12.0\totalheight} {\bf Outliers}  
           & \hspace{-0.7cm} \includegraphics[width=0.30\textwidth]{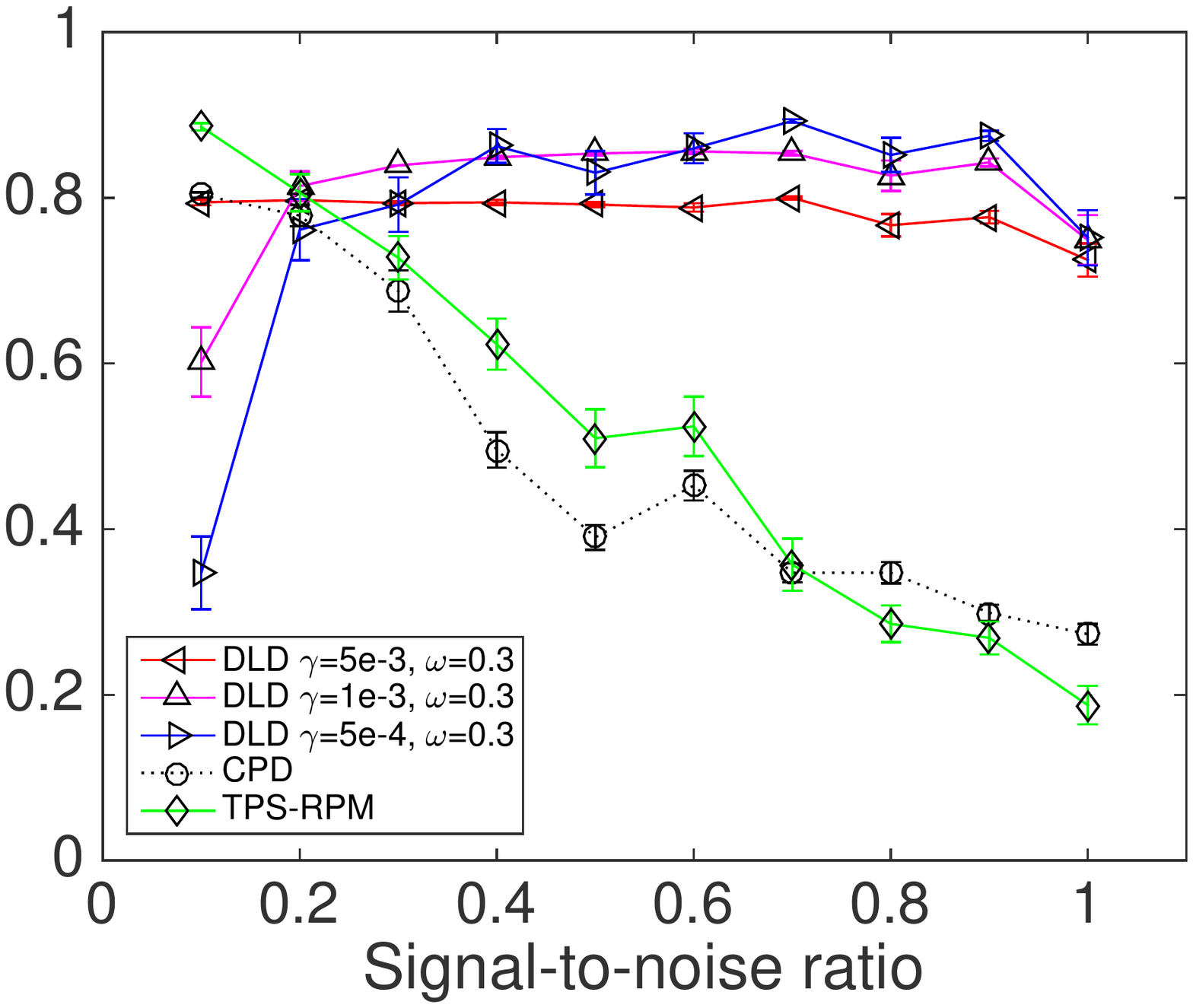}
             \hspace{-0.7cm} \includegraphics[width=0.30\textwidth]{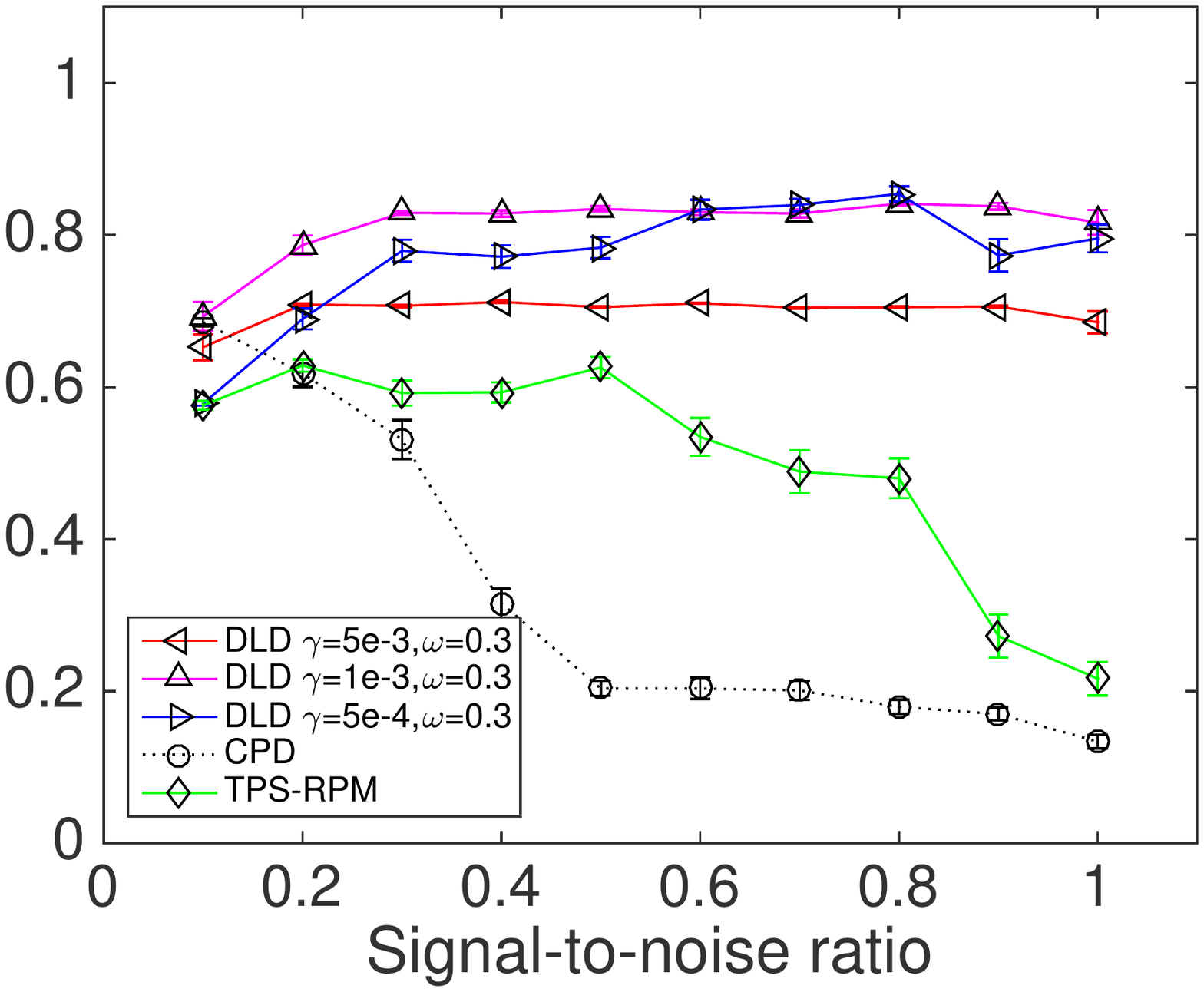}
             \vspace{-3.0cm}\\% &
         \raisebox{ 12.0\totalheight} {\bf Rotation}  
           & \hspace{-0.7cm} \includegraphics[width=0.30\textwidth]{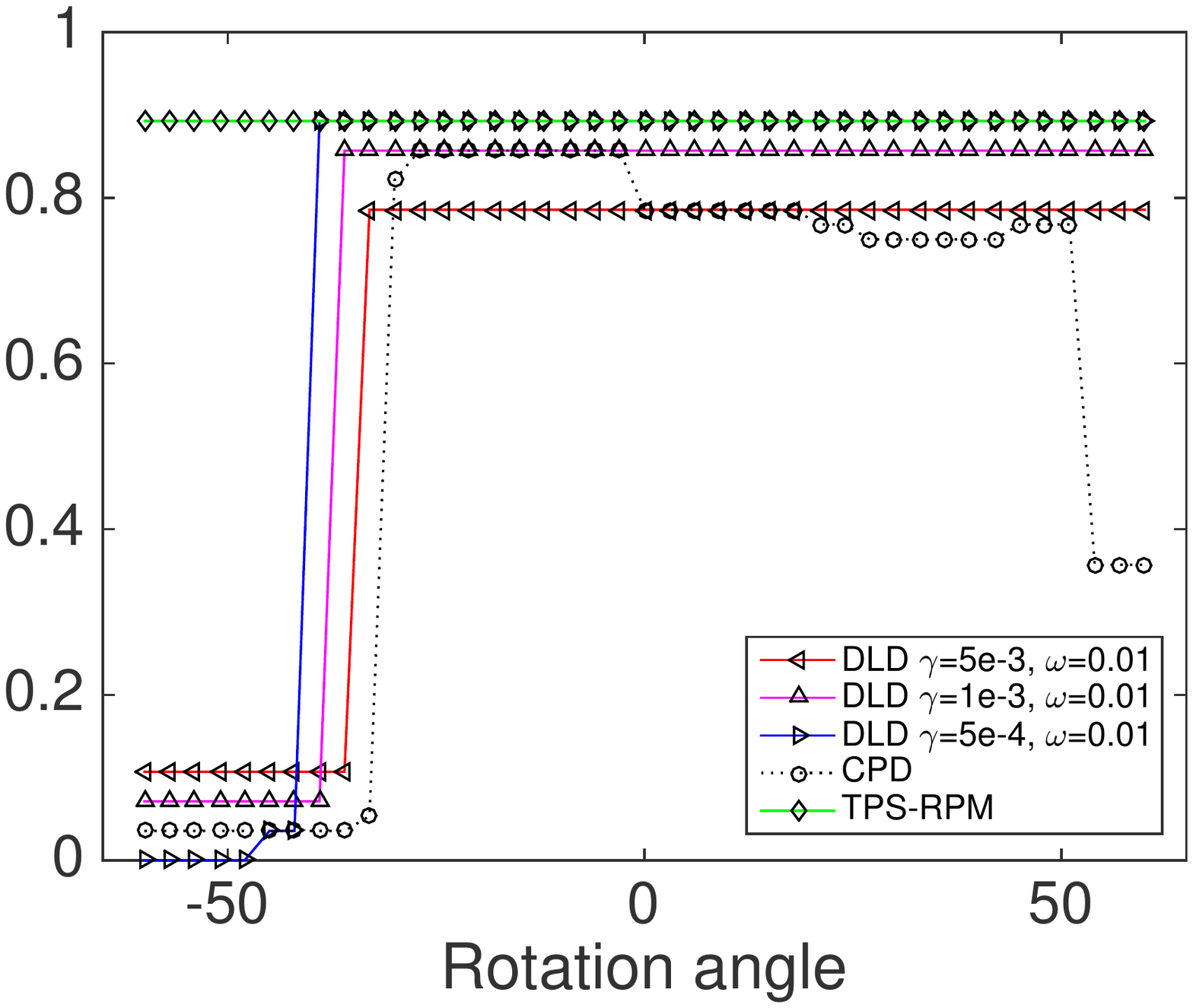} 
             \hspace{-0.7cm} \includegraphics[width=0.30\textwidth]{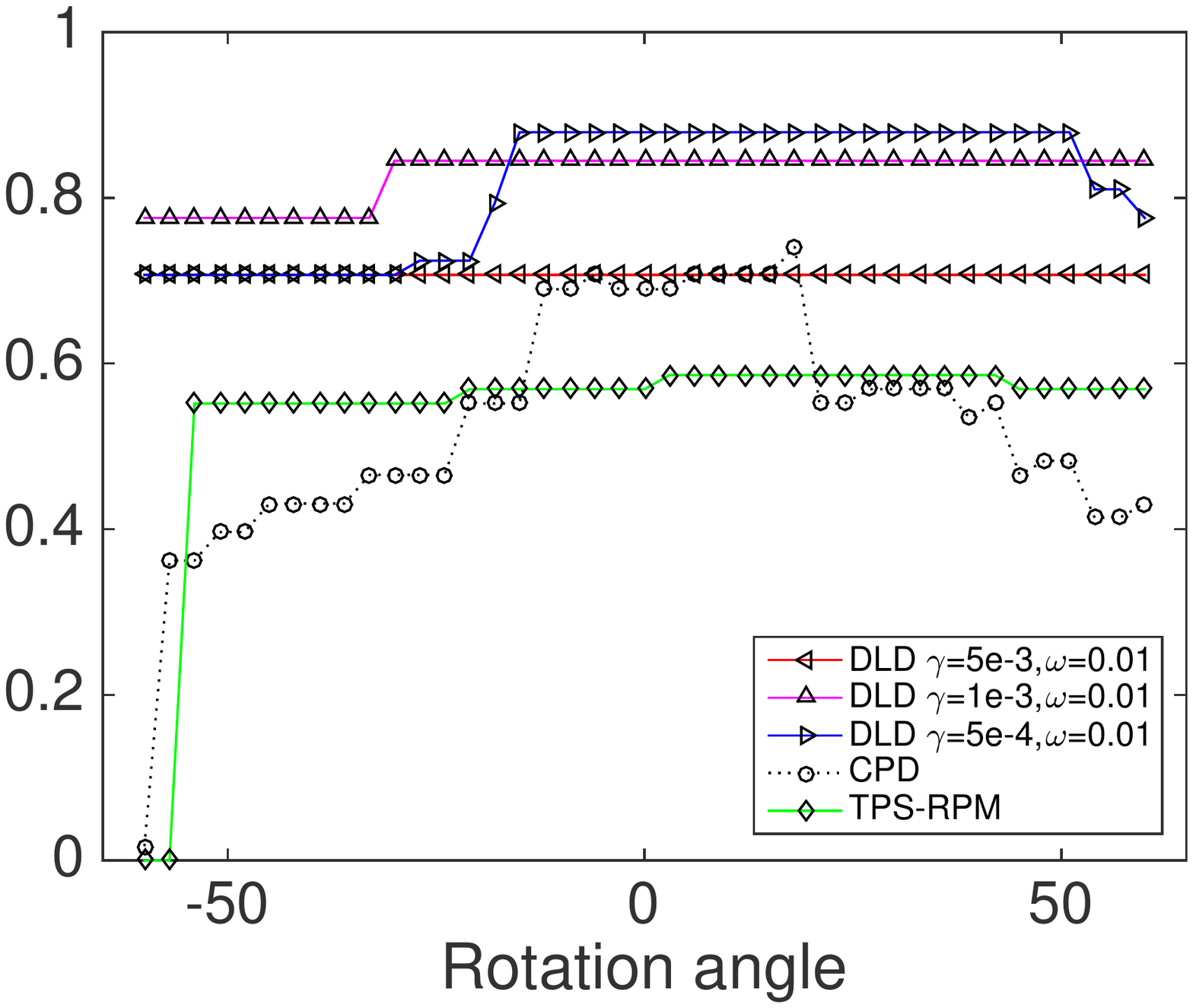} 
       \end{tabular}
    \end{center} \vspace{-1.5cm}
    \caption{ 
      Comparisons of registration performance among DLD, CPD, and TPS-RPM using hand No. $6$ in the IMM hand dataset (left),
      and face 22-4f in the IMM face dataset (right) with four types of modification: (1) replication of points with dispersion,
      (2) deletion of points, (3) addition of outliers, and (4) rotation of the whole shape.
     }
    \label{fig:comp}
  \end{figure*}
\else
  \begin{figure*}
    \begin{center} \vspace{-1.5cm}
       %\begin{tabular}{ccccc}  & Amplification & Missing & Outlier & Rotation  \end{tabular}
       \begin{tabular}{ll}
         \raisebox{ 10.0\totalheight}{\bf Target data} 
           & \hspace{-0.7cm} \includegraphics[width=0.32\textwidth]{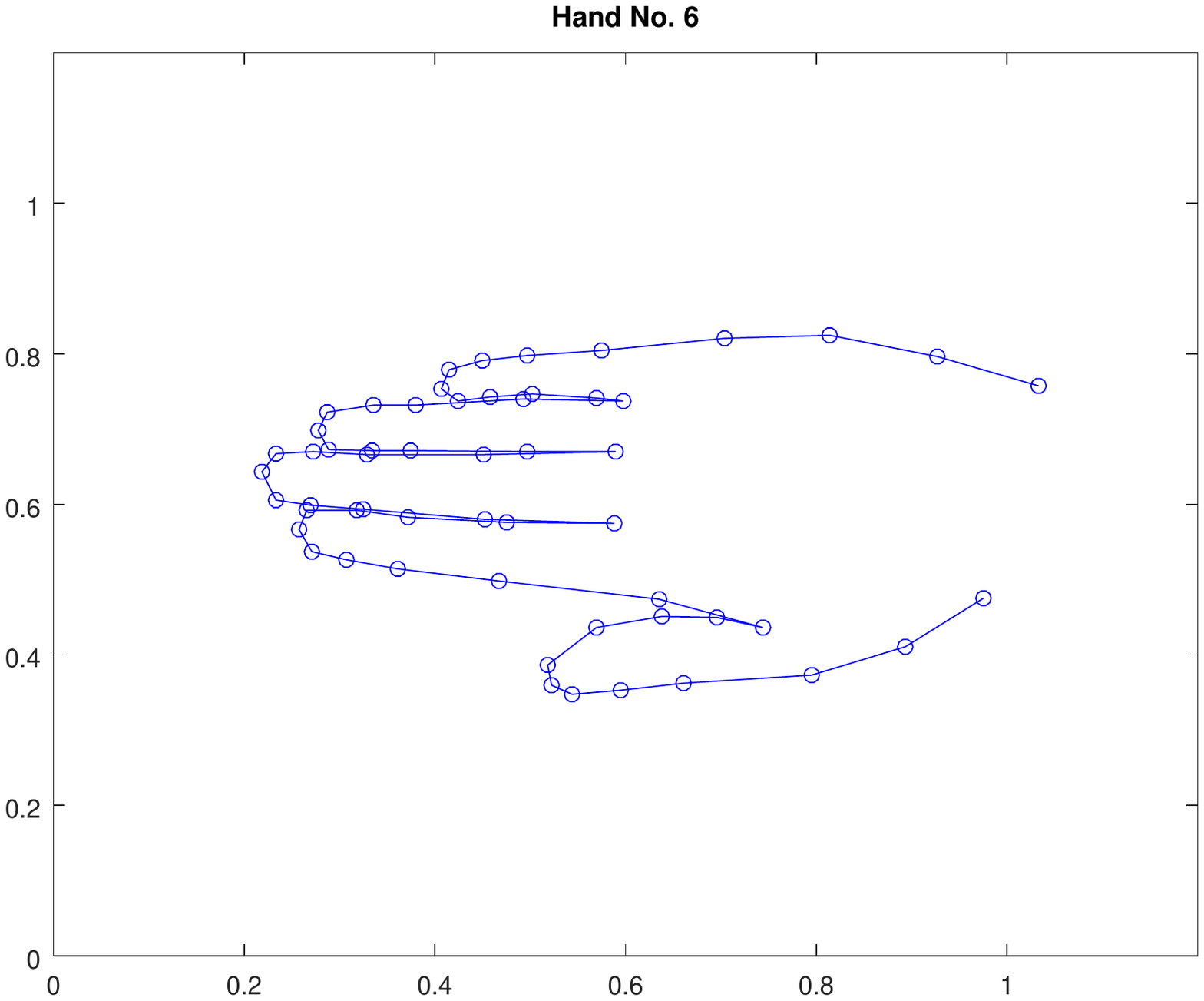}    
             \hspace{-0.7cm} \includegraphics[width=0.32\textwidth]{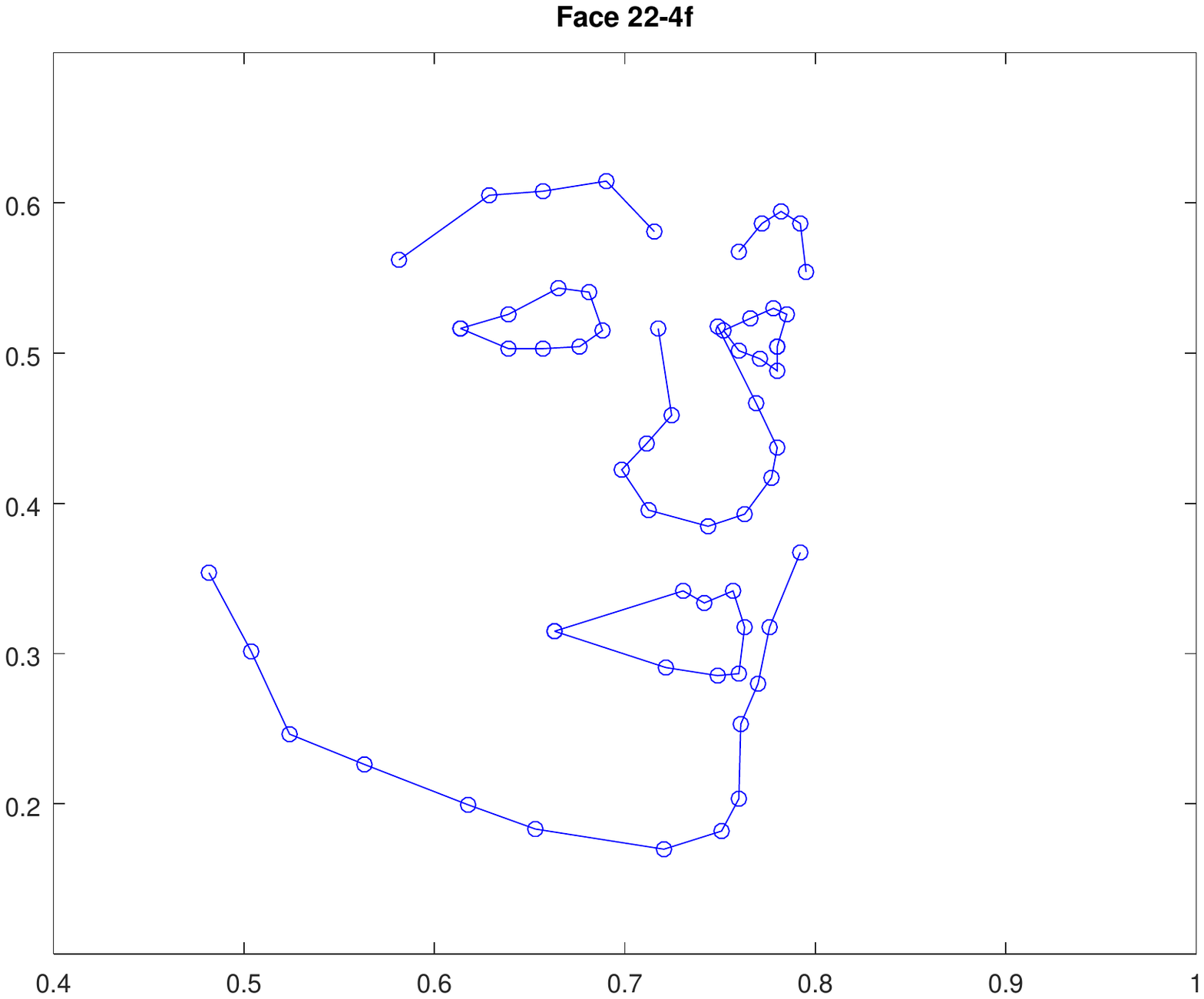}    
             \vspace{-3.0cm}\\% &
         \raisebox{ 10.0\totalheight} {\bf Replication}   
           & \hspace{-0.7cm} \includegraphics[width=0.32\textwidth]{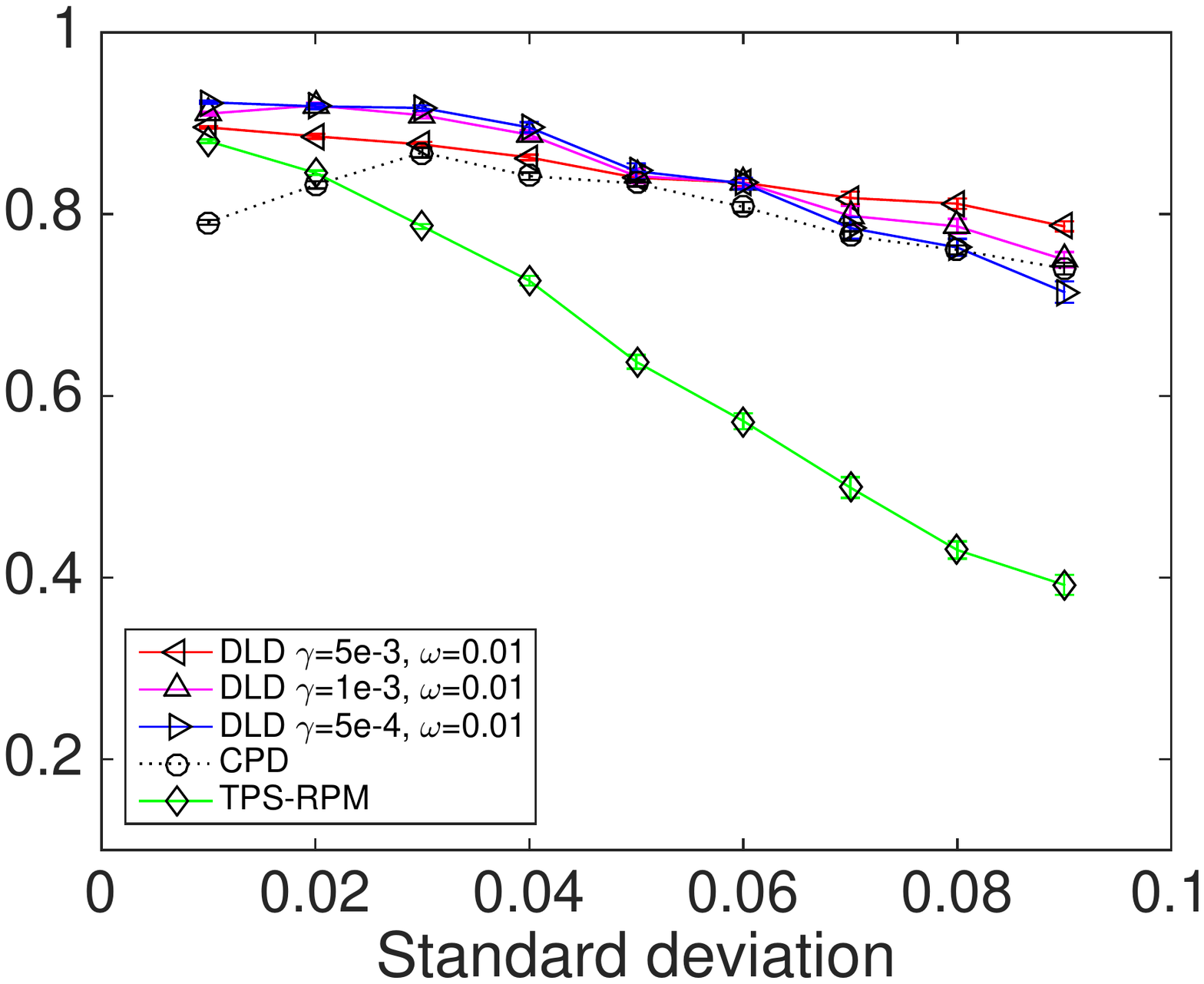}    
             \hspace{-0.7cm} \includegraphics[width=0.32\textwidth]{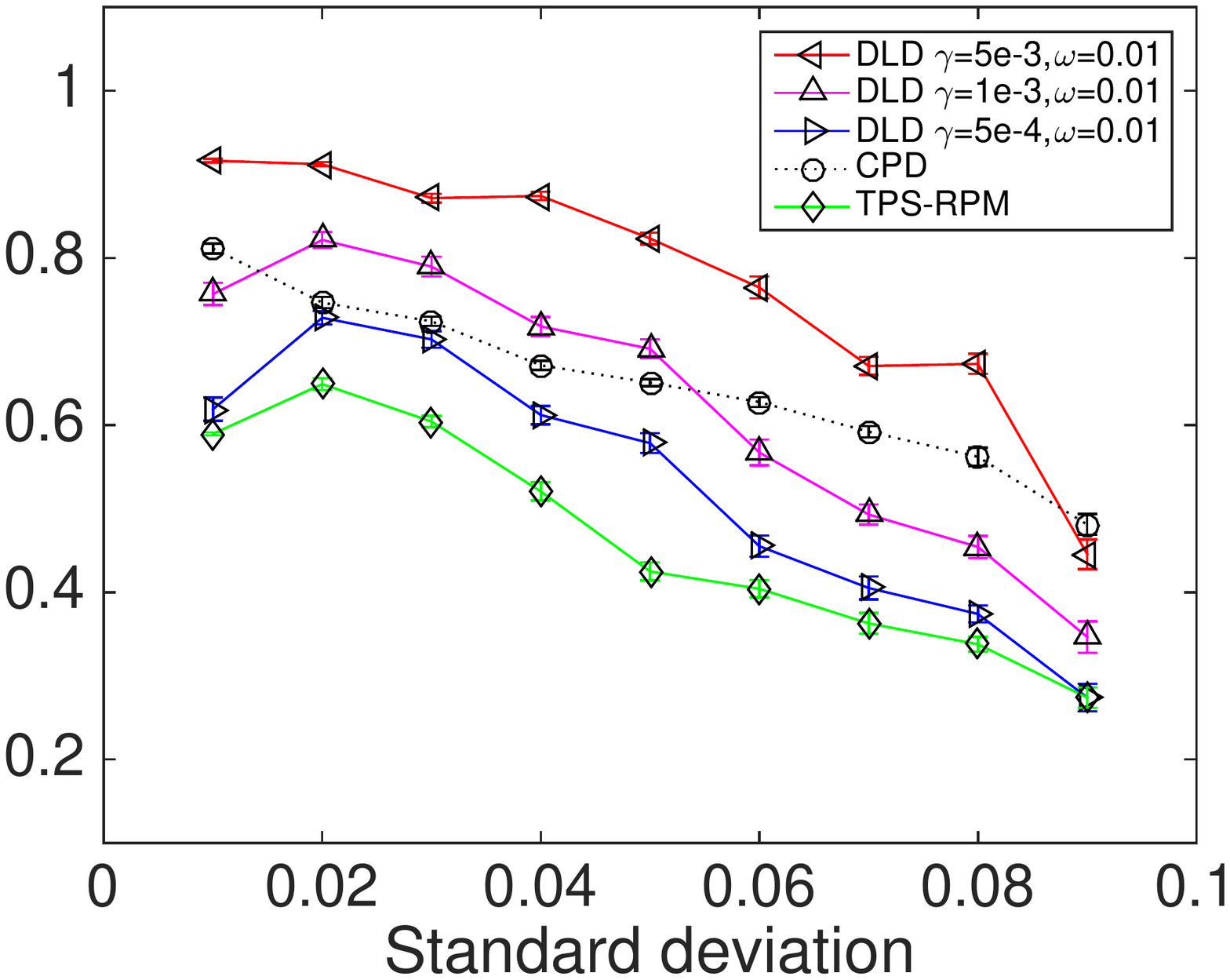}    
             \vspace{-3.0cm}\\% &
         \raisebox{ 12.0\totalheight} {\bf Deletion}  
           & \hspace{-0.7cm} \includegraphics[width=0.32\textwidth]{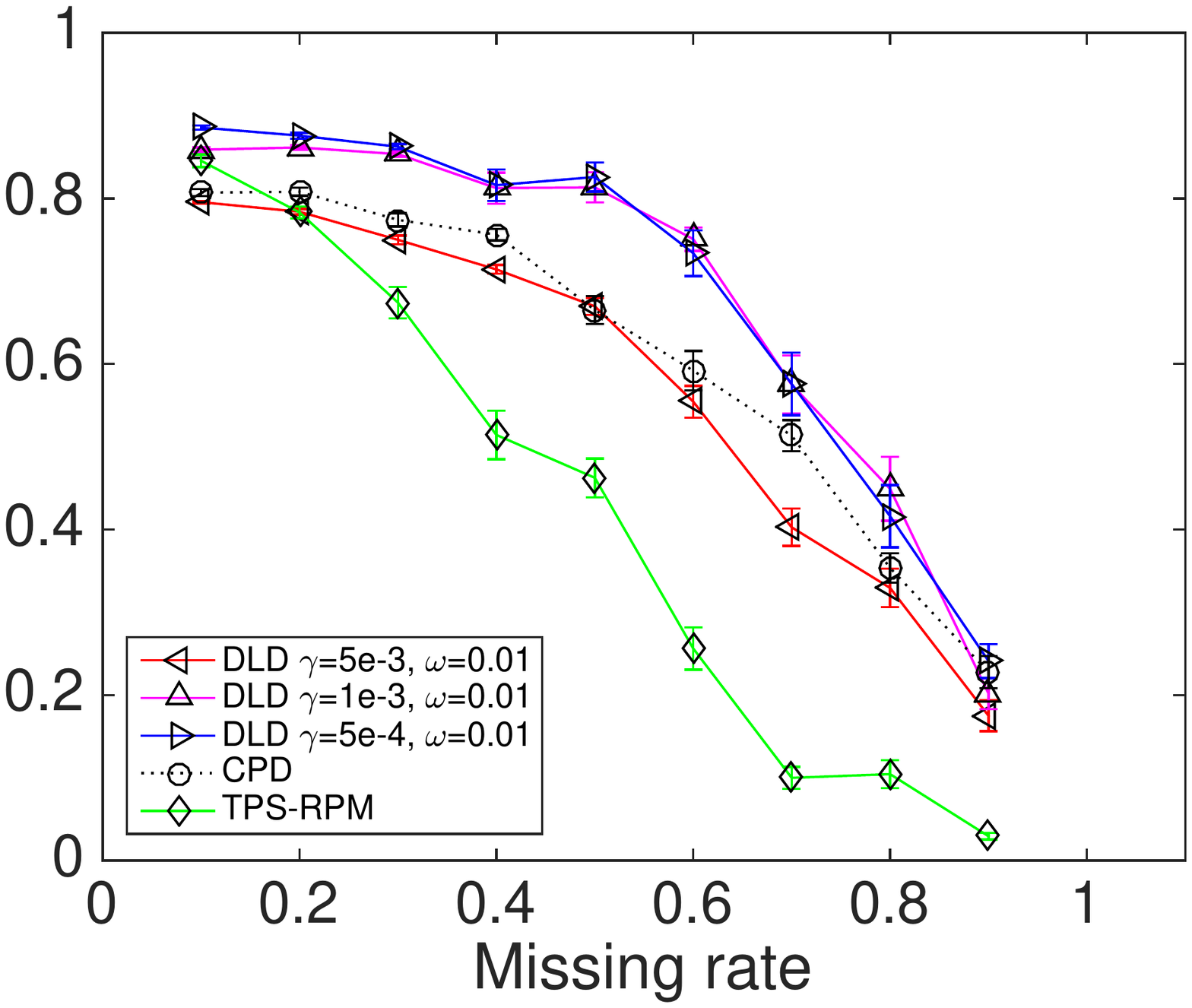} 
             \hspace{-0.7cm} \includegraphics[width=0.32\textwidth]{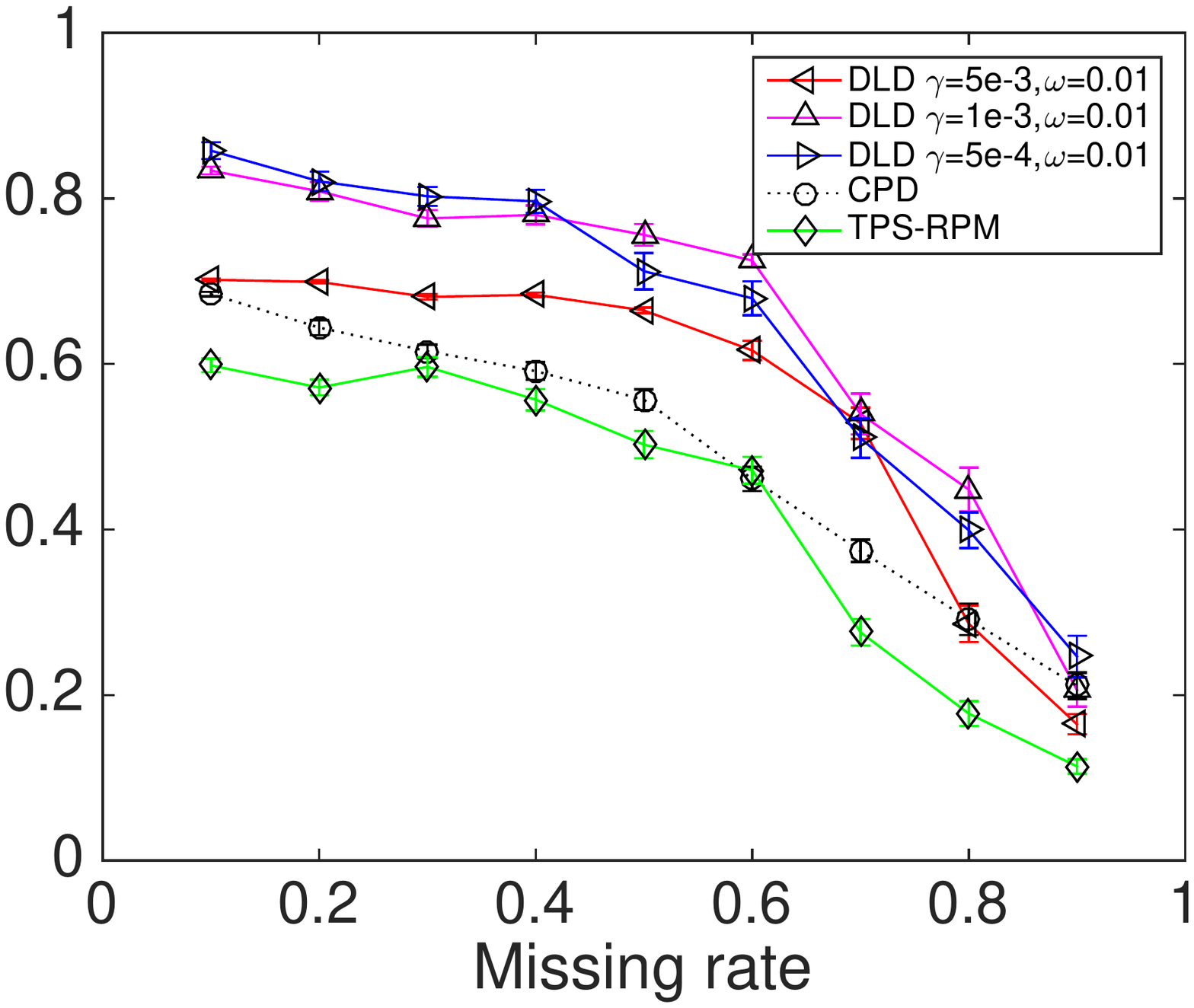} 
             \vspace{-3.0cm}\\ % &
         \raisebox{ 12.0\totalheight} {\bf Outliers}  
           & \hspace{-0.7cm} \includegraphics[width=0.32\textwidth]{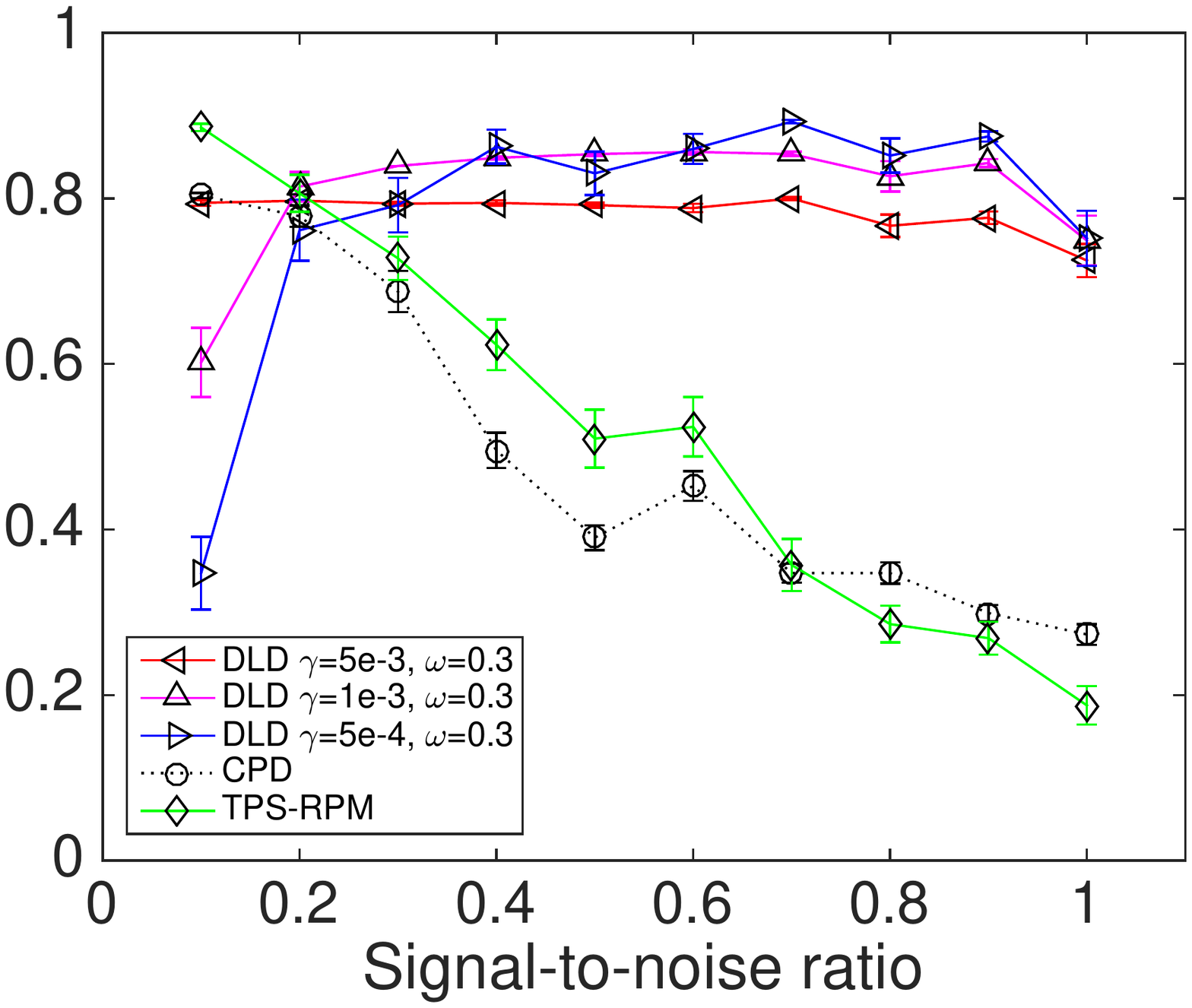}
             \hspace{-0.7cm} \includegraphics[width=0.32\textwidth]{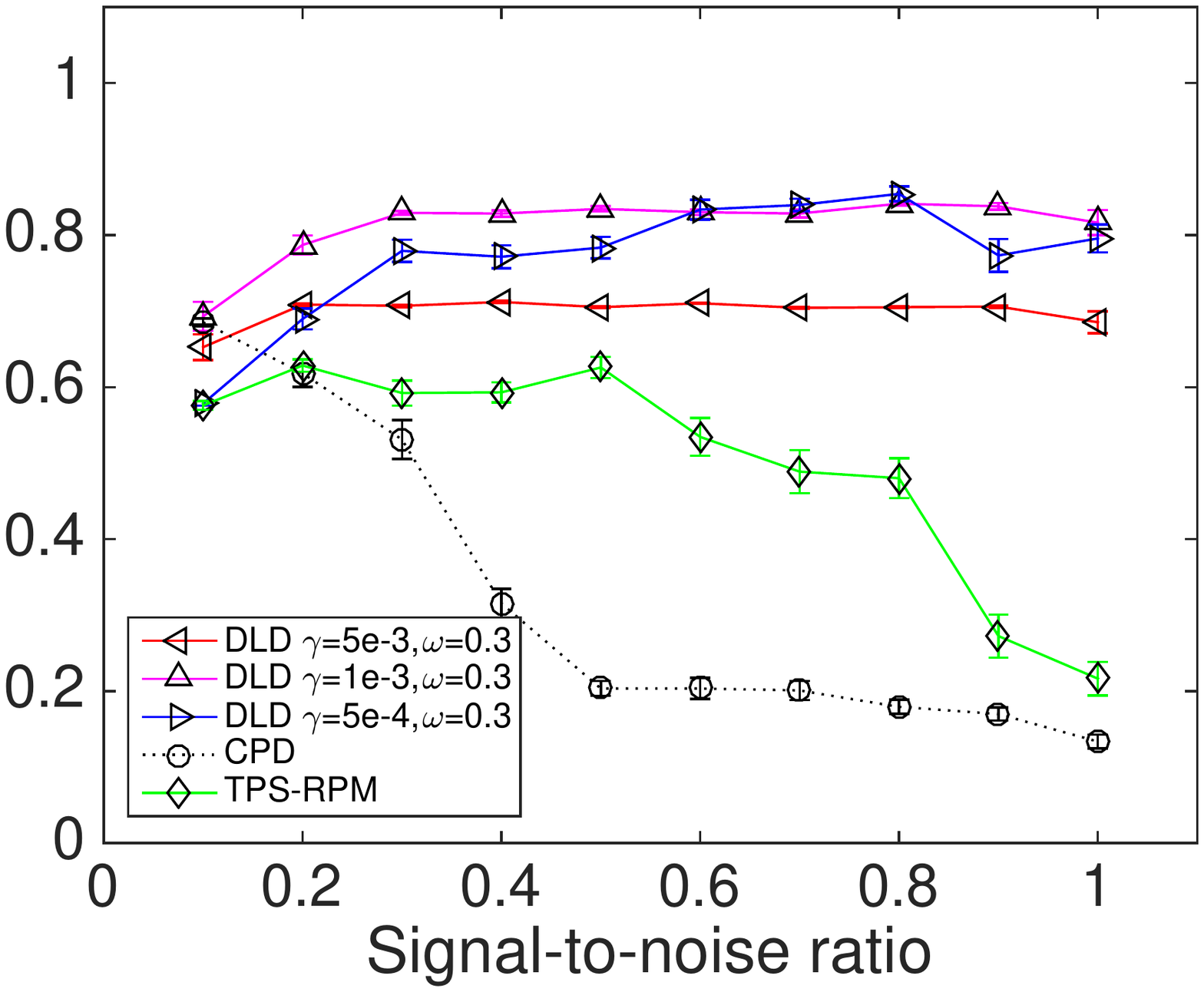}
             \vspace{-3.0cm}\\% &
         \raisebox{ 12.0\totalheight} {\bf Rotation}  
           & \hspace{-0.7cm} \includegraphics[width=0.32\textwidth]{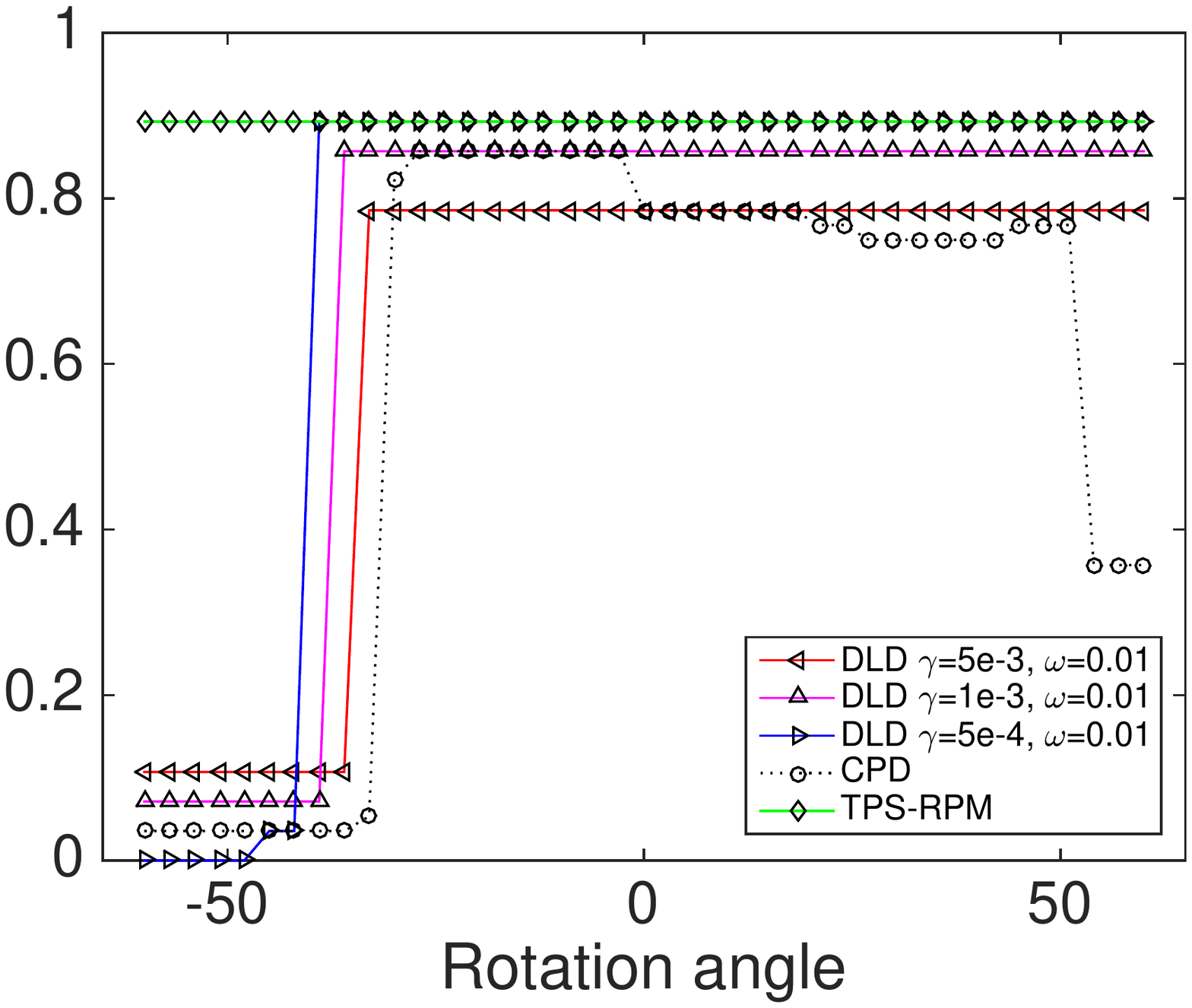} 
             \hspace{-0.7cm} \includegraphics[width=0.32\textwidth]{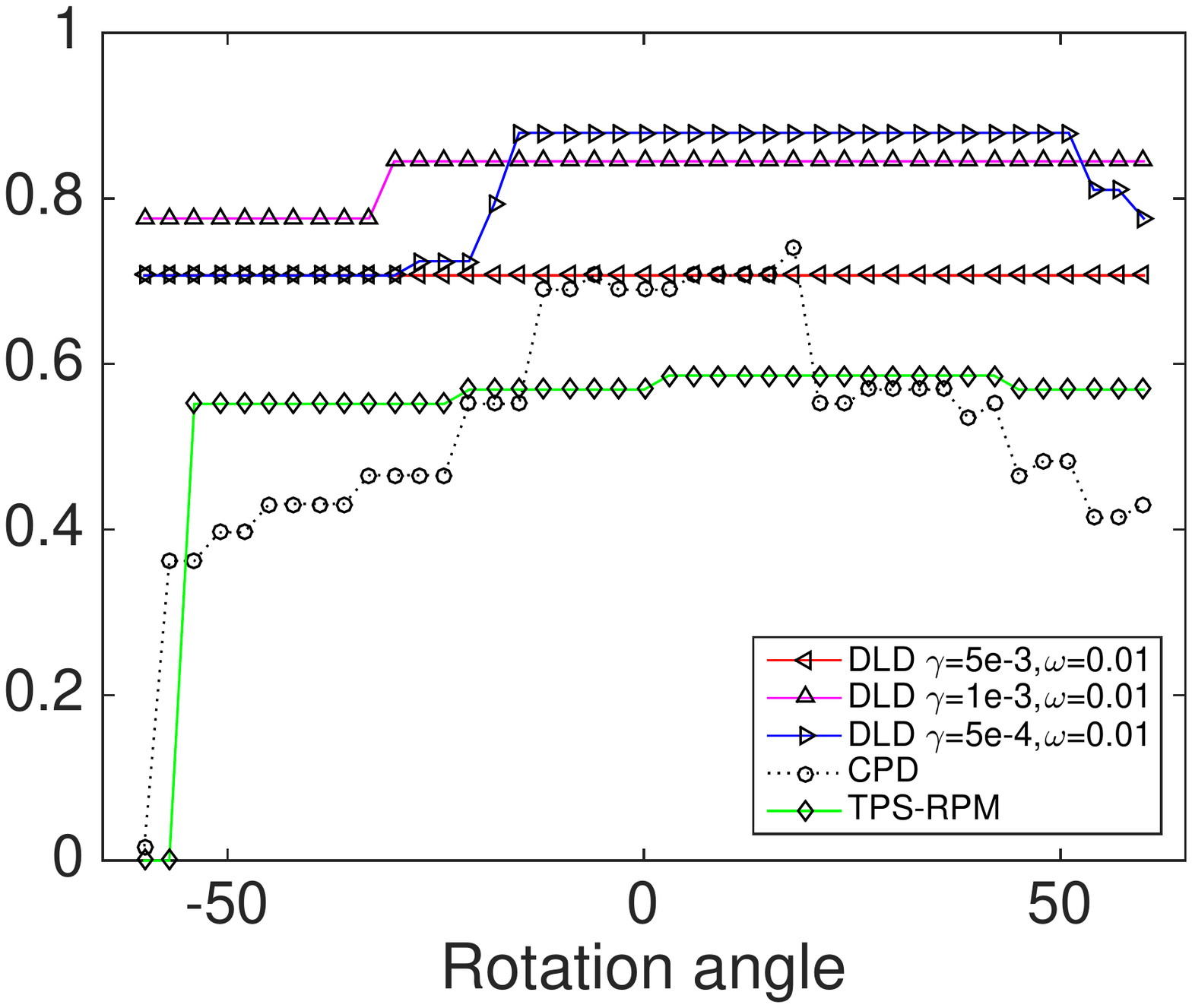} 
       \end{tabular}
    \end{center} \vspace{-1.5cm}
    \caption{ 
      Comparisons of registration performance among DLD, CPD, and TPS-RPM using hand No. $6$ in the IMM hand dataset (left),
      and face 22-4f in the IMM face dataset (right) with four types of modification: (1) replication of points with dispersion,
      (2) deletion of points, (3) addition of outliers, and (4) rotation of the whole shape.
     }
    \label{fig:comp}
  \end{figure*}
\fi

To evaluate the registration performance of the DLD more precisely, we compared the DLD, CPD, and TPS-RPM 
using the same IMM hand dataset and the IMM face dataset under various conditions. As the true target shapes 
to be estimated, we used hand No. $6$ and face $22$-$4$f shown in the first row of Figure \ref{fig:comp}.
For the IMM hand dataset, we used $39$ hand shapes
with hand No. $6$ removed from all $40$ shapes as a set of training data for the DLD, whereas 234 face shapes,
with all six faces of human No. $22$ removed, were used as the other set of training data for the DLD.

\subsubsection*{Generating validation datasets}
As in the experiment in the previous subsection, 
we generated four types of target data for hand No. $6$: 
(a) 20-time replication of target points with dispersion,
(b) random deletion of target points, 
(c) addition of outliers that follow uniform distributions, 
and (d) rotation of the whole shape.
In generating the four types of target data, we changed the following data generation parameters: 
(a) standard deviation of the Gaussian distribution for replicating target points, 
(b) missing rate, 
(c) signal-to-noise ratio, and 
(d) rotation angle.
To repeat the experiments and reduce the influence of randomness on registration performance, 
we generated target point sets 20 times for (a), (b), and (c) in the same setting.
For (d), a single set of target points was generated for each rotation angle,
since there is no randomness in the case of rotation.
 
\subsubsection*{Definition of registration accuracy}
The DLD, CPD, and TPS-RPM were applied to each target data item, and the registration accuracy was calculated 
for each. For target data (a), (b), and (c), the average and standard errors in the registration accuracy 
were calculated. Registration accuracy was defined as the rate of correct matching between a true target 
shape and a deformed mean shape following registration. More formally, it was defined as follows:
\begin{linenomath*}
\begin{align}
  \text{Accuracy} =\frac{\text{\#points with correct matching}}
                        {\text{\#points in the mean shape}    } \nonumber.
\end{align}
\end{linenomath*}
Point-by-point correspondence to compute registration accuracy was estimated using the nearest-neighbor 
matching between the deformed mean shape and the true target shape, i.e., the shape before
modifications such as deletion of points or addition of replicated points and outliers.

\subsubsection*{Choice of parameters}
We used the default parameters for the CPD and TPS-RPM implemented in their software. 
For the DLD, we fixed the number of shape variations to $K=10$ for the hand dataset and $K=30$ for the face dataset.
We set the parameter of the DLD $\omega$ to $0.01$ for (a), (b), and (d), whereas it was set to $0.30$ for (c) because
the former datasets did not include outliers and the latter included outliers.
We changed the parameter of the DLD $\gamma$ to $5.0\times 10^{-3}$, $1.0\times 10^{-3}$, and $5.0\times10^{-4}$
without the adaptive control of $\gamma$ to investigate the influence of $\gamma$ on registration performance.

\subsubsection*{Results} 
The second row in Figure \ref{fig:comp} 
shows the results of a comparison between target shapes and 
replicated target points. The $x$-axis represents the standard deviation of replicating target points
whereas the $y$-axis shows registration accuracy. 
For the hand datasets, the registration performance of DLD was insensitive to the regularization parameter
$\gamma$ and the DLD outperformed the CPD and TPS-RPM in almost all cases.
For face datasets, the DLD with $\gamma=5.0\times 10^{-3}$ achieved the best registration performance,
whereas the DLD with $\gamma=5.0\times 10^{-4}$ was less accurate than the CPD.

The results for target shapes with random missing points are shown in the third row of Figure \ref{fig:comp}.
The $x$-axis represents the missing rate and the $y$-axis registration accuracy.
The DLDs with $\gamma=1.0\times 10^{-3}$ and $5\times 10^{-4}$ were the most stable for the target shapes, 
and they outperformed CPD and TPS-RPM in nearly all cases. On the contrary, the DLD with $\gamma=5\times 10^{-3}$ was 
moderately less accurate than CPD for the hand dataset.
This suggests that a $\gamma$  %, which corresponds to the smoothness of the displacement field,
that is too large may, in some cases, reduce registration accuracy because of the large bias 
originating in the prior shape information.

The fourth row in Figure \ref{fig:comp} shows the results for target shapes with outliers following
a 2D uniform distribution with $x_1\in [0.0,1.2]$ and $x_2\in[0.0,1.2]$ for the hand dataset and 
that with $x_1\in [0.2,1.0]$ and $x_2\in [0.0,0.8]$ for the face dataset. 
The $x$-axis of each figure represents the signal-to-noise ratio and the $y$-axis represents registration accuracy. 
The DLD achieved the best registration performance for both datasets in nearly all cases, showing its robustness against outliers.
When the signal-to-noise ratio was $0.1$, the DLD was less accurate than the CPD and TPS-RPM.
This suggests that an inappropriate choice of noise probability $\omega$ reduces registration accuracy.

The fifth row in Figure \ref{fig:comp} shows the results for rotated target shapes. The $x$-axis and the
$y$-axis represent the rotation angle and registration accuracy, respectively. 
For the hand dataset, %TPS-RPM achieved the best registration performance
the registration accuracy of TPS-RPM was considerably stable and was unaffected by the 
rotation angles, at least in the range $[-\pi/3,\pi/3]$.
For the face dataset, the DLD achieved the best performance in nearly all cases,
suggesting its robustness to the rotation.

\section{Conclusion}
Many existing algorithms for solving point set registration problems 
employ the assumption of a smooth displacement field, that is,
%movements of neighboring points in a floating point set are correlated, 
the local geometry of the point set is preserved. 
Because of this assumption, registration problems are elegantly 
solved by these algorithms in various cases. This means that
the registration performance can be improved using prior shape information
instead of sacrificing the applicability of the algorithms.
%% Two keys to the success of point set registration algorithms are
%% (1) robustness to outliers and
%% (2) the assumption of a smooth displacement field.
%% %(3) simuletaneous estimation of pose and shape parameters.
In this paper, we proposed a novel point set registration algorithm based on
a Gaussian mixture model.
% where it explicitly models outliers as a probablistic distribution, 
%and thereby leads to the robustness to outliers.
We used a PCA-based statistical shape model to encode prior shape information,
which was combined with the assumption of a smooth displacement field. 
%% Therefore, our method is less appliable than general point set registration
%% algorithms but can be be more accurate due to the effective use of shape prior information
%% and the smooth displacement field.
The proposed algorithm works effectively if the target point set is rotated as 
location parameters corresponding to a similarity transformation are simultaneously estimated 
in addition to shape deformation.  Our algorithm is also scalable to
large point sets because of the presence of a linear-time algorithm,
which was verified through the application to the SCAPE dataset.
To evaluate the registration performance of the proposed algorithm, we compared \blue{
it with the supervised learning approaches, the GMM-ASM and an ICP with the Geman-McClure estimator
using the SCAPE and FLAME datasets. We also compared it with the the unsupervised learning approaches, }%
CPD and TPS-RPM using the IMM hand dataset and the IMM face dataset with four types of modifications:
replication of target points with dispersion, 
random deletion of target points, addition of outliers, and rotation of the entire shape.
The proposed algorithm outperformed \blue{the GMM-ASM, an ICP-based method with the Geman-McClure estimator},
CPD and TPS-RPM in almost all cases, showing its effectiveness for various types of point set registration problems.

\subsection*{Conflict of Interest}
None declared.

\bibliographystyle{elsarticle-num}
\bibliography{dld}

%\input{dld-bio.tex}
%%\bibliographystyle{plainnat}
%% \bibliography{dld}
%\input{dld.bbl}

\blue{
\section*{Appendix}
\subsection*{List of notations}
}

\blue{
\begin{itemize}
\setlength\itemsep{0.0mm}
\item $N$: the number of points in the target point set.
\item $M$: the number of points in the source point set (or the mean shape).
\item $D$: the dimension of the space in which point sets are embedded.
\item $K$: the number of shape variations in a statistical shape model.
\item $L$: the number of points to be sampled for the Nystr\"om approximation.
%% \item $B$: the number of the training shapes.
%% \item $x_n \in \mathbb{R}^D$: the $n$th point in the target point set $X=(x_1,\cdots,x_N)$.
%% \item $y_m \in \mathbb{R}^D$: the $m$th point in the floting point set $Y=(y_1,\cdots,y_M)$.
%% \item $u_m \in \mathbb{R}^D$: the $m$th point in the mean shape (i.e. flexible template) $\mb{u}=(u_1^T,\cdots,u_M^T)^T$.
%% \item $v_m \in \mathbb{R}^D$: the $m$th point in the deformed (or training) shape $\mb{v}=(v_1^T,\cdots,v_M^T)^T$.
%% \item $z_k $: the weight value for $k$th shape variation.
%% \item $\mb{h}_k \in \mathbb{R}^{MD}$: the $k$th shape variation.
%% \item $\mb{w} \in \mathbb{R}^{MD}$: a residual vector for a statistical shape model.
\item $X=(x_1,\cdots,x_N)^T\in \mathbb{R}^{N\times D}$: the target point set,
      where the $n$th point of which is denoted by $x_n\in\mathbb{R}^D$.
\item $Y=(y_1,\cdots,y_M)^T\in \mathbb{R}^{M\times D}$: the source point set, 
      where the $m$th point of which is denoted by $y_m\in\mathbb{R}^D$.
\item $U=(u_1,\cdots,u_M)^T\in \mathbb{R}^{M\times D}$: the matrix notation of the mean shape, 
      where the $m$th point of which is denoted by $y_m\in\mathbb{R}^D$.
\item $\mb{u}=(u_1^T,\cdots,u_M^T)^T\in \mathbb{R}^{MD}$: the vector notation of the mean shape.
\item $\mb{H}=(\mb{h}_1,\cdots,\mb{h}_K)\in \mathbb{R}^{MD\times K}$: the shape variation matrix,
      where $\mb{h}_k\in \mathbb{R}^{MD}$ is the $k$th leading shape variation.
\item $z =(z_1,\cdots,z_K)^T\in \mathbb{R}^{K} $: the weight vector corresponding to $K$ shape variations.
%% \item $\mb{v}=(v_1^T,\cdots,v_M^T)^T\in \mathbb{R}^{MD}$: the vector notation of a deformed (or training) shape.
%% \item $\mb{v}_j\in \mathbb{R}^{MD}$: the vector notation of the $j$th training shape.
%% \item $\mb{\bar{v}}\in \mathbb{R}^{MD}$: the vector notation of the mean shape calculated from training shapes.
%% \item $\mb{C}\in \mathbb{R}^{M\times M}$: the shape covariance matrix calculated from training shapes.
%% \item $H_m\in \mathbb{R}^{D\times K}$: the submatrix of $\mb{H}$ that corresponds to the $m$th point in the mean shape.
\item $\sigma^2$: the residual variance of a Gaussian distribution that constitutes the Gaussian mixture model.
\item $\omega$: the outlier probability.
\item $\gamma$: the regularization parameter for controlling the smoothness of the displacement field.
\item $N_P$: the effective number of matching points in the target point set. %that match points in the target point set.
\item $\mathcal{T}(y_m;\theta)$: a transformation model for mapping a source point set $Y$ to a target point set $X$
      where $\theta$ is a set of parameters that characterizes the transformation.
\item $\rho=(s,R,d)$: the parameter set that defines the similarity transformation, where $s$ is the scaling factor,
      $R\in\mathbb{R}^{D\times D}$ is the rotation matrix, and $d\in\mathbb{R}^{D}$ is the translation vector.
%% \item $s$: the scaling value that constitues the definition of the similarity transformation.
%% \item $R\in \mathbb{R}^{D\times D}$: the rotation matrix that constitues the definition of the similarity transformation.
%% \item $d\in \mathbb{R}^D$: the translation vector that constitues the definition of the similarity transformation.
\item $\Theta=(\theta, \sigma^2)$: the whole parameter set of a Gaussian mixture model.
%% \item $\theta$: the parameter set of the transformation model $\mathcal{T}$.
%% \item $p_{mn}=p(m|x_n;\Theta)$: the posterior probability that $x_n$ matches $y_m$ (or $u_m$).
\item $P=(p_{mn})\in \mathbb{R}^{M\times N}$: the probability matrix, the $mn$th element of which is
      the posterior probability that $x_n$ matches $y_m$ (or $u_m$).
%% \item $\lambda_k$: the eigenvalue corresponding to the $k$th leading shape variation.
%% \item $\Lambda_K=\text{d}(\lambda_1,\cdots,\lambda_K)\in \mathbb{R}^{K\times K}$: 
%%       the diagonal matrix, the $k$th diagonal elment of which is $\lambda_k$.          
\item $S$: the volume of the region in which outliers can be generated.
\item $V$: the sampled point set from the combined point set $Y\cup X$ for the Nystr\"om approximation.
\item $K_{YX}=(k_{mn})\in \mathbb{R}^{M\times N}$: the Gaussian affinity matrix between point sets $Y$ and $X$, 
      where $k_{mn}=\exp(-||y_m-x_n||^2/2\sigma^2)$.
%% \item $K_{YV}$: the Gaussian affinity matrix between point sets $Y$ and $V$.
%% \item $K_{VX}$: the Gaussian affinity matrix between point sets $V$ and $X$.
\end{itemize}
}

\end  {document}